\documentclass[pdflatex,sn-mathphys,iicol]{sn-jnl}

\usepackage{
graphicx,
amsfonts,
textcomp,
program,
amssymb,
listings,
rotating,
booktabs,
algorithmicx,
mathrsfs,
url,
appendix,
etoolbox,
algpseudocode,
multirow,
algorithm,
amsmath,
hyperref,
manyfoot,
hhline,
pifont}

\usepackage[export]{adjustbox}

\newcommand\shline{\specialrule{0.85pt}{0pt}{0pt}}

\jyear{2022}
\theoremstyle{thmstyleone}

\theoremstyle{thmstyletwo}

\theoremstyle{thmstylethree}

\begin{document}

\title[Article Title]{Deep Physics-Guided Unrolling Generalization for Compressed Sensing}

\author[]{\fnm{Bin} \sur{Chen}}\email{\scriptsize chenbin@stu.pku.edu.cn}
\author[]{\fnm{Jiechong} \sur{Song}}\email{\scriptsize songjiechong@pku.edu.cn}
\author[]{\fnm{Jingfen} \sur{Xie}}\email{\scriptsize xiejf@stu.pku.edu.cn}
\author*[]{\fnm{Jian} \sur{Zhang*}}\email{zhangjian.sz@pku.edu.cn}
\affil[]{School of Electronic and Computer Engineering, Shenzhen Graduate School, \\Peking University, Shenzhen, China}


\abstract{By absorbing the merits of both the model- and data-driven methods, deep physics-engaged learning scheme achieves high-accuracy and interpretable image reconstruction. It has attracted growing attention and become the mainstream for inverse imaging tasks. Focusing on the image compressed sensing (CS) problem, we find the intrinsic defect of this emerging paradigm, widely implemented by deep algorithm-unrolled networks, in which more plain iterations involving real physics will bring enormous computation cost and long inference time, hindering their practical application. A novel deep \textbf{P}hysics-guided un\textbf{R}olled recovery \textbf{L}earning (\textbf{PRL}) framework is proposed by generalizing the traditional iterative recovery model from image domain (ID) to the high-dimensional feature domain (FD). A compact multiscale unrolling architecture is then developed to enhance the network capacity and keep real-time inference speeds. Taking two different perspectives of optimization and range-nullspace decomposition, instead of building an algorithm-specific unrolled network, we provide two implementations: \textbf{PRL-PGD} and \textbf{PRL-RND}. Experiments exhibit the significant performance and efficiency leading of PRL networks over other state-of-the-art methods with a large potential for further improvement and real application to other inverse imaging problems or optimization models.}


\maketitle

\section{Introduction}
\label{sec:intro}
Inverse imaging is to reconstruct the latent clean image signal $\mathbf{x}\in\mathbb{R}^N$ from its noisy degraded measurement $\mathbf{y}\in \mathbb{R}^M$ with the acquisition model $\mathbf{y}= \mathbf{Ax}+\mathbf{\epsilon}$, where $\mathbf{A}:\mathbb{R}^N\rightarrow \mathbb{R}^M$ is the forward operator and $\mathbf{\epsilon}\in\mathbb{R}^M$ is the observation noise. There are wide applications with different forms of $\mathbf{A}$, like the classic image denoising with identity mapping (\textit{i.e.}, $\mathbf{A}=\mathbf{I}_N$), super-resolution (SR) with filtered downscaling operators, compressed sensing (CS) \cite{donoho2006compressed,candes2008introduction} with Gaussian matrices, accelerated magnetic resonance image (MRI) \cite{lustig2008compressed,lustig2007sparse} with subsampled Fourier transforms and sparse-view computed tomography (CT) \cite{szczykutowicz2010dual} with sparse-angle Radon transforms. In this paper, we focus on the typical block-based (or block-diagonal) image CS scheme \cite{gan2007block, mun2009block, fowler2012block, chen2022content,song2023dynamic} that divides the high-dimensional natural image into non-overlapped $B\times B$
blocks and obtains measurements block-by-block with a small-sized sampling matrix $\mathbf{A}$\footnote{{For reproducible research, the complete source code and pre-trained models of our PRL networks will be made available soon at \url{https://github.com/Guaishou74851/PRL}.}}. And we keep this work extensible to other inverse imaging tasks.

Generally, the ill-posedness of such an under-determined linear system with $M\ll N$ leads to the extreme difficulty of recovering $\mathbf{x}$ from only the observed $\mathbf{y}$. Here the CS ratio (or sampling rate) is denoted as $\gamma =M/N$. Following \cite{chen2020deep, chen2021equivariant}, we call the operators utilized in the signal acquisition process the \textit{``sampling physics"} or simply \textit{``physics"}. Specifically, in the context of CS imaging, the sampling matrix $\mathbf{A}$ is the ``physics" containing quite critical information for reconstruction from the acquired data $\mathbf{y}$. In the early stage, there are many traditional model-driven methods trying to exploit the prior knowledge of images to construct the regularizers with strong theoretical guarantees (e.g. sparsity \cite{zhao2016reducing,zhao2016video,zhang2014image,elad2010sparse,nam2013cosparse} and low-rank \cite{dong2014compressive,cai2010singular,long2019low,liu2018image,liu2019low,long2021bayesian,long2022trainable}), but they may suffer from the high computational cost and extensive empirical fine-tunings for obtaining the optimal hyperparameters and transforms.

\begin{figure*}[!t]
    \centering
    \includegraphics[width=1.0\textwidth]{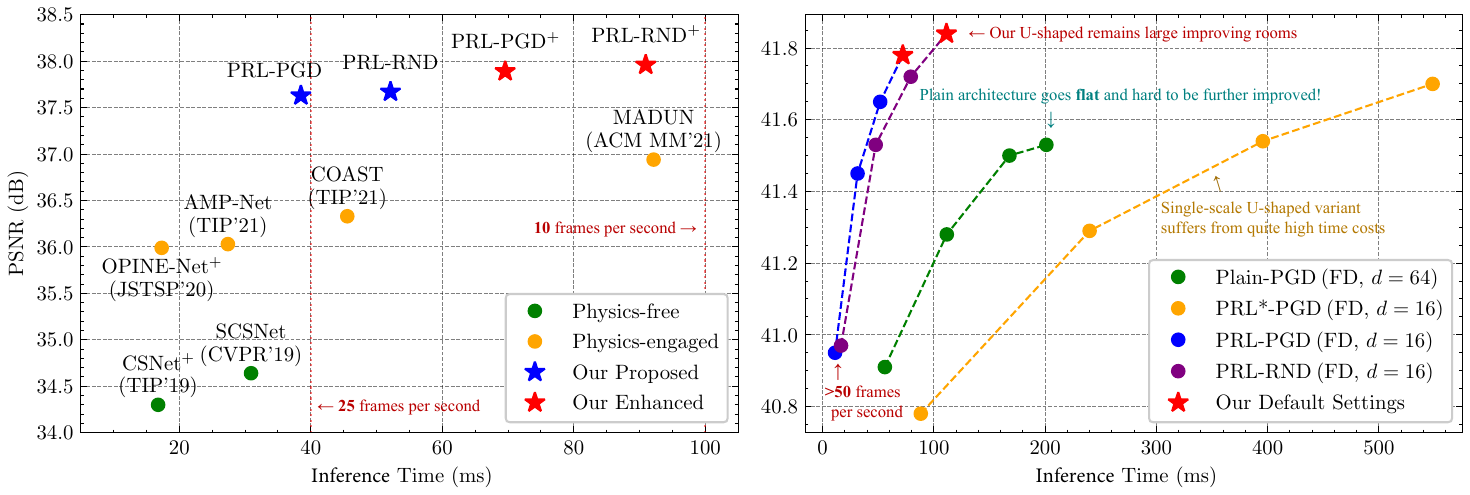}
	\caption{Comparisons of the PSNR on Set11 \cite{kulkarni2016reconnet} and the inference time for $256 \times256$ image on an 1080Ti GPU of various methods with CS ratio $\gamma=30\%$ (left) and different unrolled architectures with $\gamma=50\%$ (right). Our PRL networks exceed the other methods with remaining a large accuracy margin. They explore the improvement potential in both the optimization model- and network-level, and keep the real-time inferences faster than the ones under traditional plain unrolling scheme.}
	\label{fig:comp_time_PSNR}
\end{figure*}

In the past decade, convolutional neural network (CNN) greatly improves the image restoration accuracy and speed \cite{ravishankar2019image}. Many end-to-end networks \cite{mousavi2015deep,kulkarni2016reconnet,shi2019image,sun2020dual,shi2019scalable,fan2022global} can reconstruct at very high speeds but may leave some undesired effects and remain their intrinsic ``black-box" property \cite{huang2018some}. Some optimization-inspired networks \cite{sun2016deep,zhang2018ista,zhang2020optimization,zhang2020amp,chen2022content,zhang2023physics} break the limitation of CS domain insights and combine the merits of both model- and data-driven approaches by training a truncated deep unrolled inference. Davis \textit{et al.} \cite{gilton2019neumann} and D. Chen \textit{et al.} \cite{chen2020deep} employ the frameworks of Neumann series and range-nullspace decomposition of the forward operator, respectively, and provide two novel perspectives of sampling information utilization. These physics-engaged methods have become the mainstream for constructing well-defined interpretable structures with high recovery accuracy.

Recently, some memory augmentation mechanisms \cite{chen2020learning,song2021memory} are developed with the consideration of the information transmission limitations. Some efficient controllable and lightweight networks \cite{you2021coast,you2021ista} become a new baseline in the image CS field and are flexible and robust to the change of sampling rate. Z. Chen \textit{et al.} \cite{chen2021deep} propose a tailored residual-regressive network and a deep dilated residual channel attention network (DRCAN) to perform distortion-aware proximal mappings.

Nevertheless, most physics-engaged networks focus on learning the powerful proximal operators (denoisers) to improve the regularization with complicated image-feature fusion, but less notice the intrinsic restrictions of the existing optimization models in the \textbf{I}mage \textbf{D}omain (\textbf{ID}) (with insufficient single-channel physics utilization) and the weak representation ability of the plain architectures derived from iterative steps (with strictly sharing the same structure across all the unrolled stages). They cause performance degradation and saturation when increasing the plain unrolled iteration number, network depth, or feature channel numbers to improve the reconstruction quality. In this paper, we propose a novel deep \textbf{P}hysics-guided un\textbf{R}olled recovery \textbf{L}earning (\textbf{PRL}) framework for CS, which learns to recover a high-dimension feature instead of only an image, with generalized multiscale perception and fully-activated physics injections to resolve the existing obstacles. Benefiting from the simple and effective U-shaped residual designs \cite{he2016deep,ronneberger2015u,zhang2021plug}, our two PRL networks, derived from the proximal gradient descent algorithm \cite{parikh2014proximal,lefkimmiatis2018universal} and the range-nullspace decomposition of image signal \cite{chen2020deep}, respectively, try to exploit the sufficient utilization of physics in the multiscale \textbf{F}eature \textbf{D}omain (\textbf{FD}). They manifest significant superiority in performing efficient block-based image CS recoveries (see Fig.~\ref{fig:comp_time_PSNR}).

Our main contributions are three-fold:

\vspace{1pt}
\noindent \ding{113}~(1) We propose a deep physics-guided unrolled image CS recovery framework dubbed PRL. We first discover the serious problems of traditional deep unrolling paradigm in both the optimization model- and network-level and then generalize the existing single-scale/-path inference in image domain (ID) with poor physical utilization to the multiscale high-throughput unrolling in feature domain (FD) with sufficient physics injections.

\vspace{2pt}
\noindent \ding{113}~(2) Instead of constructing an algorithm-specific network, we provide two PRL implementations: \textbf{PRL-PGD} and \textbf{PRL-RND} from the perspectives of optimization and range-nullspace decomposition. They are both based on some widely used simple modules and can be easily further enhanced with remaining a large room for improvement.

\vspace{2pt}
\noindent \ding{113}~(3) We conduct extensive natural image CS, sparse-view CT, and one-bit CS experiments, validate the significant superiority of our proposed PRL framework by comparing it with the other state-of-the-art methods, and try to provide a clear understanding of its working principle.

\section{Background}
\label{sec:bg}
This section retrospects the deep learning-based CS methods, the proximal gradient descent (PGD) algorithm and the range-nullspace decomposition (RND) of any target image signal to facilitate our following-up analyses and discussions.

\subsection{Deep Learning for CS Imaging}
\label{sec:related_work}
Following \cite{chen2020deep}, here we group the existing deep learning-based CS approaches into the \textit{physics-free} and \textit{physics-engaged} methods with respect to the utilization of the physical knowledge (\textit{i.e.}, $\{\mathbf{A},\mathbf{y}\}$) and focus on the methods relevant to our own.

\textbf{Physics-free Methods:} By treating the CS reconstruction as a common de-aliasing problem, deep physics-free methods directly learn an inverse mapping from $\mathbf{y}$ to $\mathbf{x}$ with little knowledge about the CS sampling process. Kulkarni \textit{et al.} \cite{kulkarni2016reconnet} learn a CNN to regress an image block from its measurement. Shi \textit{et al.} \cite{shi2019image} and Sun \textit{et al.} \cite{sun2020dual} map the block measurements to the jointly recovered patches to avoid blocking artifacts. Considering the trade-off between system flexibility and complexity, Shi \textit{et al.} \cite{shi2019scalable} propose a scalable CNN and a heuristic greedy algorithm to obtain the importance order of different matrix rows. These methods enjoy their intuitive structures and fast inferences but suffer from the black-block property, difficult training and the defect of delicacy.

\begin{figure*}[!t]
	\centering
 \includegraphics[width=1.0\textwidth]{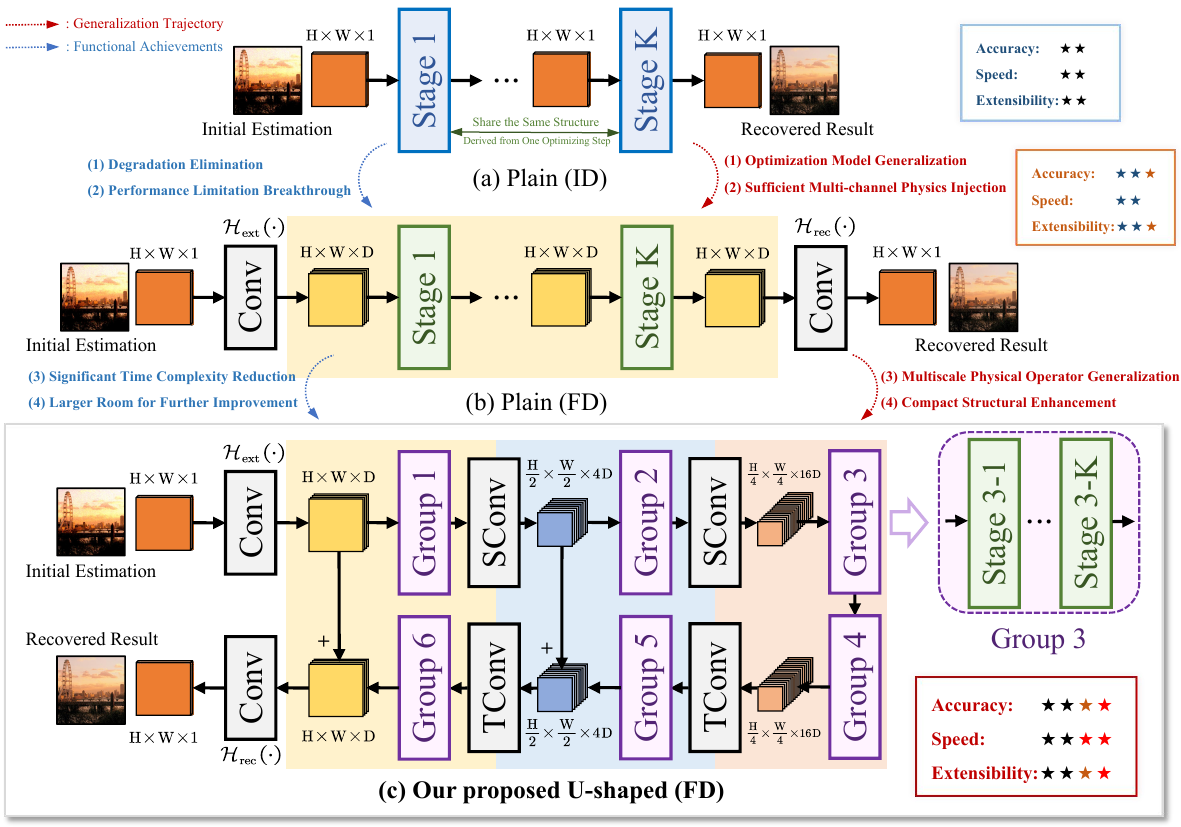}
	\caption{The previous feed-forward recovery architecture in image domain (ID) (a), generalized plain architecture in feature domain (FD) (b), and our generalized multiscale high-throughput architecture in FD (c). The processings with different background colors correspond to the features with different spatial downscalings and a unified capacity (\textit{i.e.}, $H\times W\times D$).}
	\label{fig:arch}
\end{figure*}

\textbf{Physics-engaged Methods:} With introducing the explicit physics guidance, deep physics-engaged methods feed the information of sampling matrix and measurement into the network modules. They try to take advantage of both the optimization model- and network-based methods. As an emerging paradigm for image restoration, the deep unrolling technique has been adopted in denoising \cite{chen2016trainable,lefkimmiatis2017non,lefkimmiatis2018universal}, deblurring \cite{kruse2017learning,wang2020stacking}, etc. \cite{kokkinos2018deep,zhang2020deep}, which generally formulates the CS imaging task as the following optimization problem:
\begin{equation}
	\setlength{\abovedisplayskip}{2pt}
	\setlength{\belowdisplayskip}{2pt}
	\label{eq:opt_image}
	\hat{\mathbf{x}}=\underset{\mathbf{x}}{\arg\min}f\left( \mathbf{x};\mathbf{A},\mathbf{y} \right) +\lambda g\left( \mathbf{x} \right),
\end{equation}
where $f\left( \mathbf{x};\mathbf{A},\mathbf{y} \right)$ is the data fidelity term guaranteeing the sampling consistency, and $\lambda g\left( \mathbf{x} \right)$ is the prior term assumed to be convex in general. Based on the traditional iterative shrinkage-thresholding algorithm (ISTA) \cite{blumensath2009iterative}, J. Zhang \textit{et al.} \cite{zhang2018ista} construct a 9-stage unrolled network and learn the nonlinear $\ell_1$-sparsifying transforms. J. Zhang \textit{et al.} \cite{zhang2020optimization} and Z. Zhang \textit{et al.} \cite{zhang2020amp} further integrate the data-driven learned matrix with a multi-block training strategy. You \textit{et al.} \cite{you2021coast} propose a controllable CS network that is agnostic to the sampling matrix. Song \textit{et al.} \cite{song2021memory,song2023deep} develop two inter-stage memory mechanisms to augment the information flow and compensate the high-frequency details to improve the quality of textures.

\textbf{Weaknesses:} Although the above approaches achieve impressive results, there exist two major weaknesses hindering their improvements. First, their empirical network-level developments may lack deep insights in the optimization model-level. They do not notice or break the natural limitations of traditional image domain reconstruction with the single-channel insufficient physics utilization, which leads to the difficulty in extending and the performance degradations \cite{he2016deep}. Second, the widely adopted plain single-scale/-path architectures derived from the iterative optimization steps employ the same form of operations to refine the images/features in different stages (see Fig.~\ref{fig:arch}(a)), thus restricting the improvements in network efficiency and deep representation ability. However, we argue and validate that by combining the generalization of optimization models, sufficient physics injection and advanced multiscale compact architecture, there still exists a large room for improvement for the physics-engaged networks both in recovery accuracy and efficiency.

\subsection{Proximal Gradient Descent}
\label{sec:pgd}
The proximal gradient descent (PGD) is a first-order approach suited for many large-scale linear inverse problems \cite{parikh2014proximal}, which solves the Eq.~(\ref{eq:opt_image}) by iterating between the following two update steps:
\begin{equation}
	\setlength{\abovedisplayskip}{2pt}
	\setlength{\belowdisplayskip}{2pt}
	\label{eq:pgd1}
	\mathbf{z}^{\left( k \right)}=\hat{\mathbf{x}}^{\left( k-1 \right)}-\rho \nabla f\left( \hat{\mathbf{x}}^{\left( k-1 \right)};\mathbf{A,y} \right),
	\vspace{-1pt}
\end{equation}
\begin{equation}
	\setlength{\abovedisplayskip}{2pt}
	\setlength{\belowdisplayskip}{2pt}
	\label{eq:pgd2}
	\hat{\mathbf{x}}^{\left( k \right)}=\mathtt{prox}_{\lambda g}\left( \mathbf{z}^{\left( k \right)} \right),
	\vspace{-2pt}
\end{equation}
where $k$ denotes the index of iteration, and $\rho$ is the step size. The Eq.~(\ref{eq:pgd1}) conducts the gradient descent process, while the Eq.~(\ref{eq:pgd2}) is the proximal mapping which is quite critical to solving Eq.~(\ref{eq:opt_image}) efficiently by combining with the prior knowledge.

\subsection{Range-Nullspace Decomposition}
\label{sec:rnd}
Given an orthogonal matrix $\mathbf{A}\in \mathbb{R}^{M\times N}$ satisfying $\mathbf{A}\mathbf{A}^\top =\mathbf{I}_M$\footnote{In the field of CS imaging, the condition of orthogonality $\mathbf{A}\mathbf{A}^\top =\mathbf{I}_M$ is widely implemented by orthogonalizing an i.i.d. random Gaussian matrix or a Hadamard matrix \cite{zhang2014group}.}, we define the nullspace of $\mathbf{A}$ as: $\mathcal{N}\left( \mathbf{A} \right) =\left\{ \mathbf{v}\vert\mathbf{v}\in \mathbb{R}^N,\mathbf{Av}=\mathbf{0}_M \right\}$ and the range space of $\mathbf{A}^\top$ as: $\mathcal{R}\left( \mathbf{A}^{\top} \right) =\left\{ \mathbf{A}^{\top}\mathbf{w}\vert\mathbf{w}\in \mathbb{R}^M \right\}$, then it holds that $\mathbb{R}^N=\mathcal{N}(\mathbf{A})\oplus \mathcal{R}\left( \mathbf{A}^{\top} \right)$, where $\oplus$ is the direct sum. For any $\mathbf{x}\in \mathbb{R}^N$, there is a unique range-nullspace decomposition (RND) \cite{chen2020deep,chen2021equivariant}:
\begin{equation}
	\setlength{\abovedisplayskip}{3pt}
	\setlength{\belowdisplayskip}{3pt}
	\label{eq:rnd}
	\mathbf{x}={\mathbf{A}^{\top}\mathbf{Ax}}+{\left( \mathbf{I}_N-\mathbf{A}^{\top}\mathbf{A} \right) \mathbf{x}},
\end{equation}
where $\mathbf{x}_\mathcal{R}=({\mathbf{A}^{\top}\mathbf{Ax}})\in\mathcal{R}(\mathbf{A}^\top)$ is the range space component of $\mathbf{x}$, and its nullspace component is denoted as: $\mathbf{x}_\mathcal{N}=\left[{\left( \mathbf{I}_N-\mathbf{A}^{\top}\mathbf{A} \right) \mathbf{x}}\right]\in\mathcal{N}(\mathbf{A})$, \textit{i.e.}, $\forall \mathbf{x}\in\mathbb{R}^N$, there holds that $\mathbf{x}=\mathbf{x}_\mathcal{R}+\mathbf{x}_\mathcal{N}$.

\begin{figure*}[!t]
    \centering
    \hspace{-8pt}\scalebox{0.96}{
    \begin{minipage}{0.72\textwidth}
    \includegraphics[width=1.0\textwidth]{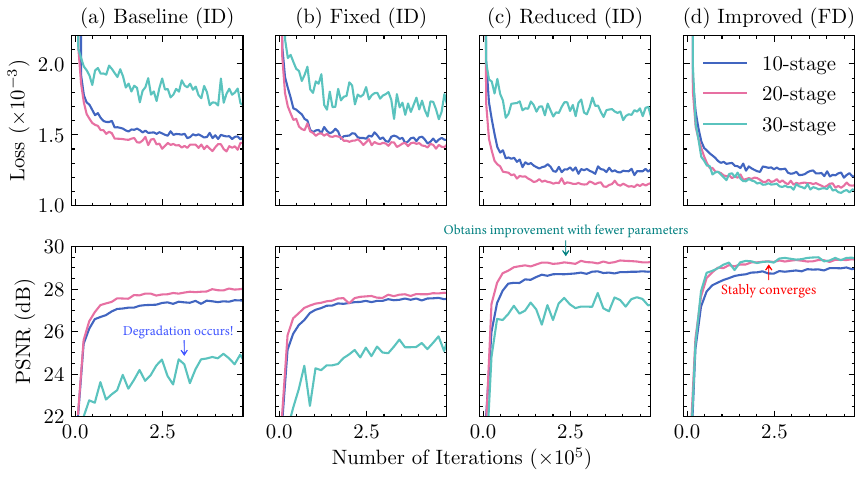}
    \end{minipage}}
    \begin{minipage}{0.27\textwidth}
    \centering \footnotesize{PSNR(dB)/\#Param.(M)\\}\normalsize
    \scalebox{0.86}{\begin{tabular}{c|ccc}
	\shline
    $K$                  & 10     & 20     & 30     \\ \hline
    \multirow{2}{*}{(a)} & 27.66  & 28.19  & \textcolor{purple}{24.59 (!)}  \\
                         & /1.479 & /2.958 & /4.437 \\
    \multirow{2}{*}{(b)} & 27.65  & 28.01  & \textcolor{purple}{25.75 (!)}  \\
                         & /1.478 & /2.957 & /4.435 \\
    \multirow{2}{*}{(c)} & 28.90  & 29.31  & \textcolor{purple}{27.38 (!)}  \\
                         & /1.477 & /2.954 & /4.431 \\
    \multirow{2}{*}{(d)} & 29.19  & 29.53  & 29.51  \\
                         & /1.645 & /3.288 & /4.932 \\ \shline
    \end{tabular}}
    \end{minipage}
	\caption{Training loss (left top), test PSNR (left bottom) and the final evaluation with parameter number (M) (right) on Set11 \cite{kulkarni2016reconnet} of the baseline (a), the stage last Conv fixed (b), the reduced with maximized information flow (c) and our improved (d) in the case of $\gamma =10\%$. An important characteristic of the degradation is that the training and test errors are both getting higher with larger depths \cite{he2016deep}. As stage number grows, (a), (b) and (c) all inevitably tend to degrade (marked by ``!"). Compared with (a) and (b), the transmission maximized (c) alleviates the inter-stage information bottleneck. Our generalized unrolling scheme in feature domain (d) further eliminates the degradation with stable convergence processes.}
	\label{fig:Train400}
\end{figure*}

\begin{figure*}[!t]
    \centering	\includegraphics[width=1.0\textwidth]{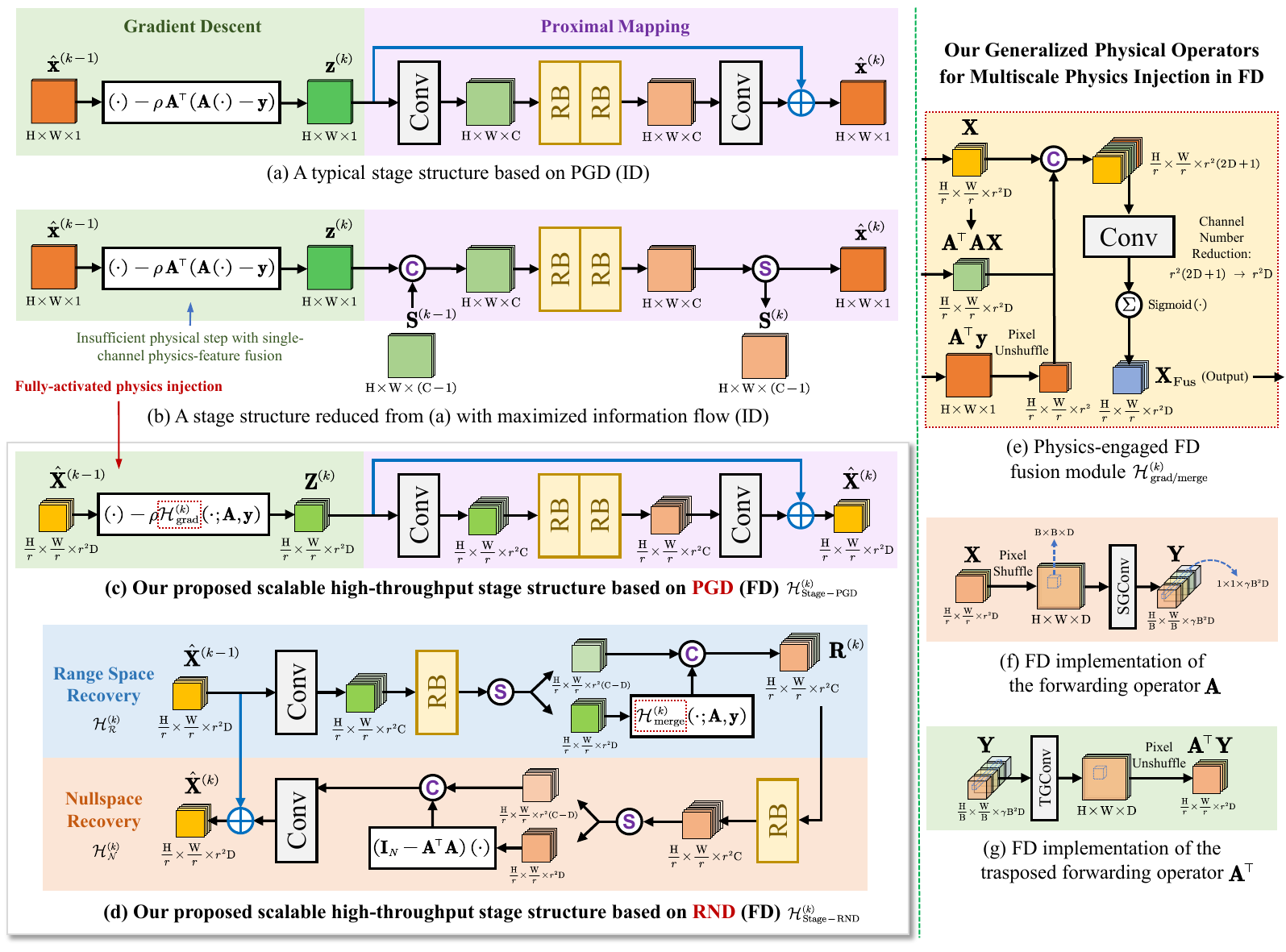}
	\caption{\textbf{(1) Left:} A typical traditional image domain (ID) stage design (a), its reduced and transmission-augmented version (b), and our scalable high-throughput structures based on PGD (c) and RND (d) in the feature domain (FD), denoted by $\mathcal{H}_\text{Stage-PGD}^{(k)}$ and $\mathcal{H}_\text{Stage-RND}^{(k)}$. The intra-stage skip connections are marked in blue. ``C", ``S" and ``+" denote the channel-wise concatenation, split and element-wise addition, respectively. \textbf{(2) Right:} Our generalized operators in FD, including the physics-engaged fusion module with an activation $\Sigma(\cdot)$ (e), the FD forward operator $\mathbf{A}(\cdot)$ (f) and its transpose $\mathbf{A}^\top(\cdot)$ (g) without introducing extra parameters by taking the sampling matrix $\mathbf{A}$ as the kernel weights for all groups of convolutions.}
	\label{fig:stage_PRL}
\end{figure*}

\section{Deep Physics-Guided Unrolling Generalization}
\label{sec:DPRL}
\subsection{Adaptive Feature Domain Reconstruction}
\label{sec:Adaptive Feature Reconstruction}
We consider the network-level effect brought by the iterative solutions of Eq.~(\ref{eq:opt_image}) in the previous deep unrolled networks \cite{zhang2018ista,zhang2020optimization,zhang2020amp,you2021coast,you2021ista,song2021memory,chen2021deep}. Essentially, they provide the decoupled paradigms to embed and supplement the physics of forward operator $\mathbf{A}$, measurement $\mathbf{y}$ and the learned image prior into the network trunk in a repetitive interpretable stage-by-stage manner. However, we discover some serious limitations of such a framework working in the image domain (ID) (with the single-channel image-level physics-feature fusion derived from the traditional optimization models) when increasing the capacity for performance improvement.

\textbf{Identify limitations by pilot experiment.} Following the above algorithm-unrolled methods \cite{zhang2018ista,zhang2020optimization,you2021coast,you2021ista}, we build a block-based CS baseline network with a widely adopted plain single-path architecture in Fig.~\ref{fig:arch}(a). It consists of $K$ iterative stages in series, and each unrolled stage takes the $\ell_2$ data fidelity $f\left( \cdot; \mathbf{A},\mathbf{y}  \right) =\frac{1}{2}\lVert \mathbf{A}\left( \cdot \right) -\mathbf{y} \rVert _{2}^{2}$ in Eq.~(\ref{eq:opt_image}) and maps PGD steps in Eq.~(\ref{eq:pgd1}) and Eq.~(\ref{eq:pgd2}) to form two separated parts in Fig.~\ref{fig:stage_PRL}(a). The first gradient descent part is analytic with $\nabla f\left( \cdot; \mathbf{A},\mathbf{y} \right) =\mathbf{A}^{\top}\left( \mathbf{A}\left( \cdot \right) -\mathbf{y} \right)$. The second proximal mapping part is implemented by a Conv layer, two residual blocks (RB) and another Conv layer to learn the ID residual with extraction, enhancement and restoration processes. Each RB adopts the classical ``Conv-ReLU-Conv" structure with an identity skip connection. Given a fixed random Gaussian matrix $\mathbf{A}$ and initialization $\hat{\mathbf{X}}^{(0)}=\mathbf{A^\top y}$ \cite{mousavi2017learning}, we evaluate this simple yet typical baseline by training three baseline networks with $C\equiv 64$ and $K\in \left\{10,20,30\right\}$ and report the loss values of the learning processes and the test results in Fig.~\ref{fig:Train400}(a). Although the intra-stage residual learning is introduced and works well when $K\le 20$ with $\le 120$ Conv layers, the notorious \textbf{degradation problem} \cite{he2016deep} emerges when we further extend $K$ to 30\footnote{Note that this phenomenon indicates that our baseline network is ``degraded" with large unrolled stage number instead of ``overfitted" as the training loss and test PSNR both become worse. It has been previously discovered and studied in \cite{he2016deep}.}. To get a deeper insight into this phenomenon, we fix the kernel weight and bias of the last Conv layer in each stage to $\left[1,0,\cdots,0\right]^\top$ and 0, respectively, to keep only one channel of the enhanced feature from RBs and discard others to obtain a weaker variant \underline{``fixed"}.  Fig.~\ref{fig:Train400}(b) shows that this partially fixed version achieves similar or even better recovery accuracy compared with the baseline when $K\in \left\{10,30\right\}$, thus exposing the inefficient channel reduction and expansion operations of each unrolled baseline stage. Then we eliminate these channel shrinkage-expansion operations as well as the intra-stage skip connections and introduce a container variable $\mathbf{S}^{(k)}\in \mathbb{R}^{N\times (C-1)}$ that conveys feature-level contents between every two adjacent stages and is zero-initialized to obtain another variant \underline{``reduced"}, as Fig.~\ref{fig:stage_PRL}(b) shows. Fig.~\ref{fig:Train400}(c) exhibits the large PSNR gains and training curves with smaller fluctuations brought by the inter-stage transmission maximization. This is also mentioned in \cite{song2021memory} but not well-studied yet, remaining the serious degradation that still occurs in the reduced version. But we find that this can be eliminated by activating sufficient feature-level physics knowledge injections while keeping the network construction concise and interpretable.

\textbf{Resolve degradation via feature-domain unrolling.} Motivated by the successful applications of high-dimensional image restoration of BM3D \cite{dabov2007image}, group-based sparse representation \cite{zhang2014group}, the recent adaptive consistency prior \cite{ren2021adaptive} and convolutional dictionary learning \cite{zheng2021deep}, we propose to learn the recovery in the feature domain (FD). Specifically, from the perspective of optimization, we first use a data-driven extractor $\mathcal{H}_\text{ext}:\mathbb{R}^N\rightarrow \mathbb{R}^{N\times C}$ to obtain an initial high-capacity representation $\hat{\mathbf{X}}^{(0)}=\mathcal{H}_\text{ext}(\mathbf{A}^\top \mathbf{y})$ and then extend Eq.~(\ref{eq:opt_image}) to the following matrix-optimization problem:
\begin{equation}
	\setlength{\abovedisplayskip}{2pt}
	\setlength{\belowdisplayskip}{3pt}
	\label{eq:opt_feature}
	\hat{\mathbf{X}}=\underset{\mathbf{X}}{\arg\min} \mathcal{F}\left( \mathbf{X};\mathbf{A},\mathbf{y} \right) +\lambda \mathcal{G}\left( \mathbf{X} \right),
\end{equation}
where the physical constraint $\mathcal{F}$ and the regularizer $\mathcal{G}$ are both data-driven learned. The Eq.~(\ref{eq:opt_feature}) is assumed to be convex and solved by the following PGD steps in the matrix form:
\begin{equation}
	\setlength{\abovedisplayskip}{3pt}
	\setlength{\belowdisplayskip}{2pt}
	\label{eq:pgdf1}
	\mathbf{Z}^{\left( k \right)}=\hat{\mathbf{X}}^{\left( k-1 \right)}-\rho \mathcal{H}_\text{grad}^{(k)}\left( \hat{\mathbf{X}}^{\left( k-1 \right)};\mathbf{A,y} \right),
	\vspace{-1pt}
\end{equation}
\begin{equation}
	\setlength{\abovedisplayskip}{2pt}
	\setlength{\belowdisplayskip}{0pt}
	\label{eq:pgdf2}
	\hat{\mathbf{X}}^{\left( k \right)}=\mathcal{H}_\text{prox}^{(k)}\left( \mathbf{Z}^{\left( k \right)} \right).
	\vspace{-2pt}
\end{equation}
Here we use two neural operators $\mathcal{H}_\text{grad}^{(k)}$ and $\mathcal{H}_\text{prox}^{(k)}$ to estimate the gradient $\nabla \mathcal{F}$ and perform the proximal mapping for each iteration, and obtain the final result by a reconstructor $\mathcal{H}_\text{rec}:\mathbb{R}^{N\times C}\rightarrow \mathbb{R}^N$ with $\hat{\mathbf{x}}=\mathcal{H}_\text{rec}(\hat{\mathbf{X}}^{(K)})$ after $K$ iterations.

{From another perspective of range-nullspace decomposition, motivated by \cite{chen2020deep}, we consider the signal RND: $[\mathbf{x}=\mathbf{A}^{\top}(\mathbf{y}-\mathbf{\epsilon})+\left( \mathbf{I}_N-\mathbf{A}^{\top}\mathbf{A} \right) \mathbf{x}]$ and propose the following iterative learnable steps to progressively reconstruct the range space component and nullspace component in ID alternately:
\begin{equation}
	\setlength{\abovedisplayskip}{3pt}
	\setlength{\belowdisplayskip}{3pt}
	\label{eq:rnd1}
	\mathbf{r}^{\left( k \right)}=\mathbf{A}^\top (\mathbf{y}-\mathbf{A}h_\mathcal{R}(\hat{\mathbf{x}}^{\left( k-1 \right)})),
\end{equation}
\begin{equation}
	\setlength{\abovedisplayskip}{3pt}
	\setlength{\belowdisplayskip}{3pt}
	\label{eq:rnd2}
	\hat{\mathbf{x}}^{\left( k \right)}=\mathbf{r}^{\left( k \right)}+\left( \mathbf{I}_N-\mathbf{A}^{\top}\mathbf{A} \right)h_\mathcal{N}(\mathbf{r}^{\left( k \right)}),
\end{equation}
where the first step is to suppress the observation noise by a range space noise estimator $h_\mathcal{R}:\mathbb{R}^N\rightarrow \mathbb{R}^N$ and the second step recovers the lost component by a nullspace reconstructor $h_\mathcal{N}:\mathbb{R}^N\rightarrow \mathbb{R}^N$. Similarly, we generalize them to FD as follows:
\begin{equation}
	\setlength{\abovedisplayskip}{3pt}
	\setlength{\belowdisplayskip}{3pt}
	\label{eq:rnd_feature1}
	\mathbf{R}^{\left( k \right)}=\mathcal{H}^{(k)}_\mathcal{R}(\hat{\mathbf{X}}^{\left( k-1 \right)};\mathbf{A},\mathbf{y}),
\end{equation}
\begin{equation}
	\setlength{\abovedisplayskip}{3pt}
	\setlength{\belowdisplayskip}{3pt}
	\label{eq:rnd_feature2}
	\hat{\mathbf{X}}^{\left( k \right)}=\mathcal{H}^{(k)}_\mathcal{N}(\hat{\mathbf{X}}^{\left( k-1 \right)},\mathbf{R}^{\left( k \right)};\mathbf{A}),
\end{equation}
where $\mathcal{H}^{(k)}_\mathcal{R/N}$ conduct the range-/null-space processings in the adaptive learned FD, respectively.}

\textbf{Relationship between iterative PGD and RND steps.} We highlight that our introduced PGD and RND iterations are two independent CS solutions, in which their ID unrollings have both been validated to be successful in inverse imaging tasks \cite{zhang2018ista,chen2020deep}. The main difference between them is that their two iterative steps have different explanations and purposes. And what they have in common is that they both iteratively use the physics information $\{\mathbf{A},\mathbf{y}\}$ for reconstruction. We will show later that both PGD and RND are two appropriate frameworks that can be applied to guide the design of PRL network modules and achieve excellent CS imaging performance.

\begin{table*}[!t]
\caption{Comparison among our provided five PRL network variants with their default setting. Following most existing deep unrolled approaches, all parameters of each PRL network are indiscriminately learned by $\ell_2$ loss end-to-end. Here, all PGD-/RND-unrolled stages $\mathcal{H}_\text{Stage-PGD/-RND}^{(k)}$ have the same learnable components of two Conv layers and two RBs. $i$ and $k$ denote the indices of group and stage, respectively. $\hat{\mathbf{x}}$ is the sampled and reconstructed result of ground truth image $\mathbf{x}$.}
\label{tab:PRL_network_summarization}
\centering
\resizebox{1.0\textwidth}{!}{
\begin{tabular}{l|ccccc}
\shline
\rowcolor[HTML]{EFEFEF} 
Our Proposed PRL Network &
  \multicolumn{1}{c|}{\cellcolor[HTML]{EFEFEF}PRL-PGD-S} &
  \multicolumn{1}{c|}{\cellcolor[HTML]{EFEFEF}PRL-PGD} &
  \multicolumn{1}{c|}{\cellcolor[HTML]{EFEFEF}PRL-PGD$^+$} &
  \multicolumn{1}{c|}{\cellcolor[HTML]{EFEFEF}PRL-RND} &
  PRL-RND$^+$ \\ \hline
  Deep Unrolling Type & \multicolumn{3}{c|}{PGD-unrolling} & \multicolumn{2}{c}{RND-unrolling} \\ \hline
Corresponding Model Equation in ID                      & \multicolumn{3}{c|}{$\hat{\mathbf{x}}=\underset{\mathbf{x}}{\arg\min}f\left( \mathbf{x};\mathbf{A},\mathbf{y} \right) +\lambda g\left( \mathbf{x} \right)$~(\ref{eq:opt_image})}                                                & \multicolumn{2}{c}{$\mathbf{x}={\mathbf{A}^{\top}\mathbf{Ax}}+{\left( \mathbf{I}_N-\mathbf{A}^{\top}\mathbf{A} \right) \mathbf{x}}$~(\ref{eq:rnd})}             \\ \hline
\multirow{2}{*}{Corresponding Iterative Steps in ID}    & \multicolumn{3}{c|}{$\mathbf{z}^{\left( k \right)}=\hat{\mathbf{x}}^{\left( k-1 \right)}-\rho \nabla f\left( \hat{\mathbf{x}}^{\left( k-1 \right)};\mathbf{A,y} \right)$~(\ref{eq:pgd1})}                                                & \multicolumn{2}{c}{$\mathbf{r}^{\left( k \right)}=\mathbf{A}^\top (\mathbf{y}-\mathbf{A}h_\mathcal{R}(\hat{\mathbf{x}}^{\left( k-1 \right)}))$~(\ref{eq:rnd1})}             \\  & \multicolumn{3}{c|}{$\hat{\mathbf{x}}^{\left( k \right)}=\mathtt{prox}_{\lambda g}\left( \mathbf{z}^{\left( k \right)} \right)$~(\ref{eq:pgd2})} & \multicolumn{2}{c}{$\hat{\mathbf{x}}^{\left( k \right)}={\mathbf{r}}^{\left( k \right)}+\left( \mathbf{I}_N-\mathbf{A}^{\top}\mathbf{A} \right)h_\mathcal{N}(\mathbf{r}^{\left( k \right)})$~(\ref{eq:rnd2})} \\ \hline \multirow{2}{*}{Generalized Iterative Steps in FD}  & \multicolumn{3}{c|}{$\mathbf{Z}^{\left( k \right)}=\hat{\mathbf{X}}^{\left( k-1 \right)}-\rho \mathcal{H}_\text{grad}^{(k)}\left( \hat{\mathbf{X}}^{\left( k-1 \right)};\mathbf{A,y} \right)$~(\ref{eq:pgdf1})} & \multicolumn{2}{c}{$\mathbf{R}^{\left( k \right)}=\mathcal{H}^{(k)}_\mathcal{R}(\hat{\mathbf{X}}^{\left( k-1 \right)};\mathbf{A},\mathbf{y})$~(\ref{eq:rnd_feature1})} \\ & \multicolumn{3}{c|}{$\hat{\mathbf{X}}^{\left( k \right)}=\mathcal{H}_\text{prox}^{(k)}\left( \mathbf{Z}^{\left( k \right)} \right)$~(\ref{eq:pgdf2})} & \multicolumn{2}{c}{$\hat{\mathbf{X}}^{\left( k \right)}=\mathcal{H}^{(k)}_\mathcal{N}(\hat{\mathbf{X}}^{\left( k-1 \right)},\mathbf{R}^{\left( k \right)};\mathbf{A})$~(\ref{eq:rnd_feature2})} \\
\hline Recovery Network Architecture & \multicolumn{5}{c}{Our proposed U-shaped architecture in FD, illustrated in Fig.~\ref{fig:arch} (c)} \\ \hline Unrolled Stage Structure & \multicolumn{3}{c|}{$\mathcal{H}_\text{Stage-PGD}^{(k)}$ illustrated in Fig.~\ref{fig:stage_PRL} (c)} & \multicolumn{2}{c}{$\mathcal{H}_\text{Stage-RND}^{(k)}$ illustrated in Fig.~\ref{fig:stage_PRL} (d)} \\ \hline
\multirow{2}{*}{Configuration of Hyper-parameters} & \multicolumn{5}{c}{$B\equiv 32$, $\rho\equiv 1$} \\ \hhline{~-----} & \multicolumn{1}{c|}{$C=D=8$, $K=5$} & \multicolumn{1}{c|}{$C=D=8$, $K=5$} & \multicolumn{1}{c|}{$C=D=16$, $K=7$} & \multicolumn{1}{c|}{$C=D=8$, $K=5$} & \multicolumn{1}{c}{$C=D=16$, $K=7$} \\ \hline
\multirow{2}{*}{Set of All Learnable Parameters} & \multicolumn{5}{c}{$\mathbf{\Theta}=\left\{\mathbf{A},\mathcal{H}_\text{ext},\mathcal{H}_\text{rec},\text{SConv}_1,\text{SConv}_2,\text{TConv}_1,\text{TConv}_1\right\}\cup\left[\bigcup_{i\in\{1,\cdots,6\},k\in\{1,\cdots,K\}}{\mathcal{H}_\text{Stage-PGD/-RND}^{(i,k)}}\right]$} \\  & \multicolumn{5}{c}{$\mathcal{H}_\text{Stage-PGD/-RND}^{(i,k)}=\left\{\mathcal{H}_\text{grad/merge}^{(i,k)},\text{Conv}_1^{(i,k)},\text{Conv}_2^{(i,k)},\text{RB}_1^{(i,k)},\text{RB}_2^{(i,k)}\right\}$} \\ \hline
Parameter Sharing Strategy& \multicolumn{1}{c|}{\begin{tabular}[c]{@{}l@{}}Intra-group stage\\parameter sharing\end{tabular}} & \multicolumn{4}{c}{No parameter sharing}\\ \hline Loss Function& \multicolumn{5}{c}{End-to-end $\ell_2$ loss (\textit{i.e.}, the mean squared error) $\mathcal{L}=\frac{1}{N}\lVert \mathbf{x}-\hat{\mathbf{x}} \rVert_2^2$ with training image patch $\mathbf{x}\in\mathbb{R}^N$} \\ \shline
\end{tabular}}
\end{table*}

\subsection{Multiscale High-throughput Unrolled Recovery Network}
Intuitively, the above feature-level reconstruction can be mapped to a more powerful architecture than the ID counterpart in Fig.~\ref{fig:arch}(b), where the image is refined and transmitted among unrolled stages in a high-dimensional FD, but we observe from \cite{sun2016deep,zhang2018ista,zhang2020optimization,zhang2020amp,you2021coast,you2021ista,song2021memory,wu2021dense} that the recovery ability of plain frameworks is generally hard to be further improved through increasing whether the stage number $K$ or the feature capacity with $C$ due to their unacceptable surge of inference time and performance saturations, thus stimulating us to build more efficient networks with simple basic modules and generalized operators as follows.

\textbf{U-shaped architecture for context perception and recovery acceleration.} To combine the merits of deep unrolling and multiscale recovery networks \cite{zamir2021multi,zhang2021plug}, we propose a 3-scale U-shaped backbone in Fig.~\ref{fig:arch}(c), in which the last two scales are equipped with skip connections. The feature channel numbers from the first scale to the third are set to $D$, $4D$ and $16D$ with $\times 1$, $\times 2$ and $\times 4$ spatial downscaling respectively for keeping a stationary capacity. There are two groups for each scale. Each group consists of $K$ iterative stages. The $2\times 2$ strided convolution (SConv) and $2\times 2$ transposed convolution (TConv) are adopted as the downscaling and upscaling operators. The FD extractor $\mathcal{H}_\text{ext}$ and ID reconstructor $\mathcal{H}_\text{rec}$ are both simply implemented by a Conv layer.

\textbf{Scalable stage structures for multiscale FD unrolling.} Based on the above two generalized PGD-/RND-solutions, we develop two scalable high-throughput stage structures, denoted by $\mathcal{H}_\text{Stage-PGD}^{(k)}$ and $\mathcal{H}_\text{Stage-RND}^{(k)}$, respectively, to recover the deep image features in different scales. \textbf{\textcolor{blue}{(1)}} \textbf{\textcolor{purple}{PGD-Unrolling:}} For the $k$-th stage of a group with $\times r$ downscaling, as Fig.~\ref{fig:stage_PRL}(c) illustrates, we map the PGD steps into two serial parts. The first part corresponding to Eq.~(\ref{eq:pgdf1}) takes the $r^2D$-channel $\hat{\mathbf{X}}^{(k-1)}$ as the input and perform the gradient descent to obtain $\mathbf{Z}^{(k)}$ through the FD gradient estimation from $\mathcal{H}_\text{grad}^{(k)}$. The second part corresponding to Eq.~(\ref{eq:pgdf2}) contains a proximal mapping operator $\mathcal{H}_\text{prox}^{(k)}$, in which a network similar to Fig.~\ref{fig:stage_PRL}(a) including a Conv layer, two RBs, and another Conv layer is learned to estimate the FD residual. \textbf{\textcolor{blue}{(2)}} \textbf{\textcolor{purple}{RND-Unrolling:}} In Fig.~\ref{fig:stage_PRL}(d), we disentangle a reconstruction step into a range space recovery part and a nullspace recovery part. The first part $\mathcal{H}_\mathcal{R}^{(k)}$ corresponding to Eq.~(\ref{eq:rnd_feature1}) first embeds $\hat{\mathbf{X}}^{(k-1)}$ into a $r^2C$-dimensional space by a Conv layer, then refines through an RB and supplement $r^2D$ channels of the enhanced feature with $\mathbf{A}$ and $\mathbf{y}$ in $\mathcal{R}(\mathbf{A}^\top)$ by a network $\mathcal{H}_\text{merge}^{(k)}$ to obtain $\mathbf{R}^{(k)}$. The second part $\mathcal{H}_\mathcal{N}^{(k)}$ processes $\mathbf{R}^{(k)}$ with a RB, projects $r^2D$ feature channels into $\mathcal{N}(\mathbf{A})$ and get the FD residual from a Conv layer.

\textbf{Generalized physical operators in FD.} To keep the simplicity and a same parameter number for PGD-/RND-networks, we design a lightweight physical fusion module for $\mathcal{H}_\text{grad}^{(k)}$ and $\mathcal{H}_\text{merge}^{(k)}$. As Fig.~\ref{fig:stage_PRL}(e) exhibits, they share the same pipeline that projects the input $\mathbf{X}$ into $\mathcal{R}(\mathbf{A}^\top)$ and fuse $\mathbf{X}$, $\mathbf{A^\top AX}$ and the upsampled $\mathbf{A^\top y}$ by PixelShuffle \cite{shi2016real}, concatenation and a Conv layer with Sigmoid activation $\Sigma (\cdot)$ to get $\mathbf{X}_\text{Fus}$. To implement the physical forward operator and its transpose in FD, we use PixelShuffle with a $B\times B$ strided group convolution (SGConv), and a $B\times B$ transposed group convolution (TGConv) with PixelUnshuffle and share all the group Conv weights with the sampling matrix $\mathbf{A}$, as Fig.~\ref{fig:stage_PRL}(f) and (g) show.

As we will demonstrate in Sec.~\ref{sec:ex}, our multiscale unrolling generalization in FD can achieve faster inferences compared with other ID methods by breaking the original task into three correlated subproblems with different spatial levels. It overcomes the degradation by sufficient physics utilizations which is helpful for network training and convergence of the iterative reconstruction. This generalization also brings a large room for further improvement of recovery performance.

\begin{table*}[!t]
	\caption{Quantitative comparison of PSNR(dB)/SSIM among different CS networks on Set11 \cite{kulkarni2016reconnet} and Urban100 \cite{huang2015single}. The traditional physics-free, existing physics-engaged, and our proposed networks are highlighted in the yellow, green and cyan backgrounds. The best and second-best results are marked in red and blue, respectively.}
	\label{tab:comp1}
	\centering
	\resizebox{1.0\textwidth}{!}{
		\begin{tabular}{l|ccccc|ccccc|c|c}
			\shline
			Dataset                                               & \multicolumn{5}{c|}{Set11}                                                                                              & \multicolumn{5}{c|}{Urban100}                                                                                             & Time (ms) & \#Param.                                         \\ \hhline{-----------~~}
			CS Ratio $\gamma$                                     & 10\%    & 20\%                                        & 30\%    & 40\%                                        & 50\%    & 10\%    & 20\%                                        & 30\%    & 40\%                                        & 50\%    & /GFLOPs   & (M)                                              \\ \hline
			\rowcolor[HTML]{FFFFC7} 
			CSNet$^+$ \cite{shi2019image}                                            & 28.34   & 31.66                                       & 34.30   & 36.48                                       & 38.52   & 23.96   & 26.95                                       & 29.12   & 30.98                                       & 32.76   & \textcolor{red}{16.767}    & \cellcolor[HTML]{FFFFC7}                         \\
			\rowcolor[HTML]{FFFFC7} 
			(TIP'19)                                              & /0.8580 & /0.9203                                     & /0.9490 & /0.9644                                     & /0.9749 & /0.7309 & /0.8449                                     & /0.8974 & /0.9273                                     & /0.9484 & /53.932   & \multirow{-2}{*}{\cellcolor[HTML]{FFFFC7}\textcolor{red}{0.621}}  \\ \hline
			\rowcolor[HTML]{FFFFC7} 
			SCSNet \cite{shi2019scalable}                                               & 28.52   & 31.82                                       & 34.64   & 36.92                                       & 39.01   & 24.22   & 27.09                                       & 29.41   & 31.38                                       & 33.31   & 30.905    & \cellcolor[HTML]{FFFFC7}                         \\
			\rowcolor[HTML]{FFFFC7} 
			(CVPR'19)                                             & /0.8616 & /0.9215                                     & /0.9511 & /0.9666                                     & /0.9769 & /0.7394 & /0.8485                                     & /0.9016 & /0.9321                                     & /0.9534 & /77.312   & \multirow{-2}{*}{\cellcolor[HTML]{FFFFC7}0.801}  \\ \hline
			\rowcolor[HTML]{E8FFE8} 
			OPINE-Net$^+$ \cite{zhang2020optimization}                                        & 29.81   & 33.43                                       & 35.99   & 38.24                                       & 40.19   & 25.90   & 29.38 & 31.97                                       & 34.27   & 36.28                                       &  \textcolor{blue}{17.305}    & \cellcolor[HTML]{E8FFE8}                         \\
			\rowcolor[HTML]{E8FFE8} 
			(JSTSP'20)                                            & /0.8904 & /0.9392                                     & /0.9596 & /0.9718                                     & /0.9800 & /0.7979 & /0.8902                                     & /0.9309 & /0.9548                                     & /0.9697 & /\textcolor{red}{36.294}   & \multirow{-2}{*}{\cellcolor[HTML]{E8FFE8}\textcolor{blue}{0.622}}  \\ \hline
			\rowcolor[HTML]{E8FFE8} 
			AMP-Net \cite{zhang2020amp}                                              & 29.40   & 33.33                                       & 36.03   & 38.28                                       & 40.34   & 25.32   & 29.01                                       & 31.63   & 33.88                                       & 35.91   & 27.378    & \cellcolor[HTML]{E8FFE8}                         \\
			\rowcolor[HTML]{E8FFE8} 
			(TIP'21)                                              & /0.8779 & /0.9345                                     & /0.9586 & /0.9715                                     & /0.9804 & /0.7747 & /0.8799                                     & /0.9248 & /0.9511                                     & /0.9673 & /\textcolor{blue}{47.934}   & \multirow{-2}{*}{\cellcolor[HTML]{E8FFE8}0.868}  \\ \hline
			\rowcolor[HTML]{E8FFE8} 
			COAST \cite{you2021coast}                                                & 30.02   & \cellcolor[HTML]{E8FFE8}                    & 36.33   & \cellcolor[HTML]{E8FFE8}                    & 40.33   & 26.17   & \cellcolor[HTML]{E8FFE8}                    & 32.48   & \cellcolor[HTML]{E8FFE8}                    & 36.56   & 45.544    & \cellcolor[HTML]{E8FFE8}                         \\
			\rowcolor[HTML]{E8FFE8} 
			(TIP'21)                                              & /0.8946 & \multirow{-2}{*}{\cellcolor[HTML]{E8FFE8}-} & /0.9618 & \multirow{-2}{*}{\cellcolor[HTML]{E8FFE8}-} & /0.9804 & /0.8070 & \multirow{-2}{*}{\cellcolor[HTML]{E8FFE8}-} & /0.9362 & \multirow{-2}{*}{\cellcolor[HTML]{E8FFE8}-} & /0.9709 & /156.658  & \multirow{-2}{*}{\cellcolor[HTML]{E8FFE8}1.122}  \\ \hline
			\rowcolor[HTML]{E8FFE8} 
			MADUN \cite{song2021memory}                                                & 29.89   & 34.09                                       & 36.90   & 39.14                                       & 40.75   & 26.23   & 30.24                                       & 33.00   & 35.10                                       & 36.69   & 92.146    & \cellcolor[HTML]{E8FFE8}                         \\
			\rowcolor[HTML]{E8FFE8} 
			(ACM MM'21)                                           & /0.8982 & /0.9478                                     & /0.9671 & /0.9769                                     & /0.9831 & /0.8250 & /0.9108                                     & /0.9457 & /0.9639                                     & /0.9746 & /390.032  & \multirow{-2}{*}{\cellcolor[HTML]{E8FFE8}3.126}  \\ \hline

			\rowcolor[HTML]{CCFFFD} 
			\cellcolor[HTML]{CCFFFD}                              & 31.40   & 35.07                                       & 37.55   & 39.60                                       & 41.47   & 28.22   & 31.71                                       & 34.12   & 36.14                                       & 38.09   & 36.873    & \cellcolor[HTML]{CCFFFD}                         \\
			\rowcolor[HTML]{CCFFFD} 
			\multirow{-2}{*}{\cellcolor[HTML]{CCFFFD}\textbf{PRL-PGD-S}}     & /0.9158 & /0.9533                                     & /0.9688 & /0.9782                                     & /0.9838 & /0.8657 & /0.9280                                     & /0.9541 & /0.9678                                     & /0.9780 & /101.778  & \multirow{-2}{*}{\cellcolor[HTML]{CCFFFD}2.558} \\ \hline

			\rowcolor[HTML]{CCFFFD} 
			\cellcolor[HTML]{CCFFFD}                              & 31.49   & 35.17                                       & 37.63   & 39.67                                       & 41.52   & 28.34   & 31.82                                       & 34.21   & 36.19                                       & 38.12   & 38.485    & \cellcolor[HTML]{CCFFFD}                         \\
			\rowcolor[HTML]{CCFFFD} 
			\multirow{-2}{*}{\cellcolor[HTML]{CCFFFD}\textbf{PRL-PGD}}     & /0.9173 & /0.9546                                     & /0.9697 & /0.9784                                     & /0.9842 & /0.8669 & /0.9294                                     & /0.9547 & /0.9688                                     & /0.9783 & /101.778  & \multirow{-2}{*}{\cellcolor[HTML]{CCFFFD}10.173} \\ \hline
			\rowcolor[HTML]{CCFFFD} 
			\cellcolor[HTML]{CCFFFD}                              & \textcolor{red}{31.70}   & \textcolor{blue}{35.37}                                       & \textcolor{blue}{37.89}   & \textcolor{blue}{39.89}                                       & \textcolor{blue}{41.78}   & \textcolor{red}{28.77}   & \textcolor{blue}{32.33}                                       & \textcolor{blue}{34.73}   & \textcolor{blue}{36.71}                                       & \textcolor{blue}{38.57}   & 69.567    & \cellcolor[HTML]{CCFFFD}                         \\
			\rowcolor[HTML]{CCFFFD} 
			\multirow{-2}{*}{\cellcolor[HTML]{CCFFFD}\textbf{PRL-PGD$^+$}} & /\textcolor{red}{0.9193} & /\textcolor{blue}{0.9561}                                     & /\textcolor{blue}{0.9709} & /\textcolor{blue}{0.9789}                                     & /\textcolor{blue}{0.9847}  & /\textcolor{red}{0.8746} & /\textcolor{blue}{0.9343}                                     & /\textcolor{blue}{0.9576} & /\textcolor{blue}{0.9707}                                     & /\textcolor{blue}{0.9794} & /562.186  & \multirow{-2}{*}{\cellcolor[HTML]{CCFFFD}55.621} \\ \hline
			\rowcolor[HTML]{CCFFFD} 
			\cellcolor[HTML]{CCFFFD}                              & 31.46   & 35.15                                       & 37.67   & 39.71                                       & 41.63   & 28.32   & 31.87                                       & 34.24   & 36.26                                       & 38.24   & 52.116    & \cellcolor[HTML]{CCFFFD}                         \\
			\rowcolor[HTML]{CCFFFD} 
			\multirow{-2}{*}{\cellcolor[HTML]{CCFFFD}\textbf{PRL-RND}}     & /0.9173 & /0.9548                                     & /0.9701 & /0.9787                                     & /0.9844 & /0.8667 & /0.9297                                     & /0.9548 & /0.9691                                     & /0.9789 & /101.778  & \multirow{-2}{*}{\cellcolor[HTML]{CCFFFD}10.173} \\ \hline
			\rowcolor[HTML]{CCFFFD} 
			\cellcolor[HTML]{CCFFFD}                              & \textcolor{blue}{31.66}   & \textcolor{red}{35.42}                                       & \textcolor{red}{37.96}   & \textcolor{red}{39.93}                                       & \textcolor{red}{41.84}   & \textcolor{blue}{28.75}   & \textcolor{red}{32.36}                                       & \textcolor{red}{34.78}   & \textcolor{red}{36.77}                                       & \textcolor{red}{38.64}   & 90.949    & \cellcolor[HTML]{CCFFFD}                         \\
			\rowcolor[HTML]{CCFFFD} 
			\multirow{-2}{*}{\cellcolor[HTML]{CCFFFD}\textbf{PRL-RND$^+$}} & /\textcolor{blue}{0.9189} & /\textcolor{red}{0.9567}                                     & /\textcolor{red}{0.9712} & /\textcolor{red}{0.9790}                                     & /\textcolor{red}{0.9850} & /\textcolor{blue}{0.8743} & /\textcolor{red}{0.9344}                                     & /\textcolor{red}{0.9580} & /\textcolor{red}{0.9710}                                     & /\textcolor{red}{0.9798} & /562.186  & \multirow{-2}{*}{\cellcolor[HTML]{CCFFFD}55.621} \\ \shline
	\end{tabular}}
\end{table*}

\begin{table*}[!t]
\caption{Evaluations of the inference time (ms) and computational cost (GFLOPs) of our PRL-PGD on a 1080Ti GPU with $\gamma =50\%$. The inference time and computational cost in milliseconds and FLOPs approximately follow the formulations $(0.0005N+5.7791)$ and $(1553009N+840)$, respectively, where $N$ is the total number of image pixels.}
\label{tab:computation_complexity}
\centering
\scalebox{0.875}{\begin{tabular}{
>{\columncolor[HTML]{EFEFEF}}c |c|c|c|c|c}
\shline
Input Image Size $N$                        & $64\times 64$ & $128\times 128$ & $256\times 256$ & $512\times 512$ & $1024\times 1024$ \\ \hline
Inference Time (ms)               & 4.691         & 11.534          & 38.485          & 128.126         & 483.564           \\ \hline
Computational Cost (GFLOPs) & 6.361         & 25.445          & 101.778         & 407.112         & 1628.448          \\ \shline
\end{tabular}}
\end{table*}

\subsection{Relationship to Other Works}
{Traditional physics-free CS recovery networks like ReconNet \cite{kulkarni2016reconnet} achieve fast feature-level inferences but lack explicit guidance from the sampling physics supplement of the forward operator $\mathbf{A}$ and measurements $\mathbf{y}$. As we will show later, physics guidance plays an essential role in easing training and improving recovery accuracy. Most physics-engaged CS methods, such as ISTA-Net$^+$ \cite{zhang2018ista} and ISTA-Net$^{++}$ \cite{you2021ista}, based on the single-scale structure in Fig.~\ref{fig:stage_PRL}(a) and physical operators in ID, are special cases of our PRL implementations with less flexibility and capacity due to their fixed and non-adaptive $\ell_2$ data fidelity term and the trivial configuration of $r=D\equiv 1$.}

{Furthermore, the recent MPRNet \cite{zamir2021multi} performs restorations progressively based on the contextualized feature from U-shaped encoder-decoders and details from an original-resolution sub-network with inter-branch communications. However, it lacks mathematical insights and interpretability as it does not consider the physical observation process. The powerful denoisers DRCAN \cite{chen2021deep} and DRUNet \cite{zhang2021plug} adopt efficient proximal mapping structures but are designed for plug-and-play ID recoveries. The DeamNet \cite{ren2021adaptive} and DCDicL \cite{zheng2021deep} combine the merits of both FD and U-shaped recoveries but are limited by the plain architecture in Fig.~\ref{fig:arch}(b) and the special denoising tasks with $\mathbf{A}=\mathbf{I}_N$. Compared to these works, PRL is more flexible and generalizable to various sampling (or degradation) models and different reconstruction frameworks.}

{While our PRL and the state-of-the-art memory augmentation-based CS networks, including MAC-Net \cite{chen2020learning}, MADUN \cite{song2021memory}, and MAPUN \cite{song2023deep}, may share some similarities, such as the use of maximized information flow to overcome the information bottleneck in traditional deep unrolled networks, and the exploitation of adaptive network components to fuse image features, they differ in several important aspects. Firstly, our PRL resolves the difficulty of unrolled network training by achieving sufficient physics-data fusion in FD without degradation, which may not be addressed by them since they are limited to ID recovery, as our pilot experiment exhibits. Secondly, the architectural unrolling generalization of PRL from a strict repetitive plain form to a more flexible and powerful U-shaped one, in which the structures and hyperparameters of different stages or network blocks can be different and not strictly forced to be the same. This generalization by PRL is the key to breaking performance saturation by multiscale image context perception with a large effective receptive field, reducing inference time cost with similar network capacity, obtaining significant room for improvement, bridging the gap between deep unrolling and generic recovery networks, and providing a viable solution for deep physics-guided generalized unrolling network design. In a nutshell, our PRL method can be essentially viewed as a significant extension of the existing deep unrolling paradigm, from the ID plain iterative unrolling to the FD U-shaped flexible unrolling. Our experiments will provide comparisons and an understanding of these strategies.}

\section{Experiments}
\label{sec:ex}
\subsection{Implementation Details}
Following \cite{you2021ista,zheng2021deep,zhang2021plug}, we employ a large training set consisting of the T91 \cite{yang2008image,dong2014learning}, Train400 \cite{chen2016trainable,zhang2017beyond}, 800 images from the DIV2K \cite{agustsson2017ntire} training set, and 4744 images from the Waterloo exploration database \cite{ma2016waterloo}. Four benchmarks: Set11 \cite{kulkarni2016reconnet}, CBSD68 \cite{martin2001database}  Urban100 \cite{huang2015single} and the DIV2K validation set are used for evaluation. The test images from Urban100 and DIV2K are 256$\times$256 center-cropped. All results are evaluated with the Peak Signal-to-Noise Ratio (PSNR) and Structural Similarity Index Measure (SSIM) \cite{wang2004image} on the Y channel (\textit{i.e.} the luminance component).

As described in Sec.~\ref{sec:DPRL}, we construct two PRL networks: \textbf{PRL-PGD} and \textbf{PRL-RND} based on the generalized PGD and RND with $B\equiv 32$, $C=D=8$, $K=5$ and $\rho \equiv 1$ which are only different in the stage structure (Fig.~\ref{fig:stage_PRL}(c) for PRL-PGD and (d) for PRL-RND) and extended by $C=D=16$ and $K=7$ to become \textbf{PRL-PGD$^+$} and \textbf{PRL-RND$^+$}. A PRL-PGD variant: \textbf{PRL-PGD-S} is developed by sharing the parameters across all the stages in each group to keep the original structure and largely reduce the total parameter number. All PRL networks are implemented in PyTorch \cite{paszke2019pytorch} and trained end-to-end by $\ell_2$ loss (\textit{i.e.} the MSE loss) and Adam \cite{kingma2014adam} with a momentum of 0.9 and a weight decay of 0.999. Mini-batches of 16 randomly cropped 128$\times$128 patches and data augmentations with eight geometric transforms including rotations, flippings and their combinations \cite{tian2020attention} are utilized for training. It takes about three days to learn a PRL-PGD on an NVIDIA Tesla V100 GPU with $4.8\times 10^5$ steps. The learning rate starts from $1\times 10^{-4}$ and decreases by 0.1 after $3.2\times 10^5$, $4\times 10^5$ and $4.4\times 10^5$ iterations, respectively. A brief summary of our default settings for PRL networks is provided in Tab.~\ref{tab:PRL_network_summarization}.

\begin{table*}
	\caption{PSNR (dB) results with the \textcolor{blue}{vertical} and \textcolor{purple}{horizontal} increments on Set11 \cite{kulkarni2016reconnet} of two PRL-PGD$^+$ settings with various physics domain dimensionalities in the case of $\gamma=10\%$.}
	\label{tab:abla_q}
	\centering
	\scalebox{0.95}{\begin{tabular}{c|cccccc|c}
		\shline
		\multirow{2}{*}{Setting} & \multicolumn{7}{c}{Physical Feature Dimensionality $q$}               \\ \hhline{~-------} 
		& 0     & 1     & 4     & 7     & 10    & 13    & 16 (default)                     \\ \hline
		$D=q$                                                                              & -     & 31.25 & 31.50$_{\textcolor{purple}{+0.25}}$ & 31.58$_{\textcolor{purple}{+0.08}}$ & 31.61$_{\textcolor{purple}{+0.03}}$ & 31.63$_{\textcolor{purple}{+0.02}}$ & \multirow{2}{*}{31.70$_{\textcolor{purple}{+0.02}}^{\textcolor{purple}{+0.07}}$} \\ \hhline{-------~}
		$D=16$                                                                             & 30.67 & 31.54$_{\textcolor{purple}{+0.87}}^{\textcolor{blue}{+0.29}}$ & 31.56$_{\textcolor{purple}{+0.02}}^{\textcolor{blue}{+0.06}}$ & 31.62$_{\textcolor{purple}{+0.06}}^{\textcolor{blue}{+0.04}}$ & 31.65$_{\textcolor{purple}{+0.03}}^{\textcolor{blue}{+0.04}}$ & 31.68$_{\textcolor{purple}{+0.03}}^{\textcolor{blue}{+0.05}}$ &                        \\ \shline
	\end{tabular}}
\end{table*}

\begin{table*}
	\caption{PSNR (dB) results with the \textcolor{red}{vertical} increments on Set11 \cite{kulkarni2016reconnet} and the parameter number (M) of two existing deep unrolled networks and their corresponding enhanced FD-unrolling network variants.}
	\label{tab:boost_existing_duns}
	\centering
	\scalebox{0.9}{\begin{tabular}{c|ccc}
		\shline
		\multirow{2}{*}{Method} & \multicolumn{3}{c}{CS Ratio $\gamma$}  \\ \hhline{~---} 
		& 10\%         & 30\%         & 50\%         \\ \hline
		OPINE-Net$^+$ \cite{zhang2020optimization}(ID)   & 29.81/0.175 & 35.99/0.412 & 40.19/0.649 \\
		OPINE-Net$^+$ (FD)      & 30.10\scriptsize~${\textcolor{red}{+0.29}}$\normalsize/0.212 & 36.35\scriptsize~${\textcolor{red}{+0.36}}$\normalsize/0.449 & 40.50\scriptsize~${\textcolor{red}{+0.31}}$\normalsize/0.687 \\ \hline
		MADUN \cite{song2021memory}(ID)            & 29.89/3.126 & 36.90/3.336 & 40.75/3.545 \\
		MADUN (FD)              & 30.16\scriptsize~${\textcolor{red}{+0.27}}$\normalsize/3.402 & 37.12\scriptsize~${\textcolor{red}{+0.22}}$\normalsize/3.612 & 40.93\scriptsize~${\textcolor{red}{+0.18}}$\normalsize/3.821 \\ \shline
	\end{tabular}}
\end{table*}

\subsection{Comparison with State-of-the-Arts}
\begin{figure}[!t]
	\begin{center}
		\hspace{-6pt}\scalebox{0.542}{
			\begin{tabular}[b]{c@{ } c@{ }  c@{ } c@{ } c@{ }}
				\multirow{4}{*}{\includegraphics[width=.312\textwidth,valign=t]{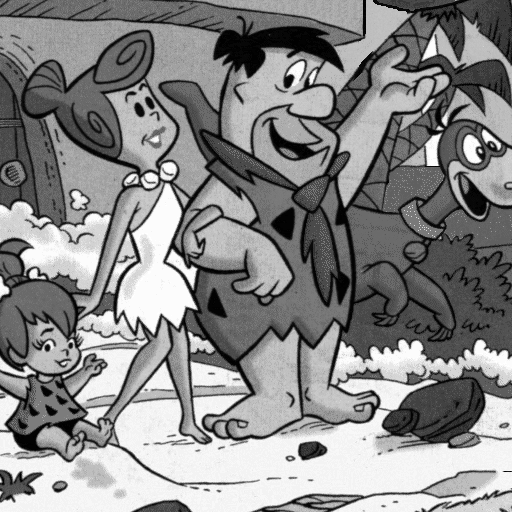}} &   
				\includegraphics[trim={0 10 480 470 },clip,width=.13\textwidth,valign=t]{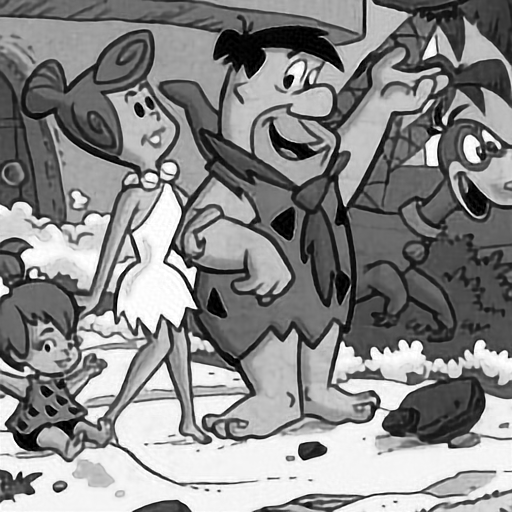}&
				\includegraphics[trim={0 10 480 470 },clip,width=.13\textwidth,valign=t]{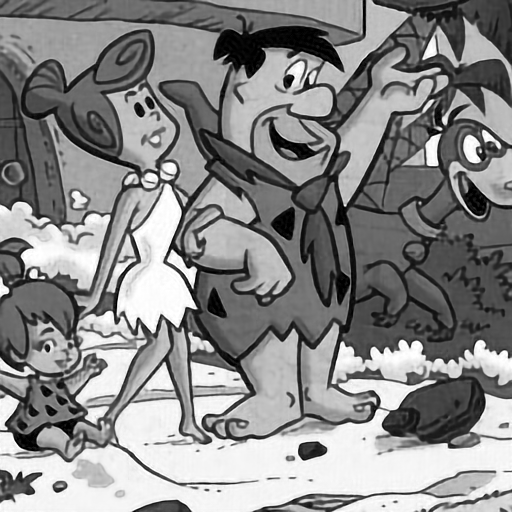}&   
				\includegraphics[trim={0 10 480 470 },clip,width=.13\textwidth,valign=t]{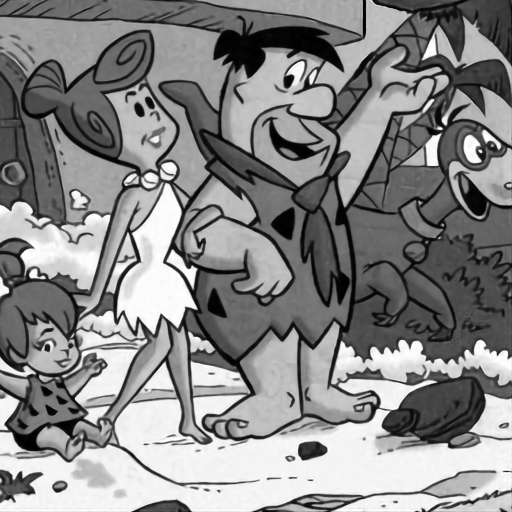}&
				\includegraphics[trim={0 10 480 470 },clip,width=.13\textwidth,valign=t]{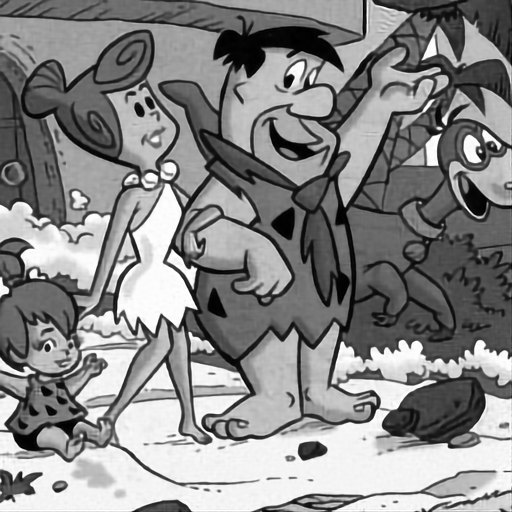}
				
				\\
				&  \scriptsize~24.07/0.7822 &\scriptsize~24.64/0.7996  & \scriptsize~28.17/0.8692& \scriptsize~26.50/0.8423\\
				& \scriptsize~CSNet$^+$& \scriptsize~SCSNet& \scriptsize~OPINE-Net$^+$& \scriptsize~AMP-Net\\
				
				&
				\includegraphics[trim={0 10 480 470 },clip,width=.13\textwidth,valign=t]{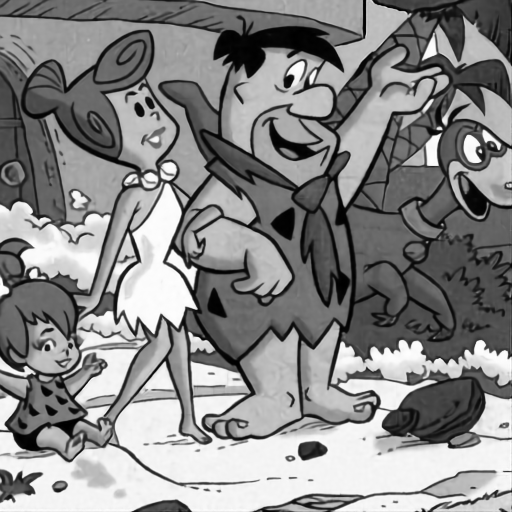}&
				\includegraphics[trim={0 10 480 470 },clip,width=.13\textwidth,valign=t]{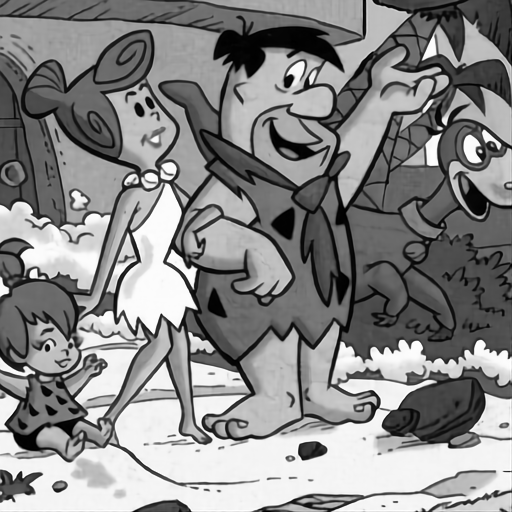}&
				\includegraphics[trim={0 10 480 470 },clip,width=.13\textwidth,valign=t]{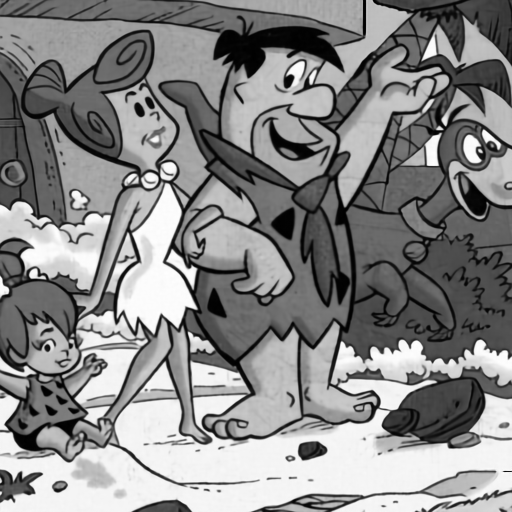}&  
				\includegraphics[trim={0 10 480 470 },clip,width=.13\textwidth,valign=t]{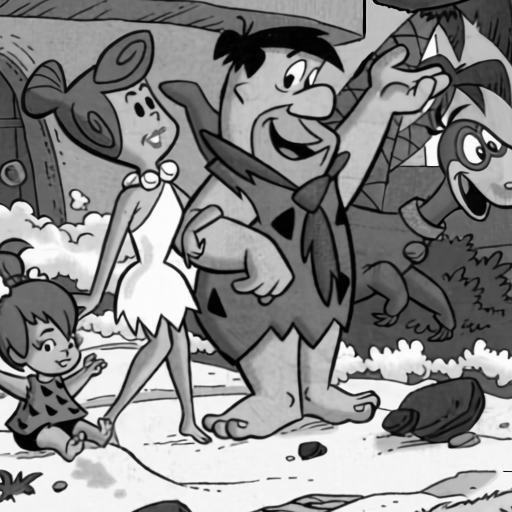}\\
				
				\scriptsize~PSNR(dB)/SSIM& \scriptsize~28.50/0.8739& \scriptsize~27.66/0.8732& \scriptsize~\textcolor{blue}{29.88}/\textcolor{blue}{0.8885}
				& \scriptsize~\textcolor{red}{30.15}/\textcolor{red}{0.8907}\\
				\scriptsize~Ground Truth& \scriptsize~COAST& \scriptsize~MADUN& \scriptsize~\textbf{PRL-PGD}& \scriptsize~\textbf{PRL-PGD$^+$}
				\\
		\end{tabular}}
  
		\hspace{-6pt}\scalebox{0.542}{
			\begin{tabular}[b]{c@{ } c@{ }  c@{ } c@{ } c@{ }}
				\multirow{4}{*}{\includegraphics[width=.312\textwidth,valign=t]{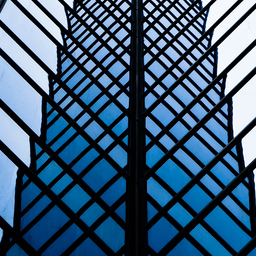}} &   
				\includegraphics[trim={140 226 86 0 },clip,width=.13\textwidth,valign=t]{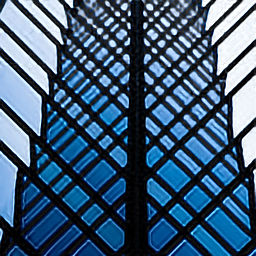}&
				\includegraphics[trim={140 226 86 0 },clip,width=.13\textwidth,valign=t]{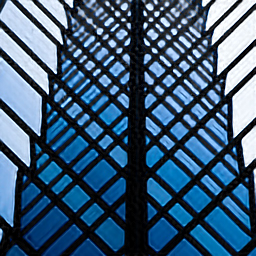}&   
				\includegraphics[trim={140 226 86 0 },clip,width=.13\textwidth,valign=t]{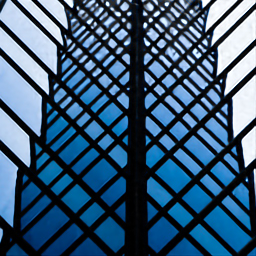}&
				\includegraphics[trim={140 226 86 0 },clip,width=.13\textwidth,valign=t]{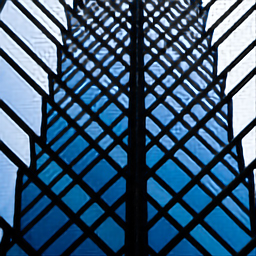}
				
				\\
				&  \scriptsize~17.91/0.6389 &\scriptsize~19.05/0.7309  & \scriptsize~23.54/0.9378& \scriptsize~20.16/0.8404\\
				& \scriptsize~CSNet$^+$& \scriptsize~SCSNet& \scriptsize~OPINE-Net$^+$& \scriptsize~AMP-Net\\
				
				&
				\includegraphics[trim={140 226 86 0 },clip,width=.13\textwidth,valign=t]{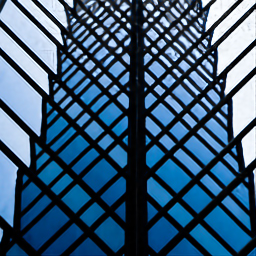}&
				\includegraphics[trim={140 226 86 0 },clip,width=.13\textwidth,valign=t]{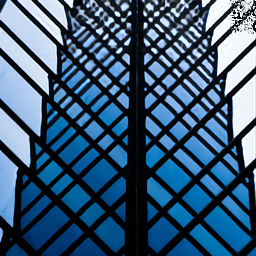}&
				\includegraphics[trim={140 226 86 0 },clip,width=.13\textwidth,valign=t]{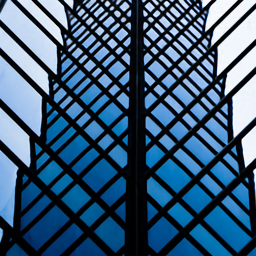}&  
				\includegraphics[trim={140 226 86 0 },clip,width=.13\textwidth,valign=t]{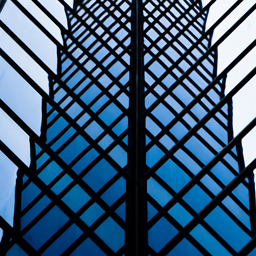}\\
				
				\scriptsize~PSNR(dB)/SSIM& \scriptsize~23.22/0.9370& \scriptsize~20.86/0.9309& \scriptsize~\textcolor{blue}{30.36}/\textcolor{blue}{0.9811}
				& \scriptsize~\textcolor{red}{31.70}/\textcolor{red}{0.9840}\\
				\scriptsize~Ground Truth& \scriptsize~COAST& \scriptsize~MADUN&\scriptsize~\textbf{PRL-PGD}& \scriptsize~\textbf{PRL-PGD$^+$}
				\\
		\end{tabular}}
	\end{center}
	\caption{Visual comparisons on recovering two images from Set11 \cite{kulkarni2016reconnet} (top) and Urban100 \cite{huang2015single} (bottom) benchmarks respectively under the setting of $\gamma=10\%$.}
	\label{fig: vis_comp1}
\end{figure}

We evaluate our PRL by comparing PRL-PGD and PRL-RND with six representative CS networks: CSNet$^+$ \cite{shi2019image}, SCSNet \cite{shi2019scalable}, OPINE-Net$^+$ \cite{zhang2020optimization} (9-stage), AMP-Net \cite{zhang2020amp} (9-stage), COAST \cite{you2021coast} (20-stage) and MADUN \cite{song2021memory} (25-stage). All of them are enhanced by the deblocking strategies with a data-driven adaptively learned sampling matrix. Tab.~\ref{tab:comp1} reports PSNR/SSIM scores of different methods on Set11 \cite{kulkarni2016reconnet} and Urban100 \cite{huang2015single} for five CS ratios. Due to the deficiency of explicit physical information guidance, the physics-free CSNet$^+$ \cite{shi2019image} and SCSNet \cite{shi2019scalable} remain a large performance gap to the physics-engaged networks. By taking the random projection augmentation and memory enhancement schemes respectively, COAST \cite{you2021coast} and MADUN \cite{song2021memory} surpass OPINE-Net$^+$ \cite{zhang2020optimization} and AMP-Net \cite{zhang2020amp} in nearly all cases. Integrating the proposed adaptive FD recovery scheme and multiscale high-throughput unrolled architecture, our PRL networks achieve the best accuracies with the shortest distances of 1.44dB/0.0187 when $\gamma =10\%$ and even 0.77dB/0.0011 when $\gamma =50\%$ from existing best results. To compare the efficiency of these methods, we evaluate the average inference time and computational complexity (GFLOPs) in processing a 256$\times$256 image on a 1080Ti GPU with $\gamma =50\%$, report in the last two columns of Tab.~\ref{tab:comp1} and visualize the results in Fig.~\ref{fig:comp_time_PSNR}(left). Although the default PRL networks benefit from their immense capacities with large-scale parameters, they enjoy real-time speeds (about 26 frames per second for PRL-PGD) and high sampling-recovery efficiencies. Tab.~\ref{tab:computation_complexity} further provides the evaluations of the time and computational complexity of PRL-PGD with different 
image sizes. The visual comparisons in Fig.~\ref{fig: vis_comp1} manifest the consistent superiority of PRL with keeping a significant leading in PSNR/SSIM and producing more details and textures than the competing methods. We note that the performance gains of PRL mainly come from its larger capacity ($\ge 30$ stages), compact architecture and learned abundant image prior instead of only the parameter number increase since the weight-sharing of PRL-PGD-S (with 2.6M parameters) brings little accuracy drops. Its accuracy can not be directly achieved by previous methods by adding more parameters. More analyses about the effect of our unrolling scheme will be provided in the following section and our \textbf{Appendices}.

\subsection{Ablation Studies}
\label{sec:abla}
\textbf{Effect of FD Recovery:} We apply the high-throughput PGD stage structure in Fig.~\ref{fig:stage_PRL}(c) with $r\equiv 1$ and $C=D=64$ to the plain FD architecture in Fig.~\ref{fig:arch}(b) to obtain an \underline{improved} network and train it with the same setting as the baseline. Fig.~\ref{fig:Train400}(d) exhibits that compared with the reduced (c), our FD-improved achieves more stable convergences and an average PSNR gain of 0.88dB while eliminating the degradation via activating the sufficient physics-guided refinements and transmissions. Similar phenomenons are observed in other settings including $C=32$ and $K\ge 80$ with other changes in the learning rate and training set. We also find that the adaptive gradient descent step sizes \cite{zhang2018ista,you2021ista} bring few effects. To get a deeper insight into FD, we investigate the PRL-PGD$^+$ with two settings: $D=q$ and $D=16$ and various physical feature dimensionalities $q$, where the first strictly limits the channel numbers of the features transmitted between adjacent stages, and the second keeps the high-throughput property but reduces the physical channel number to $r^2\times q$ by disabling $r^2\times(16-q)$ kernels with fixing weights/biases to zero in $\mathcal{H}_\text{grad}^{(k)}$ for each scale, \textit{i.e.} there are only $r^2\times q$ channels are physics-engaged. Tab.~\ref{tab:abla_q} shows that for the first setting, the higher feature capacities will lead to better results by enlarging the inter-stage flow. For the second setting, there exists a sharp performance increase (0.87dB) when $q\le 1$ and a mild PSNR growth when $q>1$. In addition, the comparison between these two settings indicates that the physics-free feature part also contains meaningful and supplemental contents, thus revealing the great importance of physics utilization and transmission bottleneck elimination. We further apply the FD unrolling to two existing networks in ID: OPINE-Net$^+$ \cite{zhang2020optimization} and MADUN \cite{song2021memory}. Tab.~\ref{tab:boost_existing_duns} exhibits the breakthrough on their optimal versions with the PSNR boosts of 0.36/0.20dB and 
parameter increment of about only 9\%, thus verifying the effectiveness of our FD generalization.

\begin{table*}[!t]
	\caption{PSNR (dB) results on Set11 \cite{kulkarni2016reconnet} with $\gamma=50\%$ and the inference time of FD high-throughput networks with three different types of architectures: the plain architecture, our PRL$^*$ and PRL ones.}
	\label{tab:abla_arch}
	\centering
	\scalebox{0.8}{
	\begin{tabular}{c|cccc|c}
		\shline
		Stage Number $K$                                        & 10      & 20       & 30       & 40       & \begin{tabular}[c]{@{}c@{}}Feature Domain Dimensionality $d$\end{tabular} \\ \hline
		\rowcolor[HTML]{CBFFCA} 
		\cellcolor[HTML]{CBFFCA}                                & 40.75   & 41.14    & 41.25    & 41.31    & \cellcolor[HTML]{CBFFCA}                                                    \\
		\rowcolor[HTML]{CBFFCA} 
		\cellcolor[HTML]{CBFFCA}                                & /25.685 & /50.824  & /76.239  & /101.688 & \multirow{-2}{*}{\cellcolor[HTML]{CBFFCA}32}                                \\ \hhline{>{\arrayrulecolor[HTML]{CBFFCA}}->{\arrayrulecolor{black}}|-----}
		\rowcolor[HTML]{CBFFCA} 
		\cellcolor[HTML]{CBFFCA}                                & 40.91   & 41.28    & 41.50    & 41.53    & \cellcolor[HTML]{CBFFCA}                                                    \\
		\rowcolor[HTML]{CBFFCA} 
		\multirow{-4}{*}{\cellcolor[HTML]{CBFFCA}Plain(FD)-PGD} & /55.929 & /111.608 & /167.920 & /201.092 & \multirow{-2}{*}{\cellcolor[HTML]{CBFFCA}64}                                \\ \hline \hline
		Group Stage Number $K$                                  & 1       & 3        & 5        & 7        & \begin{tabular}[c]{@{}c@{}}Feature Domain Dimensionality $d$\end{tabular} \\ \hline
		\rowcolor[HTML]{FFEBC4} 
		\cellcolor[HTML]{FFEBC4}                                & 40.61   & 41.17    & 41.41    & 41.56    & \cellcolor[HTML]{FFEBC4}                                                    \\
		\rowcolor[HTML]{FFEBC4} 
		\cellcolor[HTML]{FFEBC4}                                & /36.534 & /106.462 & /177.535 & /240.427 & \multirow{-2}{*}{\cellcolor[HTML]{FFEBC4}8}                                 \\ \hhline{>{\arrayrulecolor[HTML]{FFEBC4}}->{\arrayrulecolor{black}}|-----}
		\rowcolor[HTML]{FFEBC4} 
		\cellcolor[HTML]{FFEBC4}                                & 40.78   & 41.29    & 41.54    & 41.70    & \cellcolor[HTML]{FFEBC4}                                                    \\
		\rowcolor[HTML]{FFEBC4} 
		\multirow{-4}{*}{\cellcolor[HTML]{FFEBC4}PRL$^*$-PGD}   & /88.218 & /239.967 & /395.687 & 548.295  & \multirow{-2}{*}{\cellcolor[HTML]{FFEBC4}16}                                \\ \hline
		\rowcolor[HTML]{C0D0FC} 
		\cellcolor[HTML]{C0D0FC}                                & 40.77   & 41.31    & 41.52    & 41.59    & \cellcolor[HTML]{C0D0FC}                                                    \\
		\rowcolor[HTML]{C0D0FC} 
		\cellcolor[HTML]{C0D0FC}                                & /\textcolor{red}{8.709}  & /\textcolor{red}{24.017}  & /\textcolor{red}{41.402}  & /\textcolor{red}{54.743}  & \multirow{-2}{*}{\cellcolor[HTML]{C0D0FC}8}                                 \\ \hhline{>{\arrayrulecolor[HTML]{C0D0FC}}->{\arrayrulecolor{black}}|-----}
		\rowcolor[HTML]{C0D0FC} 
		\cellcolor[HTML]{C0D0FC}                                & \textcolor{blue}{40.95}   & \textcolor{blue}{41.45}    & \textcolor{blue}{41.65}    & \textcolor{blue}{41.78}    & \cellcolor[HTML]{C0D0FC}                                                    \\
		\rowcolor[HTML]{C0D0FC} 
		\multirow{-4}{*}{\cellcolor[HTML]{C0D0FC}PRL-PGD}       & /\textcolor{blue}{11.169} & /\textcolor{blue}{31.397}  & /\textcolor{blue}{51.746}  & /\textcolor{blue}{72.067}  & \multirow{-2}{*}{\cellcolor[HTML]{C0D0FC}16}                                \\ \hline
		\rowcolor[HTML]{E2C6FF} 
		\cellcolor[HTML]{E2C6FF}                                & 40.80   & 41.37    & 41.63    & 41.70    & \cellcolor[HTML]{E2C6FF}                                                    \\
		\rowcolor[HTML]{E2C6FF} 
		\cellcolor[HTML]{E2C6FF}                                & /13.226 & /37.480  & /62.147  & /87.125  & \multirow{-2}{*}{\cellcolor[HTML]{E2C6FF}8}                                 \\ \hhline{>{\arrayrulecolor[HTML]{E2C6FF}}->{\arrayrulecolor{black}}|-----}
		\rowcolor[HTML]{E2C6FF} 
		\cellcolor[HTML]{E2C6FF}                                & \textcolor{red}{40.97}   & \textcolor{red}{41.53}    & \textcolor{red}{41.72}    & \textcolor{red}{41.84}    & \cellcolor[HTML]{E2C6FF}                                                    \\
		\rowcolor[HTML]{E2C6FF} 
		\multirow{-4}{*}{\cellcolor[HTML]{E2C6FF}PRL-RND}       & /16.534 & /47.712  & /79.213  & /111.307 & \multirow{-2}{*}{\cellcolor[HTML]{E2C6FF}16}                                \\ \shline
	\end{tabular}}
\end{table*}

\begin{table*}[!t]
	\caption{\textbf{(1) Left:} PSNR (dB) results with the corresponding \textcolor{red}{vertical} gains on Urban100 \cite{huang2015single} of the PRL networks with five different FD fusion module settings $\mathcal{H}_\text{grad/merge}^{(k)}$ when $\gamma=30\%$. \textbf{{(2) Right:}} PSNR results on Set11 \cite{kulkarni2016reconnet} of the PRL networks with four different configurations of the identity skip connections when $\gamma=10\%$.}
	\label{tab:abla_g}
	\begin{minipage}{0.5\textwidth}
	\centering
		\scalebox{0.70}{\begin{tabular}{c|cccc}
		\shline
        $\mathcal{H}_\text{grad/merge}^{(k)}$    & \begin{tabular}[c]{@{}c@{}}PRL\\-PGD\end{tabular}&\begin{tabular}[c]{@{}c@{}}PRL\\-PGD$^+$\end{tabular}&\begin{tabular}[c]{@{}c@{}}PRL\\-RND\end{tabular}&\begin{tabular}[c]{@{}c@{}}PRL\\-RND$^+$\end{tabular}\\ \hline
        $\mathbf{A}^\top(\mathbf{A}(\cdot)-\mathbf{y})$ & 33.98   & 34.45                          & 34.01   & 34.49       \\ \hline
        \textcolor{purple}{1$\times$1} Conv                                 & 34.06\scriptsize~${\textcolor{red}{+0.08}}$   & 34.55\scriptsize~${\textcolor{red}{+0.10}}$                          & 34.08\scriptsize~${\textcolor{red}{+0.07}}$   & 34.57\scriptsize~${\textcolor{red}{+0.08}}$       \\ \hline
        \begin{tabular}[c]{@{}c@{}}\textcolor{purple}{1$\times$1} Conv\\+ \textcolor{blue}{Sigmoid}\end{tabular}                       & 34.10\scriptsize~${\textcolor{red}{+0.04}}$   & 34.58\scriptsize~${\textcolor{red}{+0.03}}$                          & 34.17\scriptsize~${\textcolor{red}{+0.09}}$   & 34.62\scriptsize~${\textcolor{red}{+0.05}}$       \\ \hline
        \textcolor{purple}{3$\times$3} Conv                                 & 34.15\scriptsize~${\textcolor{red}{+0.05}}$   & 34.65\scriptsize~${\textcolor{red}{+0.07}}$                          & 34.17\scriptsize~${\textcolor{red}{+0.09}}$   & 34.69\scriptsize~${\textcolor{red}{+0.07}}$       \\ \hline
        \begin{tabular}[c]{@{}c@{}}\textcolor{purple}{3$\times$3} Conv\\+ \textcolor{blue}{Sigmoid}\end{tabular}                        & 34.21\scriptsize~${\textcolor{red}{+0.06}}$   & 34.73\scriptsize~${\textcolor{red}{+0.08}}$                          & 34.24\scriptsize~${\textcolor{red}{+0.07}}$   & 34.78\scriptsize~${\textcolor{red}{+0.09}}$       \\ \shline
        \end{tabular}}
    \end{minipage}
    \begin{minipage}{0.5\textwidth}
    \scalebox{0.74}{\begin{tabular}{cc|cc}
    	\shline
        \begin{tabular}[c]{@{}c@{}}Encoder-to-decoder\\ Connection\end{tabular} & \begin{tabular}[c]{@{}c@{}}Intra-stage\\ Connection\end{tabular} & \begin{tabular}[c]{@{}c@{}}PRL\\-PGD\end{tabular}&\begin{tabular}[c]{@{}c@{}}PRL\\-RND\end{tabular}\\ \hline
        $\times$                                                                & $\times$                                                         & 31.04   & 18.57   \\
        $\checkmark$                                                            & $\times$                                                         & 31.27\scriptsize~${\textcolor{red}{+0.23}}$   & 29.87\scriptsize~${\textcolor{red}{+11.30}}$   \\
        $\times$                                                                & $\checkmark$                                                     & 31.39\scriptsize~${\textcolor{red}{+0.12}}$   & 31.32\scriptsize~${\textcolor{red}{+1.45}}$   \\
        $\checkmark$                                                            & $\checkmark$                                                     & 31.49\scriptsize~${\textcolor{red}{+0.10}}$   & 31.46\scriptsize~${\textcolor{red}{+0.14}}$ \\ \shline  
        \end{tabular}}
    \end{minipage}
\end{table*}

\begin{figure*}[!t]
	\centering
 \includegraphics[width=1.0\textwidth]{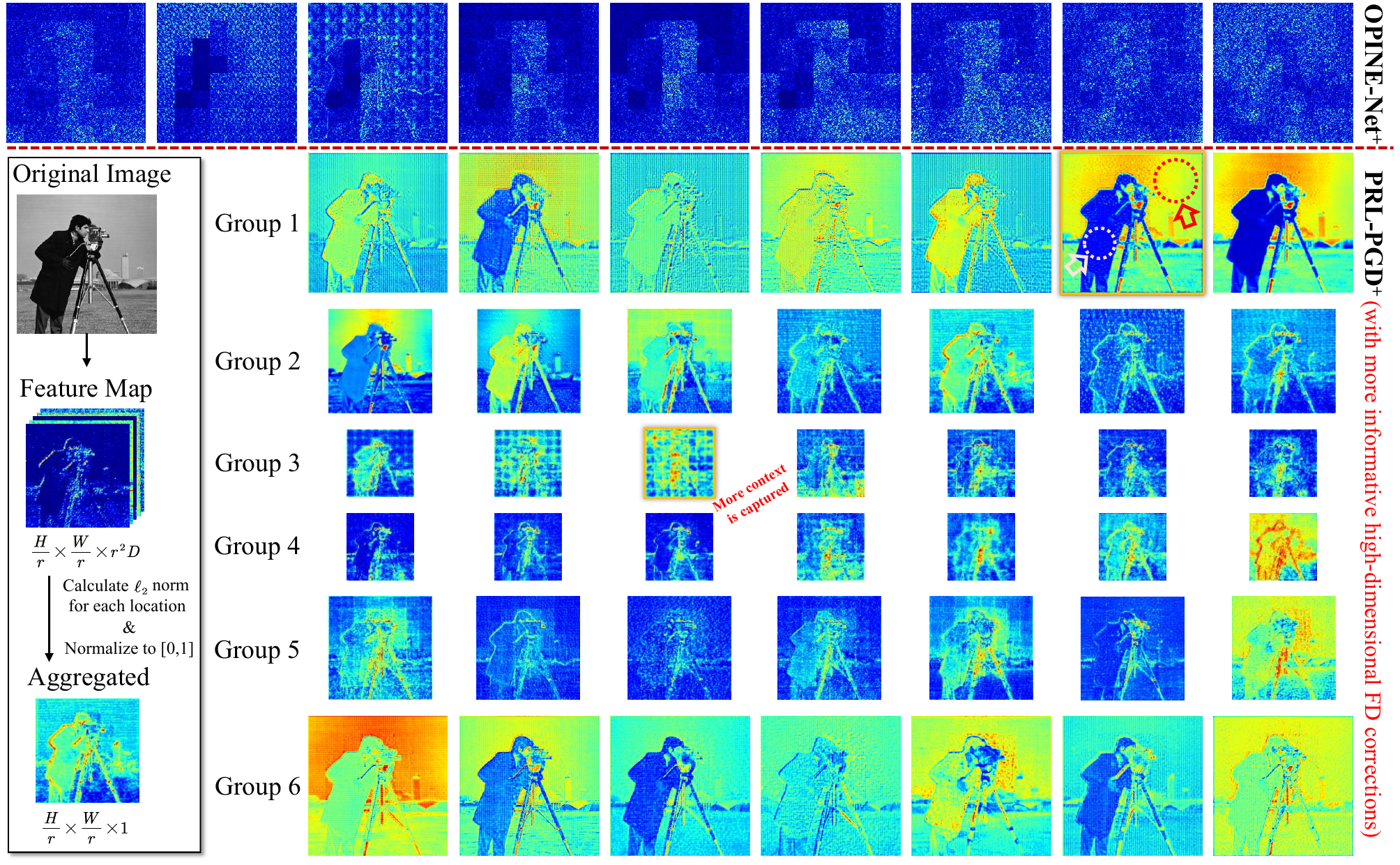}
	\caption{2-D visualizations of the gradient components $\rho^{(k)}\mathbf{A}^\top(\mathbf{A}(\cdot)-\mathbf{y})$ and $\rho\mathcal{H}_\text{grad}^{(k)}(\cdot)$ in all the unrolled stages when recovering an image from Set11 \cite{kulkarni2016reconnet} with $\gamma=10\%$. The 1st and 2nd-7th rows correspond to the 9-stage OPINE-Net$^+$ \cite{zhang2020optimization} and the 1st-6th groups of our proposed PRL-PGD$^+$, respectively.}
	\label{fig:abla_g}
\end{figure*}

\begin{figure*}[!t]
    \centering
    \includegraphics[width=1.0\textwidth]{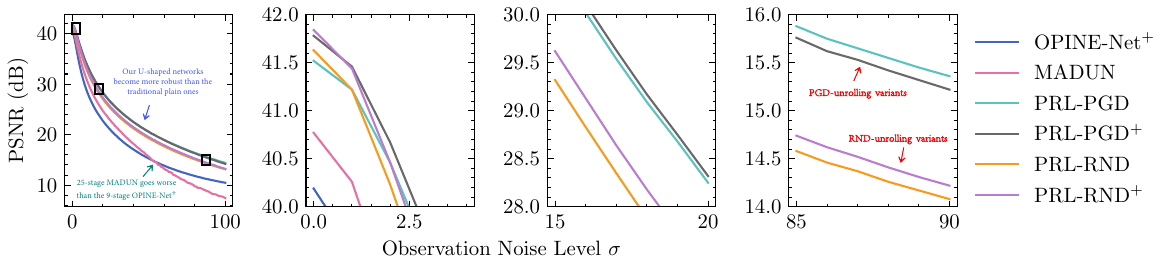}
	\caption{PSNR curves of various networks on Set11 \cite{kulkarni2016reconnet} with $\gamma=50\%$ when suffering from the disturbance of additive white Gaussian noise $\epsilon$ in the observation process $\mathbf{y}=\mathbf{Ax}+\epsilon$ with the standard deviation (noise level) $\sigma\in\left[0,100\right]$.}
	\label{fig:abla_noise}
\end{figure*}

\begin{table*}
\caption{Quantitative comparisons of PSNR(dB)/SSIM among ten sparse-view CT reconstruction methods and our PRL-PGD network (with $K=1$, marked by \textbf{PRL-PGD-1}) on the Apple CT benchmark \cite{coban2020apple}. The star marks indicate the training time and number of parameters: \ding{72} ($\le$ 1 day/$\le 10^5$), \ding{72}\ding{72} (2-5 days/$10^5$-$10^6$), \ding{72}\ding{72}\ding{72} (6-14 days/$10^6$-$10^8$) and \ding{72}\ding{72}\ding{72}\ding{72} ($>$14 days/$>10^8$). The best and second-best accuracy scores are marked in red and blue, respectively.}
\label{tab:CT}
\centering
\resizebox{1.0\textwidth}{!}{\begin{tabular}{c|cccccccccc
>{\columncolor[HTML]{EFEFEF}}c }
\shline 
Method &
  FBP &
  CGLS &
  TV &
  iCT-UNet &
  CINN &
  MS-D-CNN &
  U-Net &
  ISTA U-Net &
  Learned P. D. &
  FISTA-Net &
  \textbf{PRL-PGD-1} \\ \hline
 &
  13.97 &
  15.25 &
  15.95 &
  19.28 &
  19.46 &
  20.55 &
  19.78 &
  20.48 &
  \textcolor{blue}{22.00} &
  21.05 &
  \textcolor{red}{29.89} \\
\multirow{-2}{*}{2-Angle} &
  /0.4379 &
  /0.6149 &
  /0.6614 &
  /\textcolor{blue}{0.7413} &
  /0.6740 &
  /0.6460 &
  /0.6758 &
  /0.6855 &
  /0.7401 &
  /0.7067 &
  /\textcolor{red}{0.8032} \\ \hline
 &
  17.09 &
  21.81 &
  29.00 &
  29.95 &
  34.84 &
  34.38 &
  33.51 &
  34.54 &
  \textcolor{blue}{35.85} &
  35.06 &
  \textcolor{red}{38.22} \\
\multirow{-2}{*}{10-Angle} &
  /0.5840 &
  /0.6189 &
  /0.7825 &
  /0.8465 &
  /\textcolor{blue}{0.8713} &
  /0.8367 &
  /0.8029 &
  /0.8540 &
  /0.8702 &
  /0.8498 &
  /\textcolor{red}{0.9122} \\ \hline
 &
  30.39 &
  33.05 &
  39.27 &
  36.07 &
  39.59 &
  \textcolor{blue}{39.85} &
  39.62 &
  38.86 &
  38.72 &
  39.15 &
  \textcolor{red}{40.85} \\
\multirow{-2}{*}{50-Angle} &
  /0.7143 &
  /0.7803 &
  /\textcolor{blue}{0.9149} &
  /0.8778 &
  /0.9134 &
  /0.9128 &
  /0.9133 &
  /0.8966 &
  /0.9008 &
  /0.8993 &
  /\textcolor{red}{0.9314} \\ \hline
Training Time &
  - &
  - &
  - &
  \ding{72}\ding{72} &
  \ding{72}\ding{72} &
  \ding{72}\ding{72}\ding{72}\ding{72} &
  \ding{72}\ding{72} &
  \ding{72}\ding{72}\ding{72} &
  \ding{72}\ding{72}\ding{72}\ding{72} &
  \ding{72}\ding{72}\ding{72} &
  \ding{72} \\ \hline
\#Param. &
  - &
  - &
  - &
  \ding{72}\ding{72}\ding{72}\ding{72} &
  \ding{72}\ding{72}\ding{72} &
  \ding{72} &
  \ding{72}\ding{72} &
  \ding{72}\ding{72}\ding{72} &
  \ding{72}\ding{72} &
  \ding{72} &
  \ding{72}\ding{72}\ding{72}\\
  \shline
\end{tabular}}
\end{table*}

\textbf{Effect of Multiscale Architecture:} Our architecture handles the insufficient performance saturations and the extending difficulties of plain ones \cite{zhang2020optimization,zhang2020amp,you2021ista} by deep multiscale unrolling. We construct plain and single-scale U-shaped variants where the former adopts the plain FD architecture in Fig.~\ref{fig:arch}(b) and the latter marked as PRL$^*$ substitutes the SConv and TConv in Fig.~\ref{fig:arch}(c) with normal Conv layer for enlarging feature capacities ($H\times W\times r^2D$) with the loss of multiscale perceptions. Tab.~\ref{tab:abla_arch} and Fig.~\ref{fig:comp_time_PSNR}(right) investigate the comparable networks with different $K$s and FD dimensionalities $C=D=d$. We observe that the single-scale PRL$^*$ surpasses the plain variant narrowly with huge increases in time complexity. PRL benefits from the high flexibilities close to PRL$^*$ and boosts the accuracy with large reductions of inference time as it mainly refines the features on smaller scales. We also discover that the performance of plain variants is going to be saturated when $K\ge 40$ and $d\ge 64$, but there remains an extending space for PRL. Note that even the most lightweight PRL-PGD variant with $K=1$ and 2.6M parameters obtains higher accuracy and is $\times$11 faster than the MADUN \cite{song2021memory} (about 115 frames per second) with 3.1M parameters, validating the necessity of multiscale unrolling generalization.

\textbf{Study of FD Fusion Module and Skip Connections:} Our FD fusion module learns the generalized physical constraint $\mathcal{F}$ in Eq.~(\ref{eq:opt_feature}) for PGD and estimates the range space component in Eq.~(\ref{eq:rnd_feature1}) for RND. Tab.~\ref{tab:abla_g}(left) reports the evaluations on PRL with five fusion module settings and shows higher accuracy compared with the fixed or less flexible operators brought by our default data-driven ones with larger receptive fields and Sigmoid nonlinearity in the last column. Fig.~\ref{fig:abla_g} collects the gradient components in all stages of OPINE-Net$^+$ \cite{zhang2020optimization} and PRL-PGD$^+$, and visualize the high-dimensional gradient features by aggregating each spatial location with $\ell_2$ norm (a greater value indicates a larger adjustment in FD). Compared with the pixel-level corrections brought by OPINE-Net$^+$ with fixed ID fidelity term, our PRL-PGD$^+$ identifies the image edges and distinguishes between the fore-/back-grounds in the front stages. It also effectively captures the context by exploiting the inter-scale FD relationships. As a simple but effective scheme, the additive skip connection \cite{he2016deep} has already been widely used in deep image restorations \cite{zhang2018image}. Our PRL also employs the encoder-to-decoder and intra-stage skip connections to exploit the long- and short-term dependencies. Tab.~\ref{tab:abla_g}(right) compares the PRL networks with four different settings. For PRL-PGD, these connections play a key role in enhancing the maintenance of high-frequency components in the decoder and proximal mapping in each stage. We also find that they are much more necessary for the PRL-RND with such an iterative range-nullspace complement scheme.

\textbf{PRL-PGD vs. PRL-RND:} Our multiscale FD unrolling generalizations correspond to the two distinct perspectives of the optimization and range-nullspace decomposition, which are implemented by the same basic modules and parameter numbers. From Tab.~\ref{tab:comp1} and Tab.~\ref{tab:abla_arch}, we observe that PRL-RND performs better than PRL-PGD when $\gamma>10\%$ but is hindered by its lower inference speeds. PRL-PGD can enjoy not only faster inferences but also simpler pipelines with higher robustness. Fig.~\ref{fig:abla_noise} investigates the PSNR of various networks with different levels of observation noise and reveals the lower robustness of the PRL-RND which works well with slight disturbance but goes worse than the PRL-PGD when the noise intensity increases. We also find that the 25-stage MADUN \cite{song2021memory} becomes worse than the 9-stage OPINE-Net$^+$ \cite{zhang2020optimization} when the noise level $\sigma\ge54$. Both of them incorporate fewer parameters and larger feature capacity with 32 channels but suffer from more significant performance drops than the PRL networks, thus verifying the undesirable noise sensitivity brought by the traditional plain architectures as the depth grows and favorable robustness achieved by the PRL networks benefited from their compact structures and adaptively learned rich image prior.

\begin{figure*}
	\begin{center}
		\resizebox{1.01\textwidth}{!}{
			\begin{tabular}[b]{c@{ } c@{ }  c@{ } c@{ } c@{ } c@{ } c@{ }  c@{ } c@{ } c@{ } c@{ } c@{ }}
                \scriptsize Ground Truth&
                \scriptsize FBP&
                \scriptsize CGLS&
                \scriptsize TV&
                \scriptsize iCT-UNet&
                \scriptsize CINN&
                \scriptsize MS-D-CNN&
                \scriptsize U-Net&
                \scriptsize ISTA U-Net&
                \scriptsize Learned P. D.&
                \scriptsize FISTA-Net&
                \scriptsize \textbf{PRL-PGD-1}\\
                
                \includegraphics[width=.11\textwidth,valign=t]{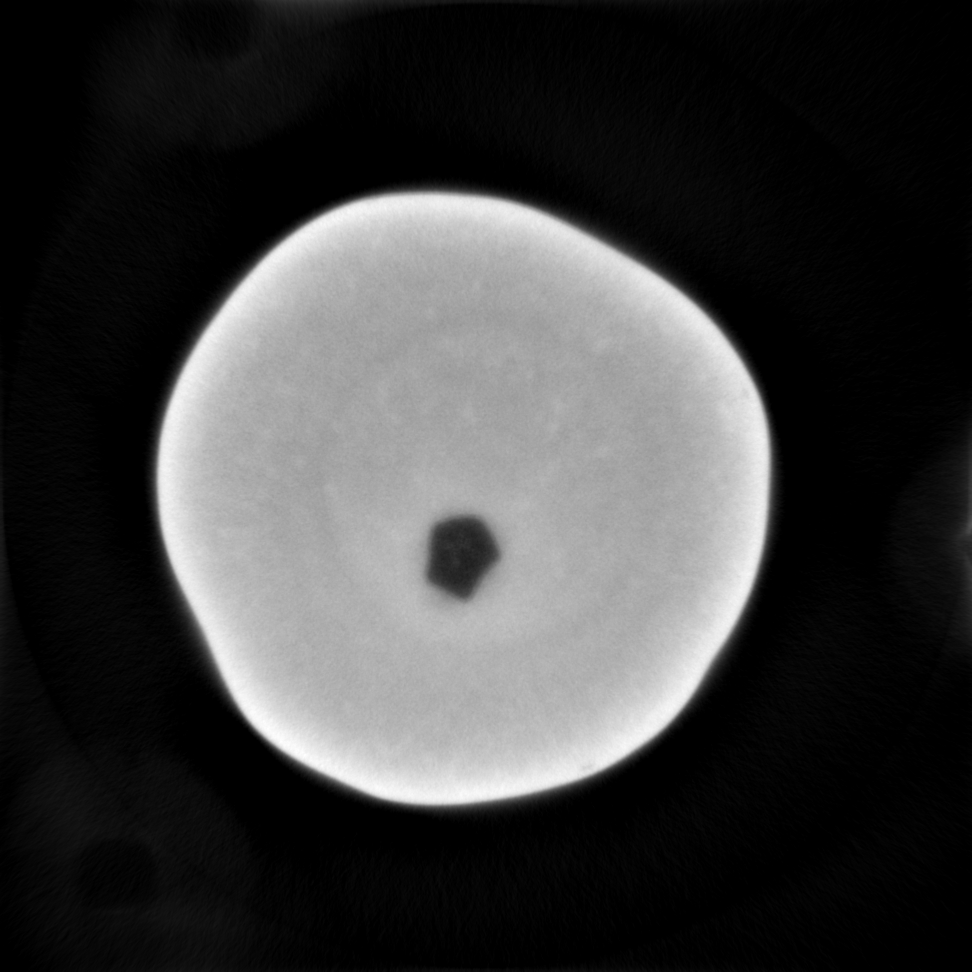} &
                \includegraphics[width=.11\textwidth,valign=t]{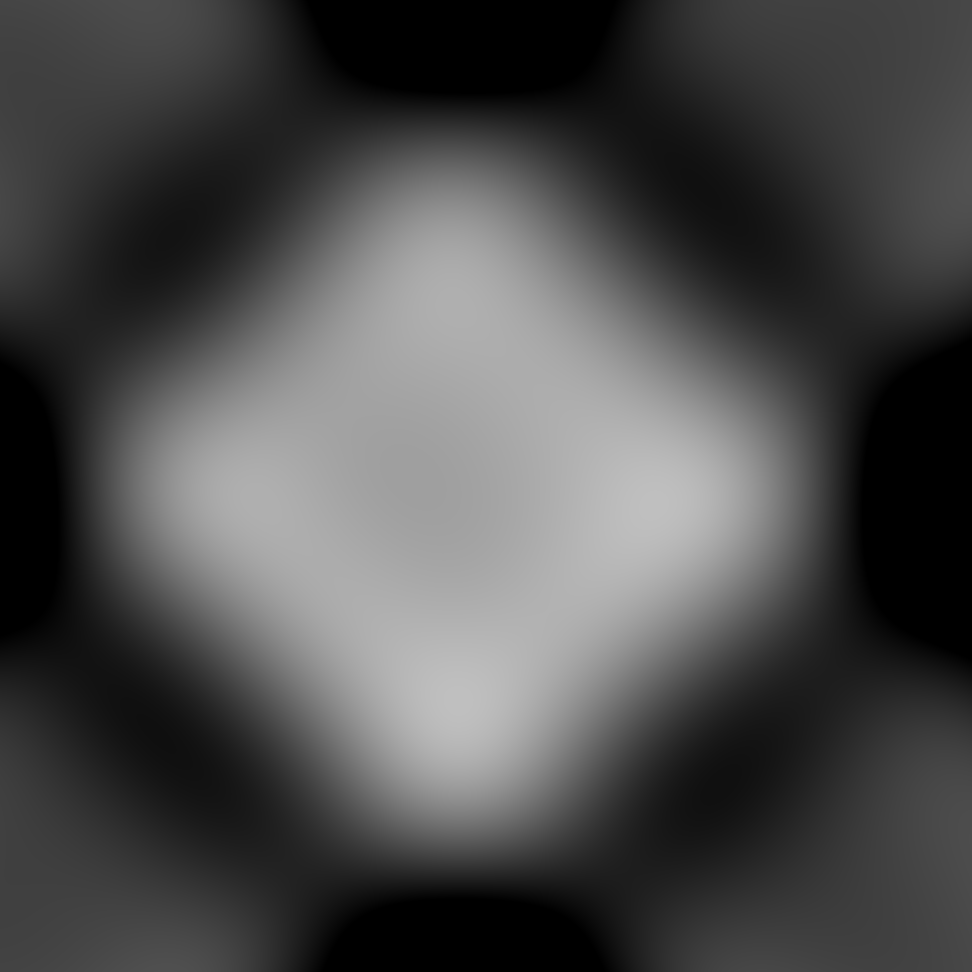} &
                \includegraphics[width=.11\textwidth,valign=t]{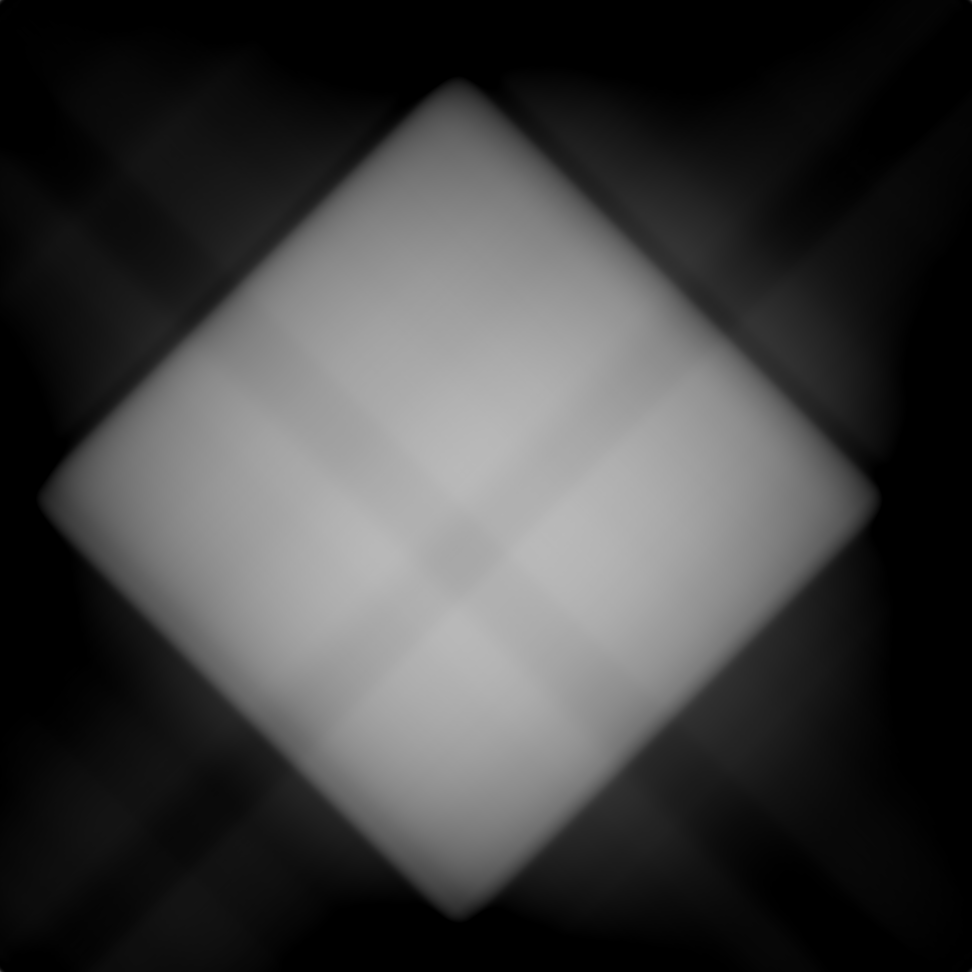} &
                \includegraphics[width=.11\textwidth,valign=t]{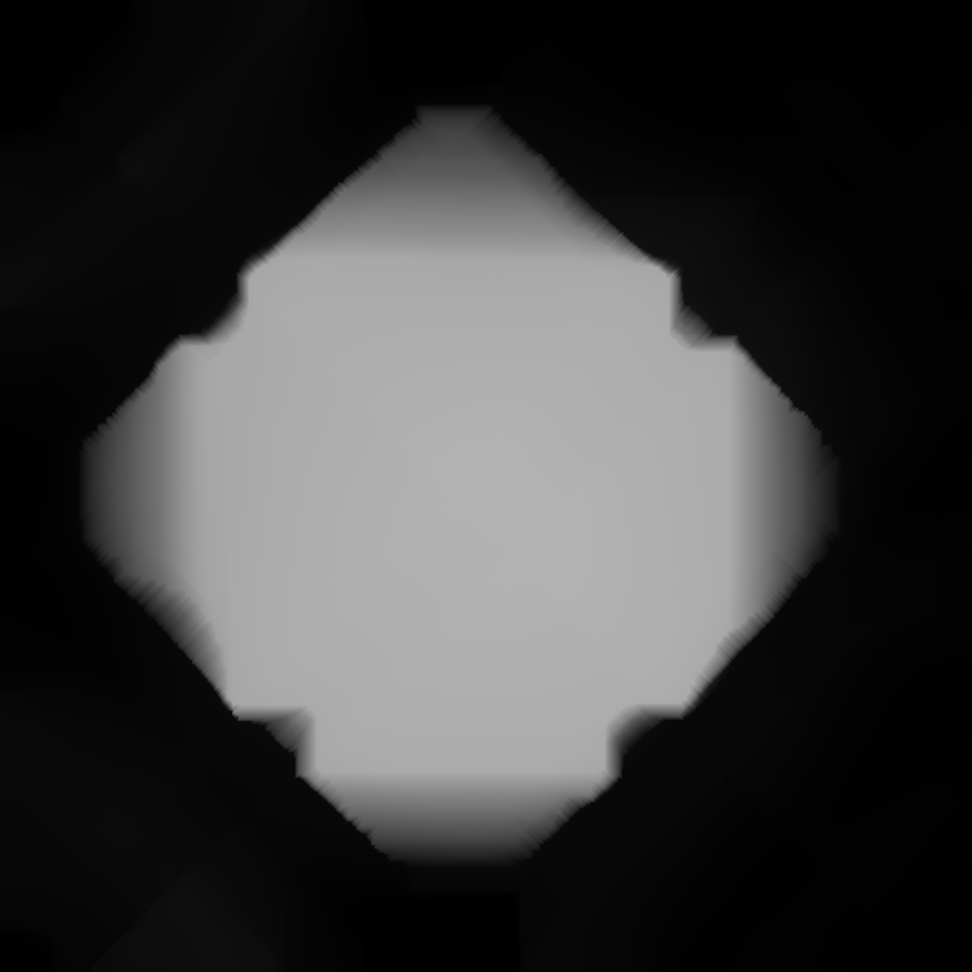} &
                \includegraphics[width=.11\textwidth,valign=t]{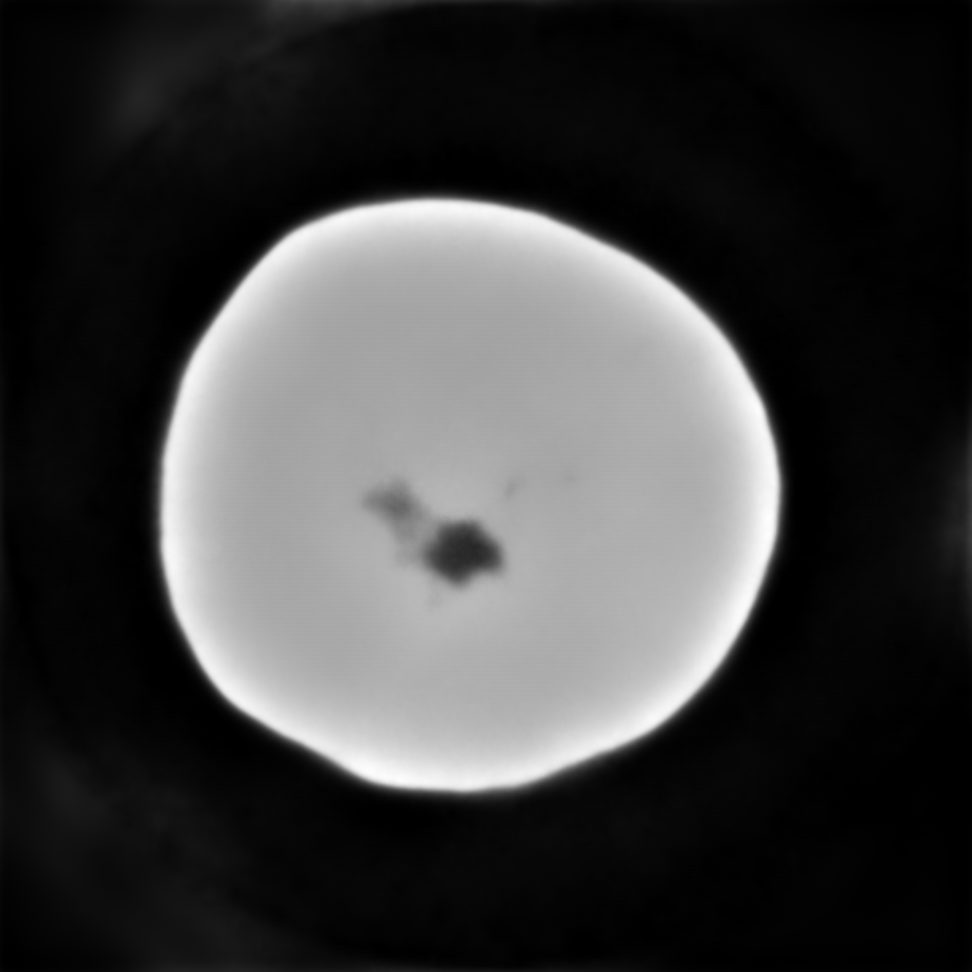} &
                \includegraphics[width=.11\textwidth,valign=t]{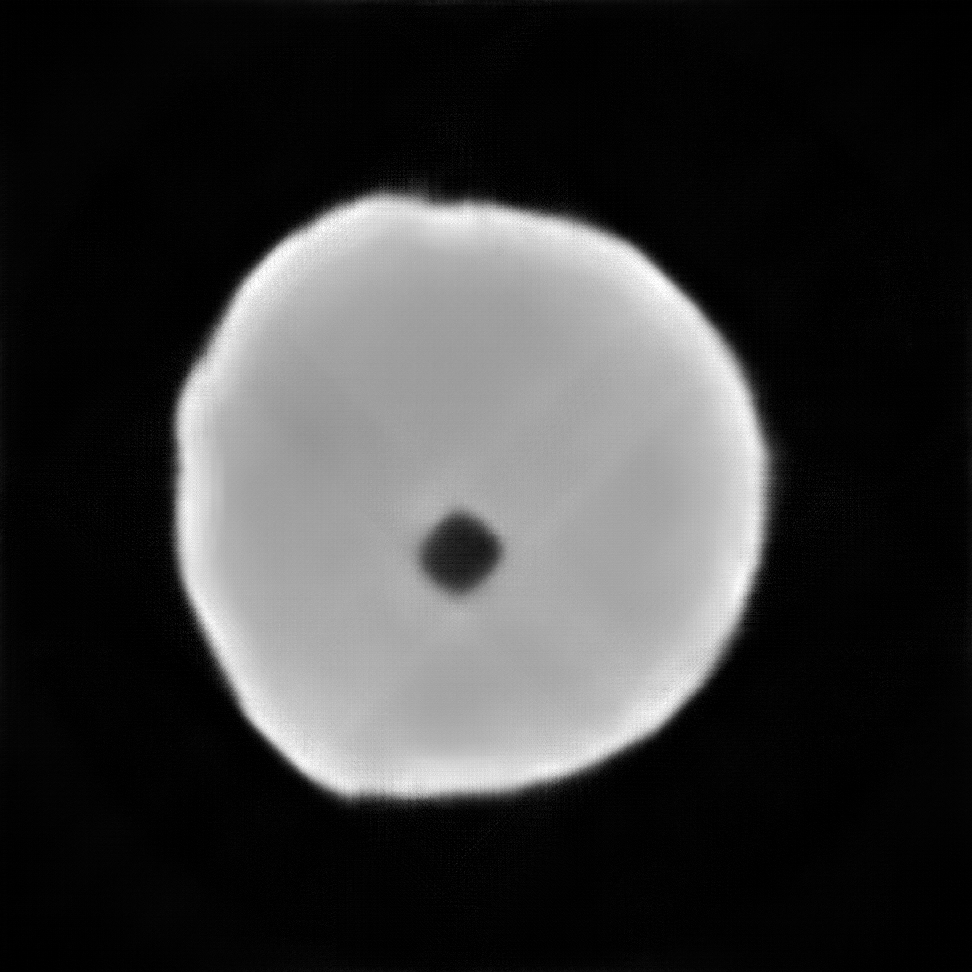} &
                \includegraphics[width=.11\textwidth,valign=t]{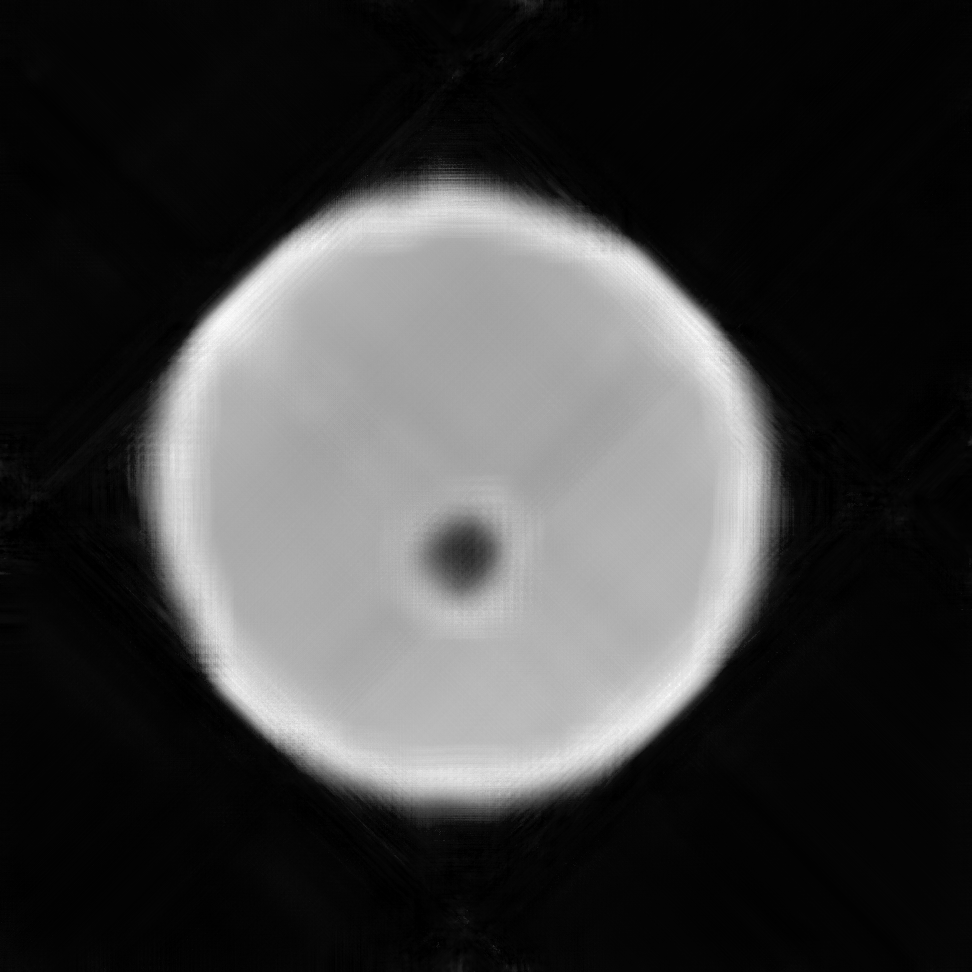} &
                \includegraphics[width=.11\textwidth,valign=t]{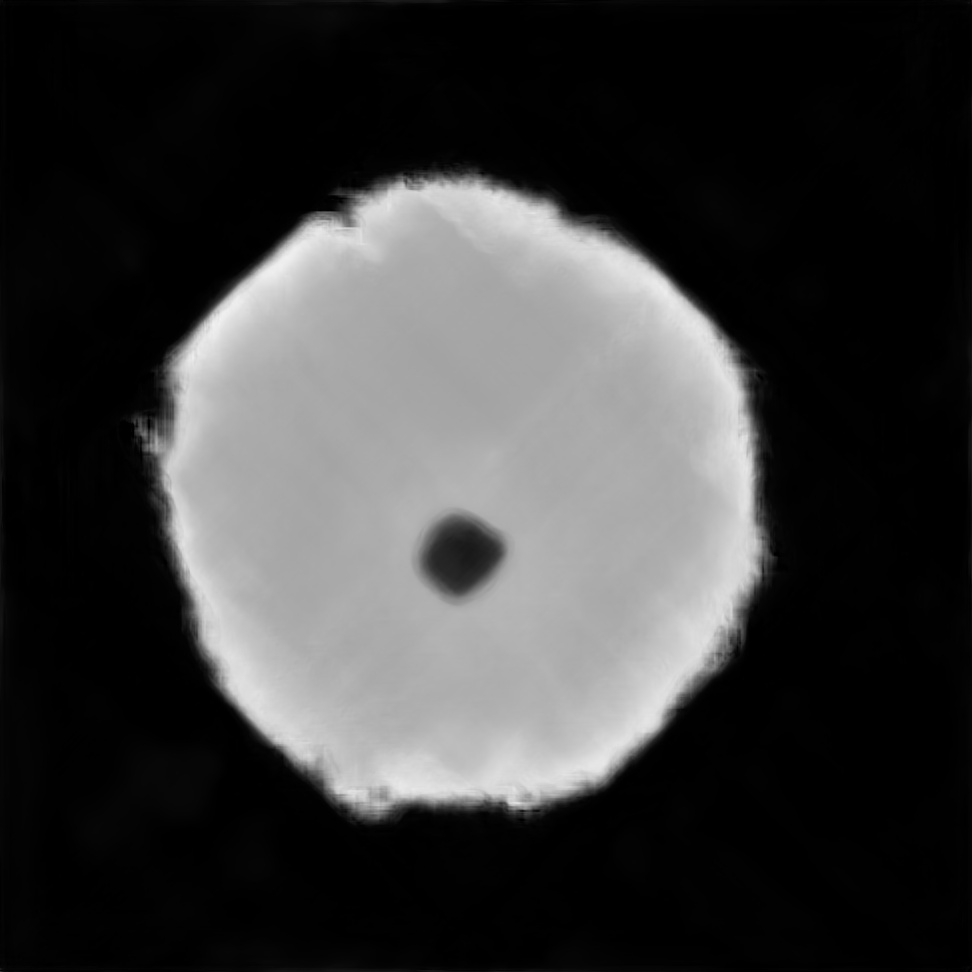} &
                \includegraphics[width=.11\textwidth,valign=t]{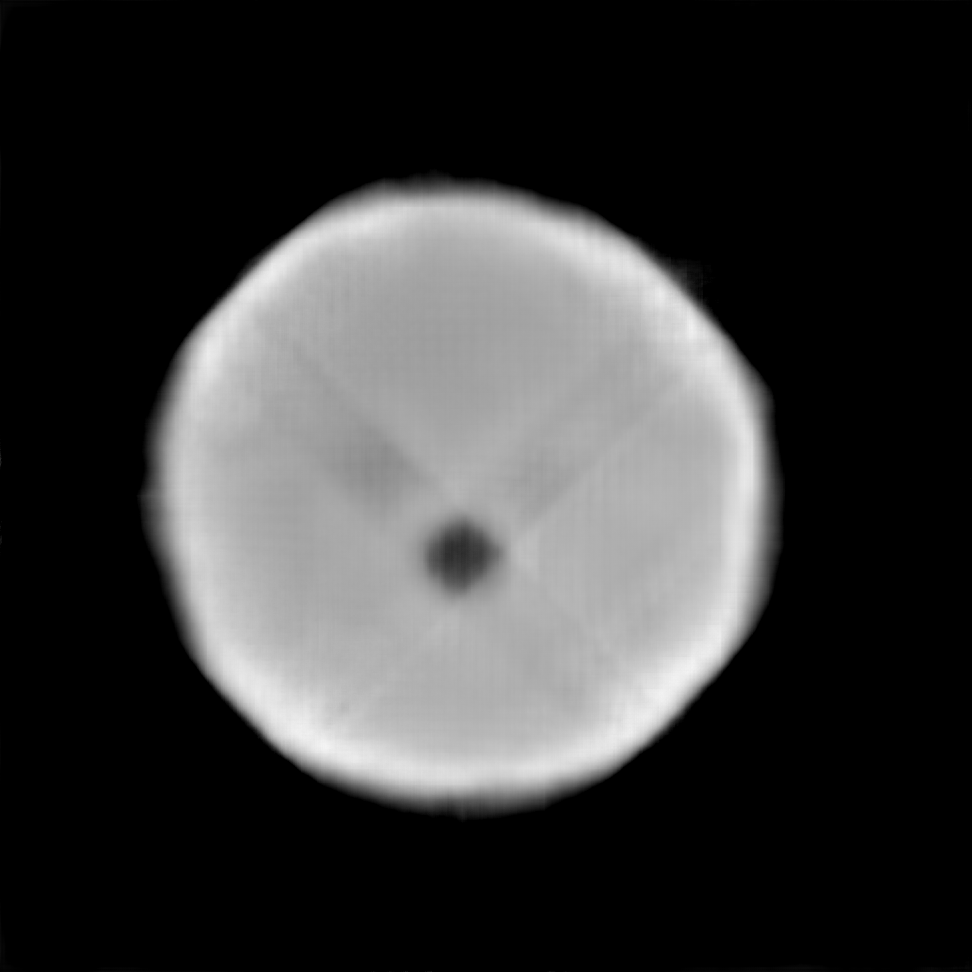} &
                \includegraphics[width=.11\textwidth,valign=t]{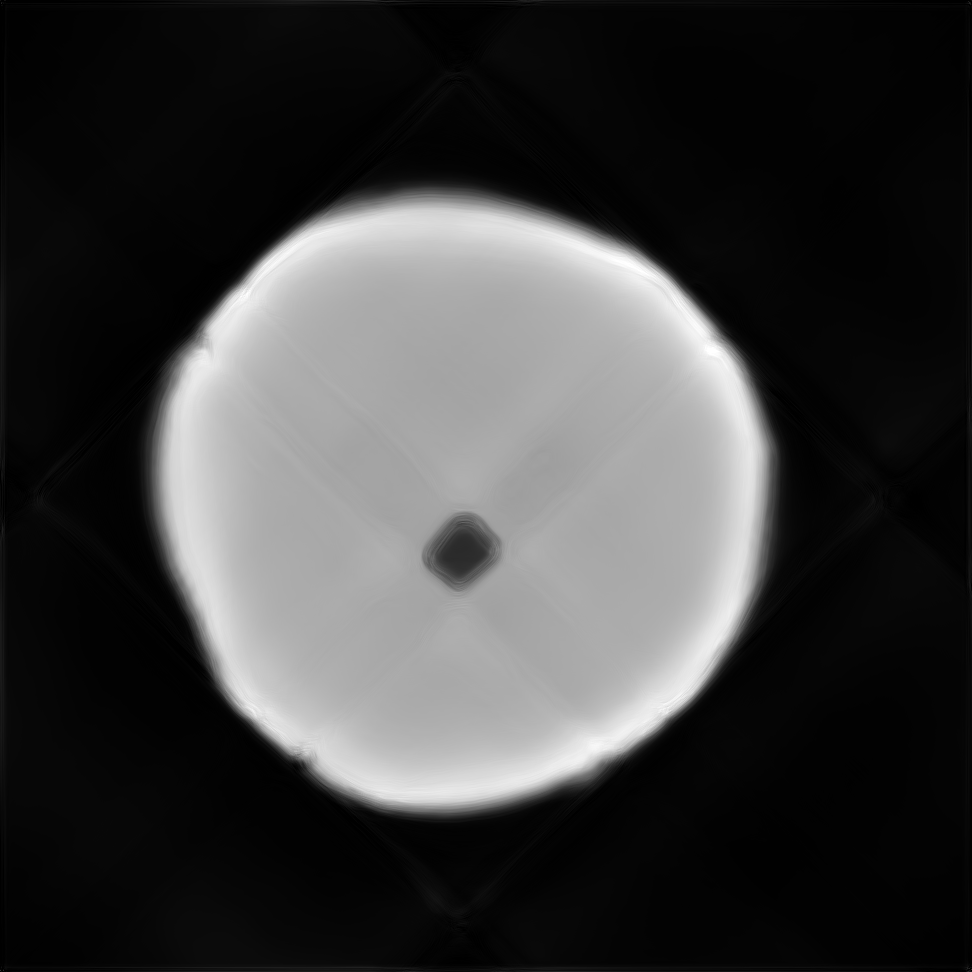} &
                \includegraphics[width=.11\textwidth,valign=t]{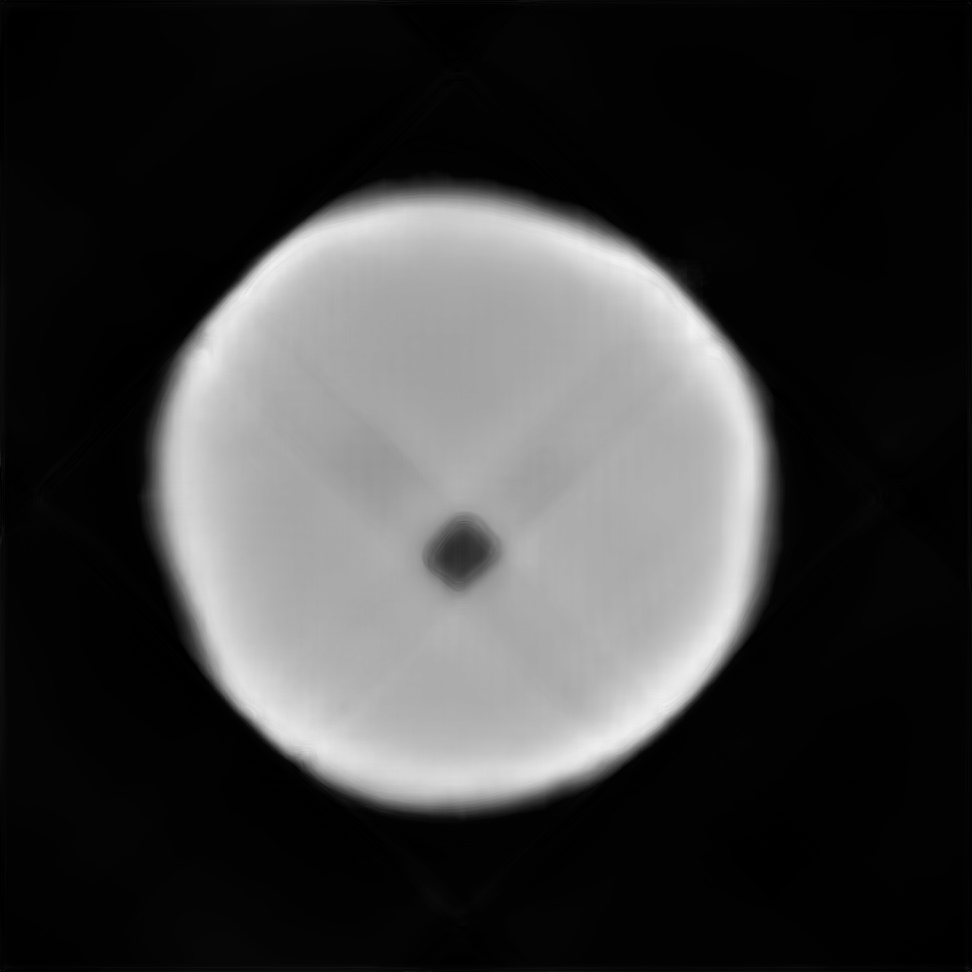} &
                \includegraphics[width=.11\textwidth,valign=t]{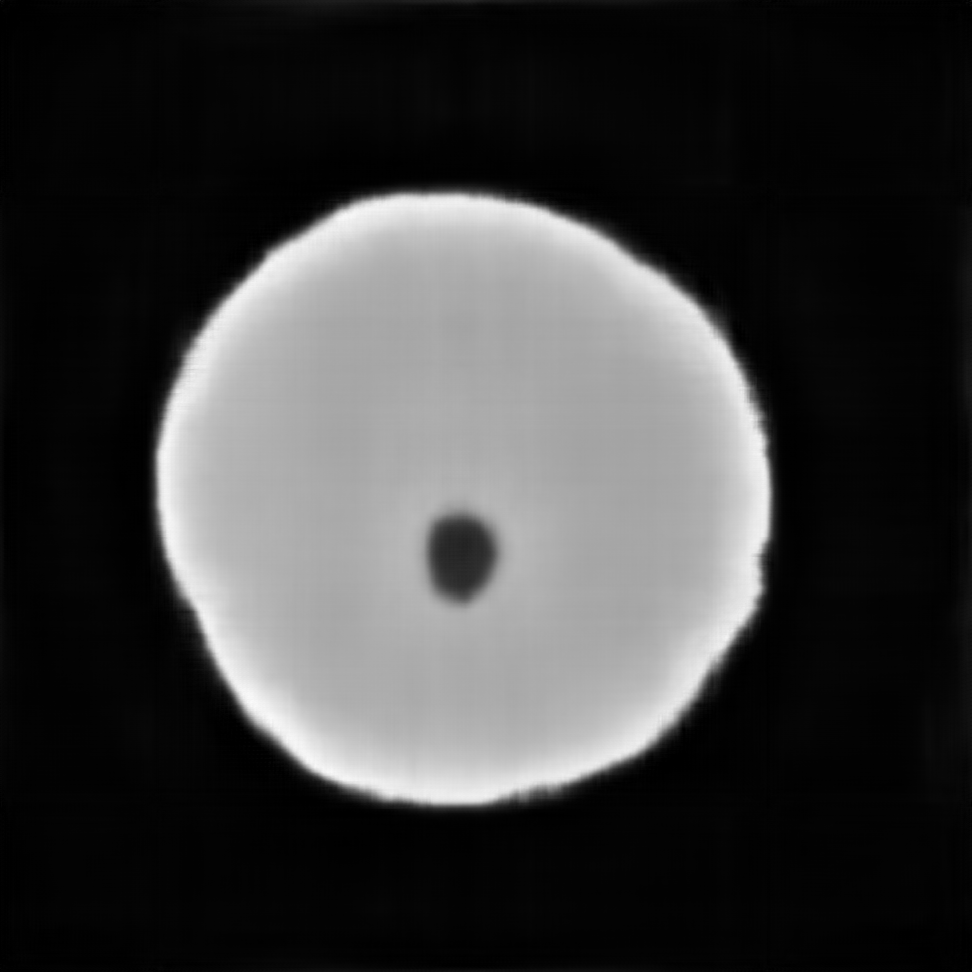}\\
                \scriptsize PSNR~(dB)&
                \scriptsize 13.62&
                \scriptsize 16.03&
                \scriptsize 18.12&
                \scriptsize 21.61&
                \scriptsize 22.29&
                \scriptsize 24.52&
                \scriptsize 22.75&
                \scriptsize 24.49&
                \scriptsize 25.94&
                \scriptsize \textcolor{blue}{26.03}&
                \scriptsize \textcolor{red}{28.12}\\
                \scriptsize /SSIM&
                \scriptsize /0.3308&
                \scriptsize /0.5499&
                \scriptsize /0.6675&
                \scriptsize /\textcolor{blue}{0.7698}&
                \scriptsize /0.6910&
                \scriptsize /0.6792&
                \scriptsize /0.7058&
                \scriptsize /0.6533&
                \scriptsize /0.7476&
                \scriptsize /0.7586&
                \scriptsize /\textcolor{red}{0.7741}\\

                \includegraphics[width=.11\textwidth,valign=t]{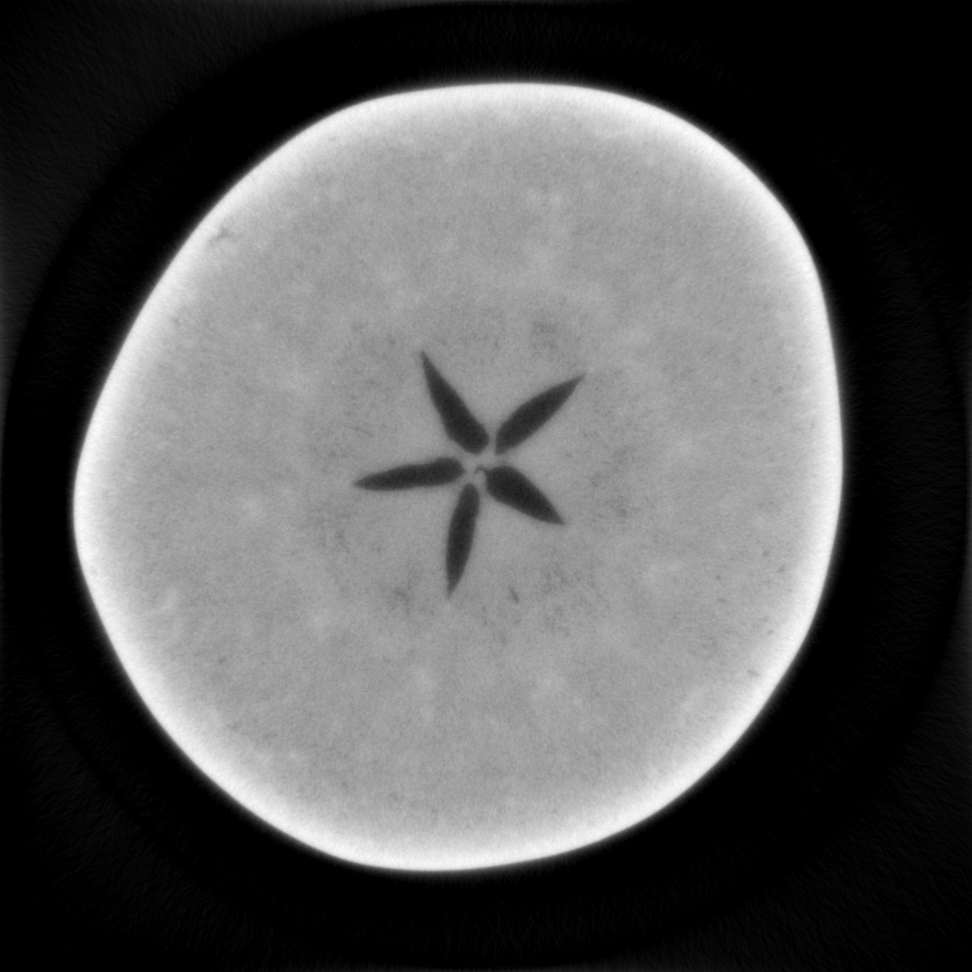} &
                \includegraphics[width=.11\textwidth,valign=t]{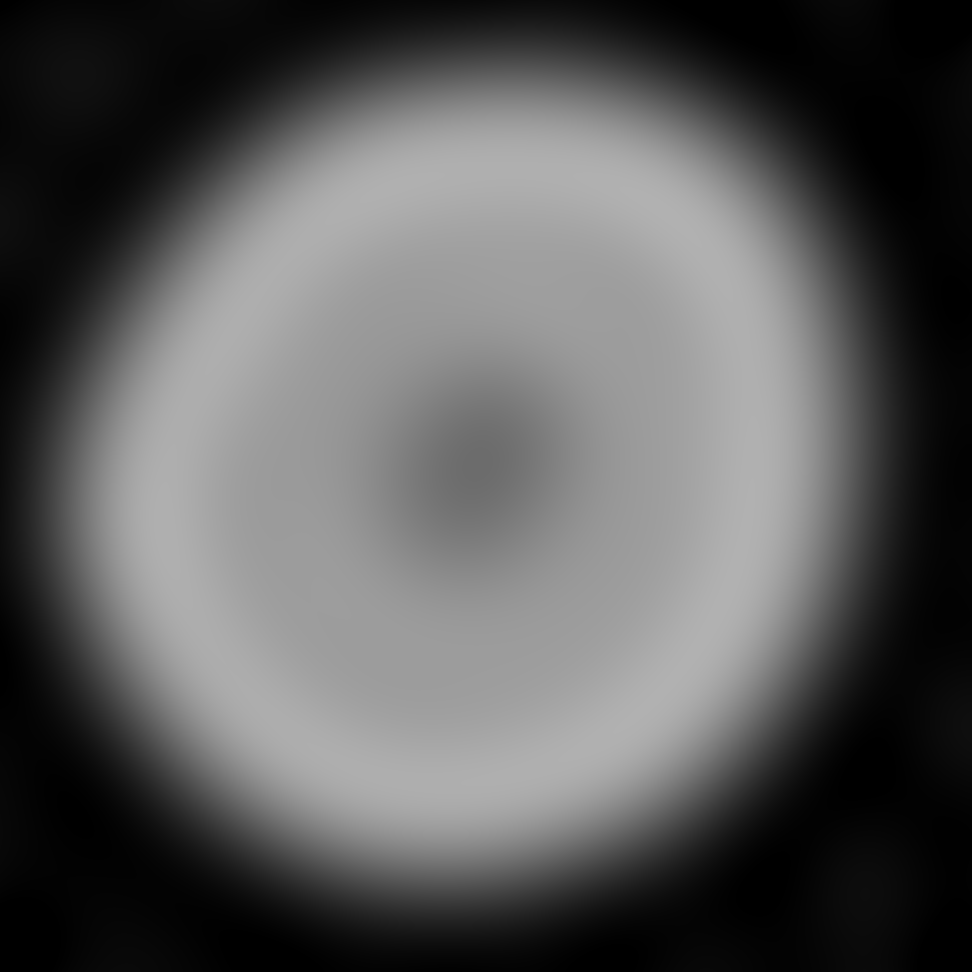} &
                \includegraphics[width=.11\textwidth,valign=t]{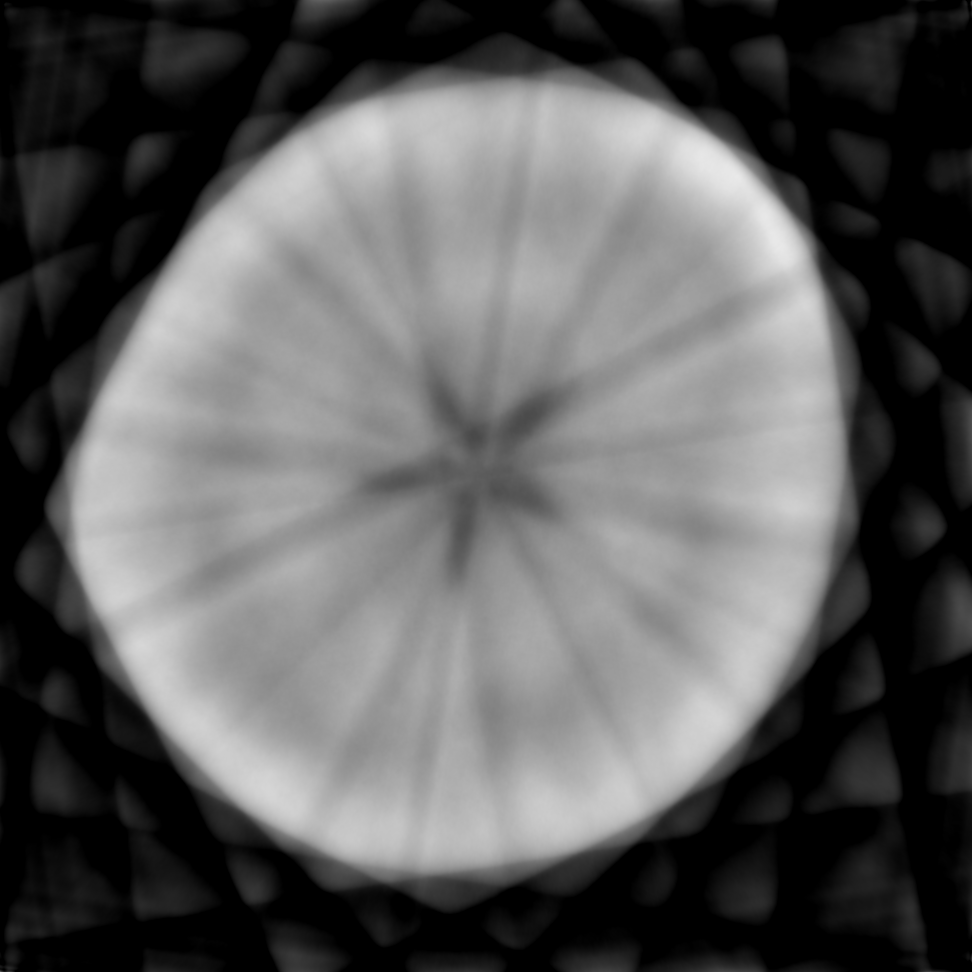} &
                \includegraphics[width=.11\textwidth,valign=t]{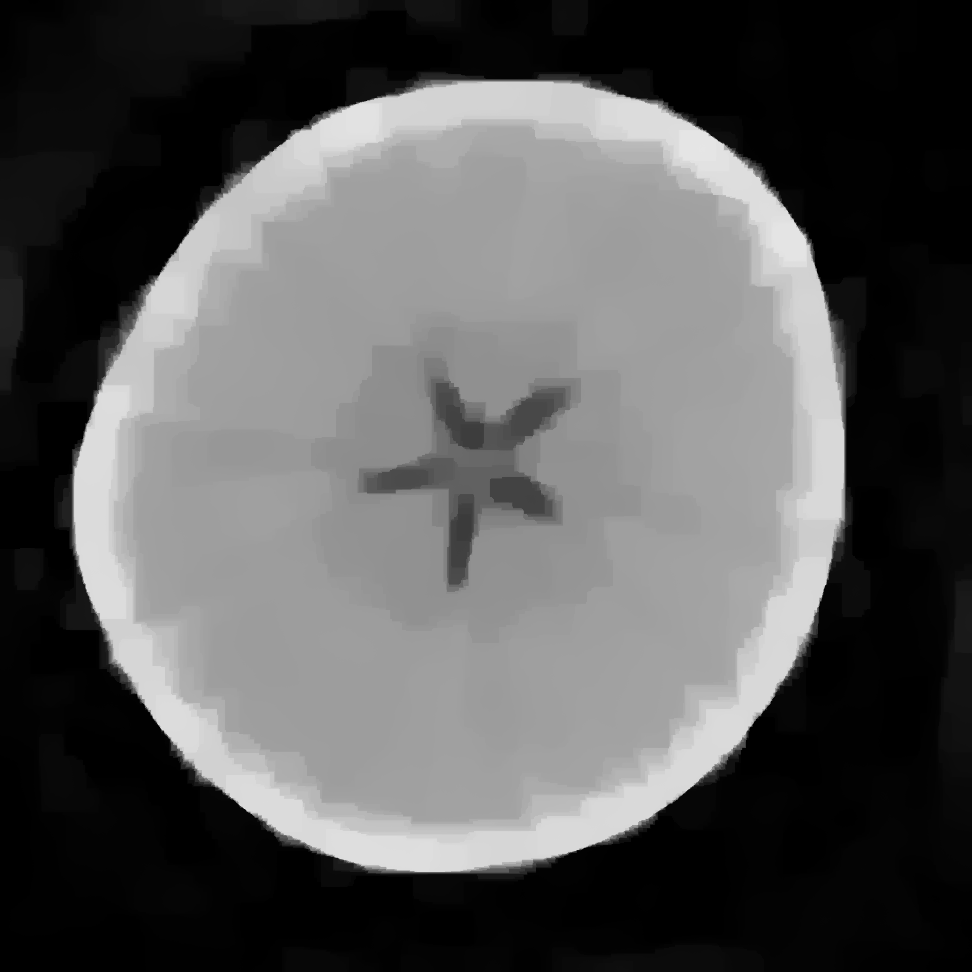} &
                \includegraphics[width=.11\textwidth,valign=t]{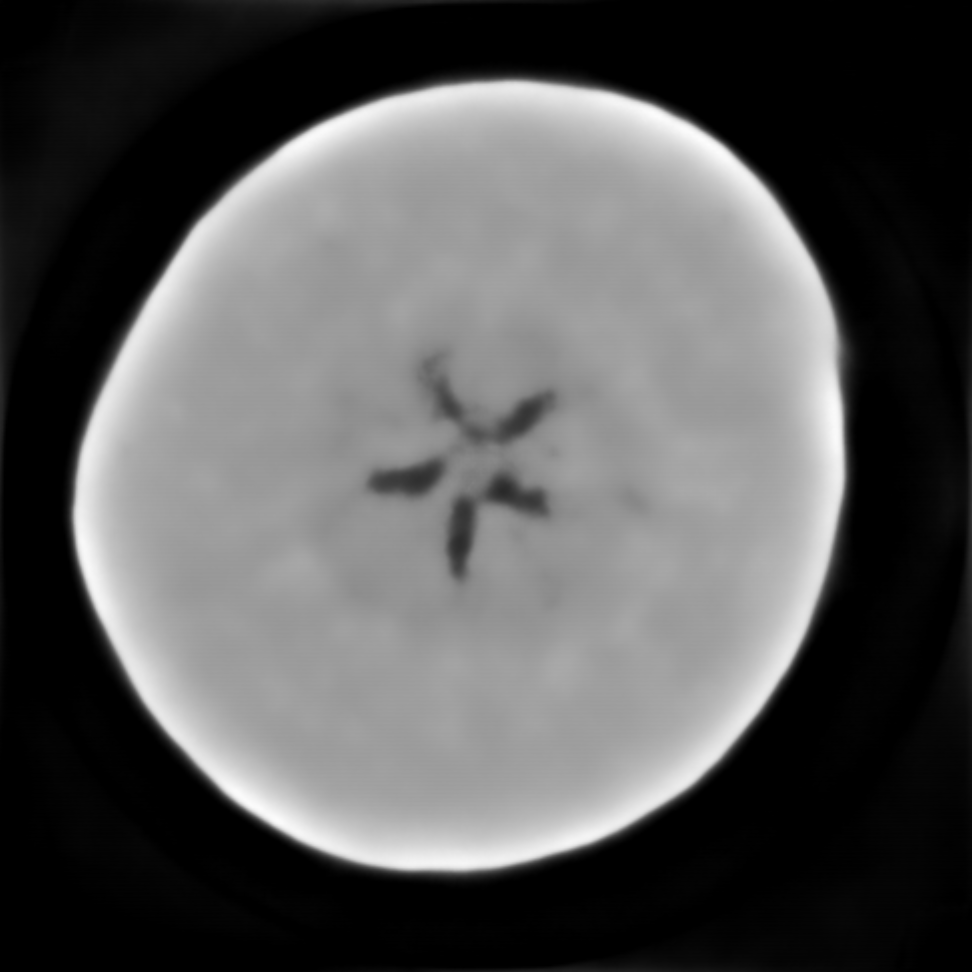} &
                \includegraphics[width=.11\textwidth,valign=t]{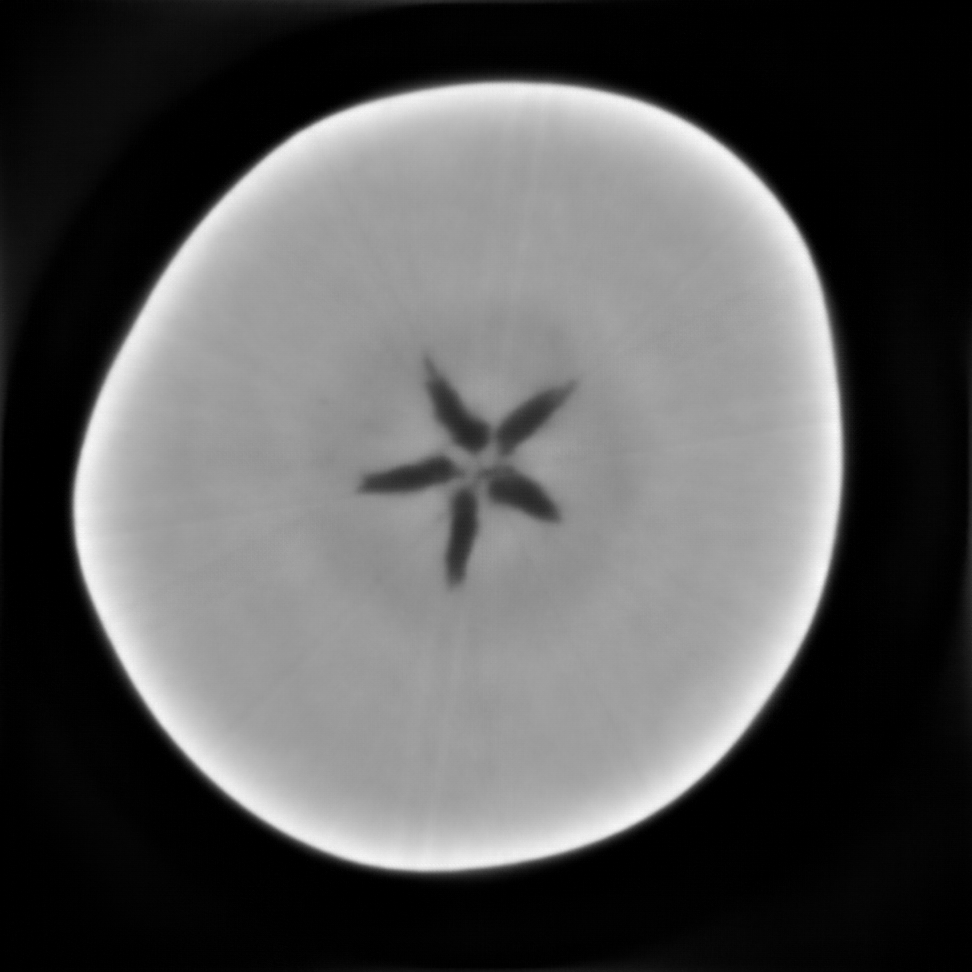} &
                \includegraphics[width=.11\textwidth,valign=t]{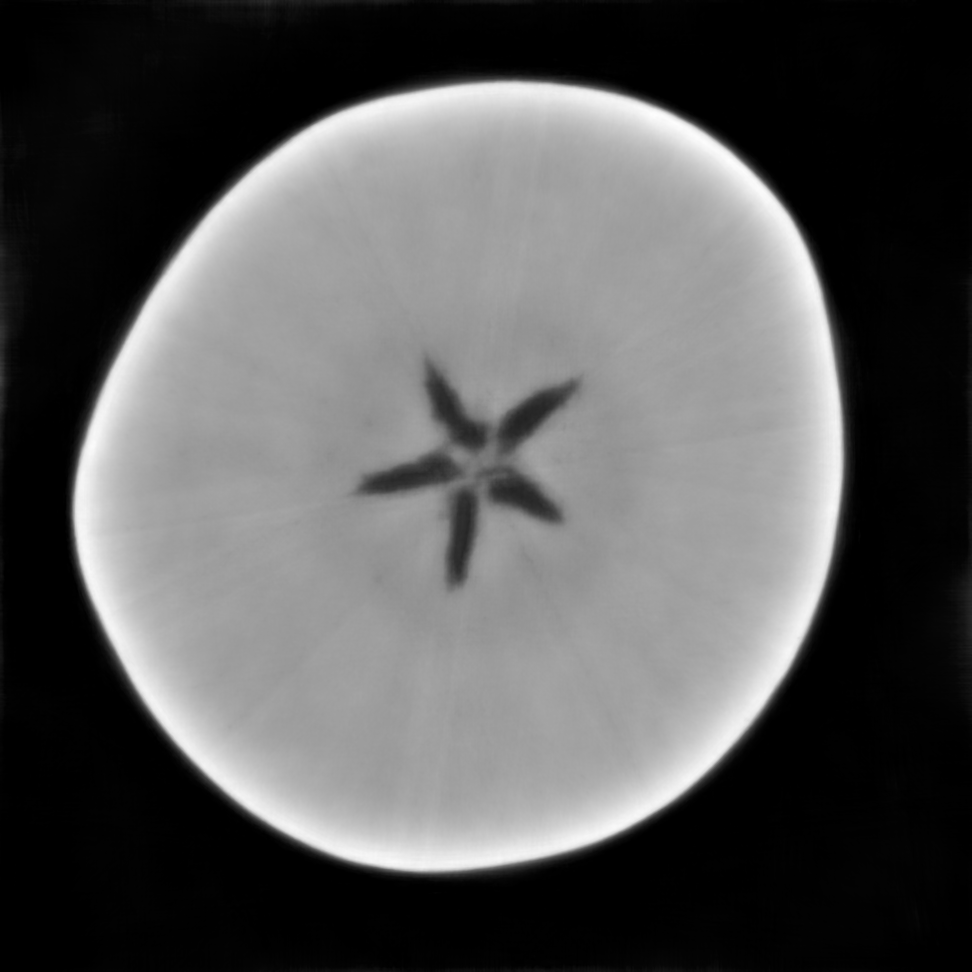} &
                \includegraphics[width=.11\textwidth,valign=t]{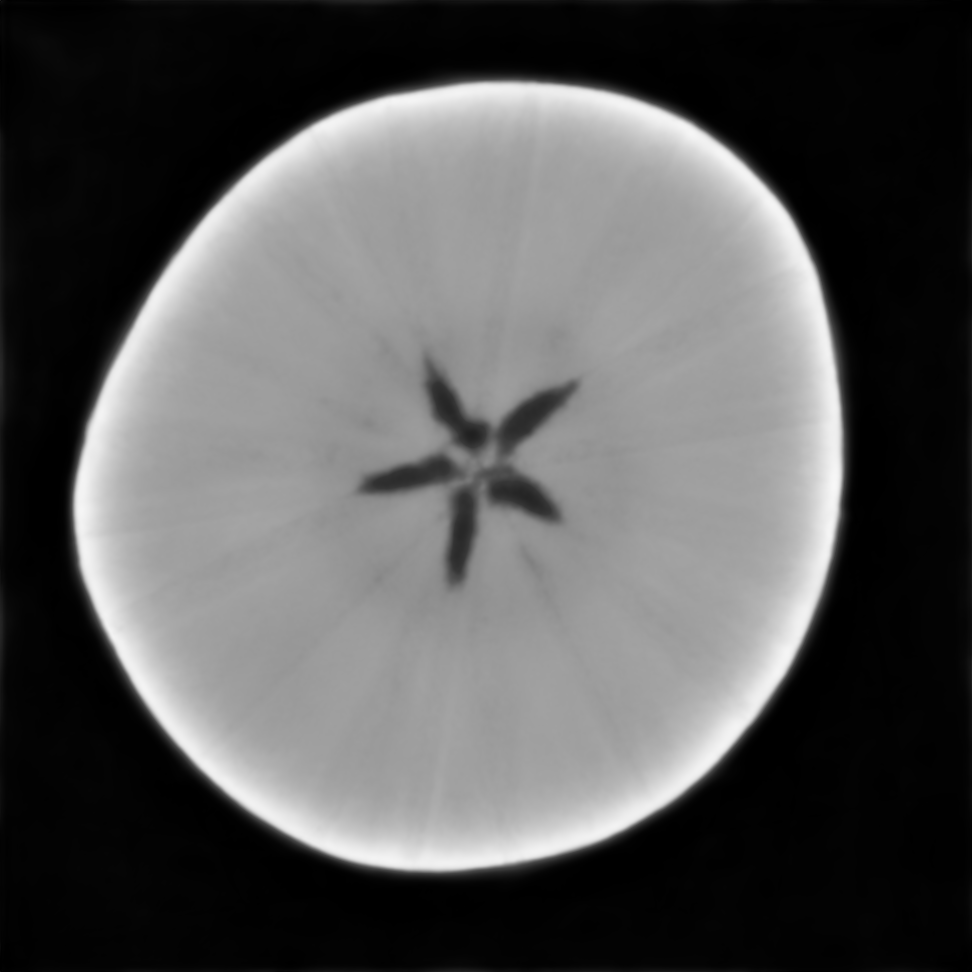} &
                \includegraphics[width=.11\textwidth,valign=t]{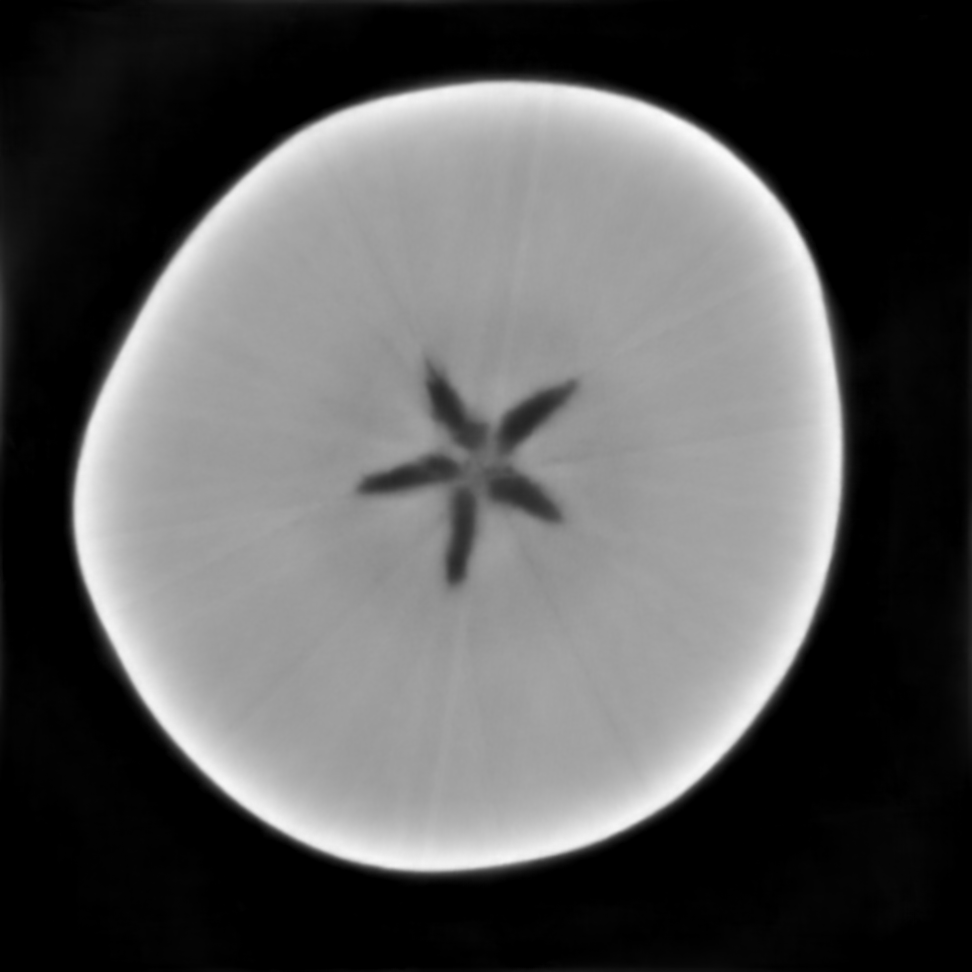} &
                \includegraphics[width=.11\textwidth,valign=t]{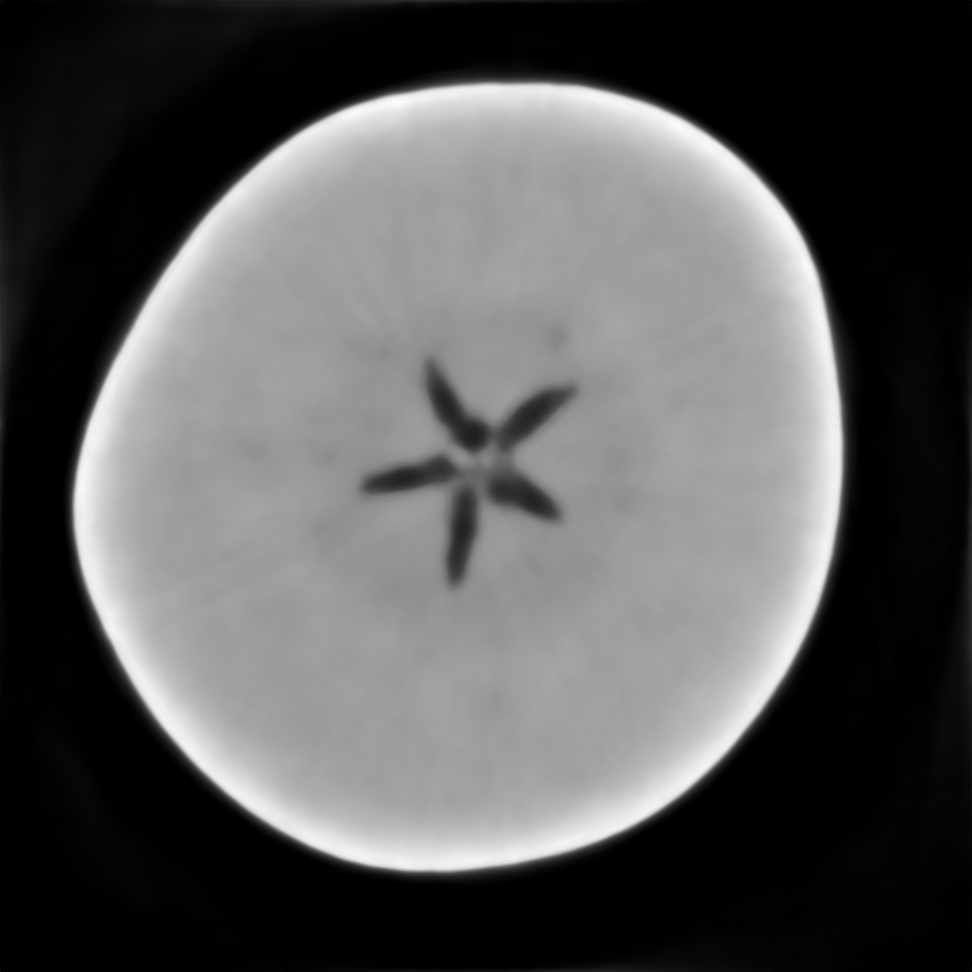} &
                \includegraphics[width=.11\textwidth,valign=t]{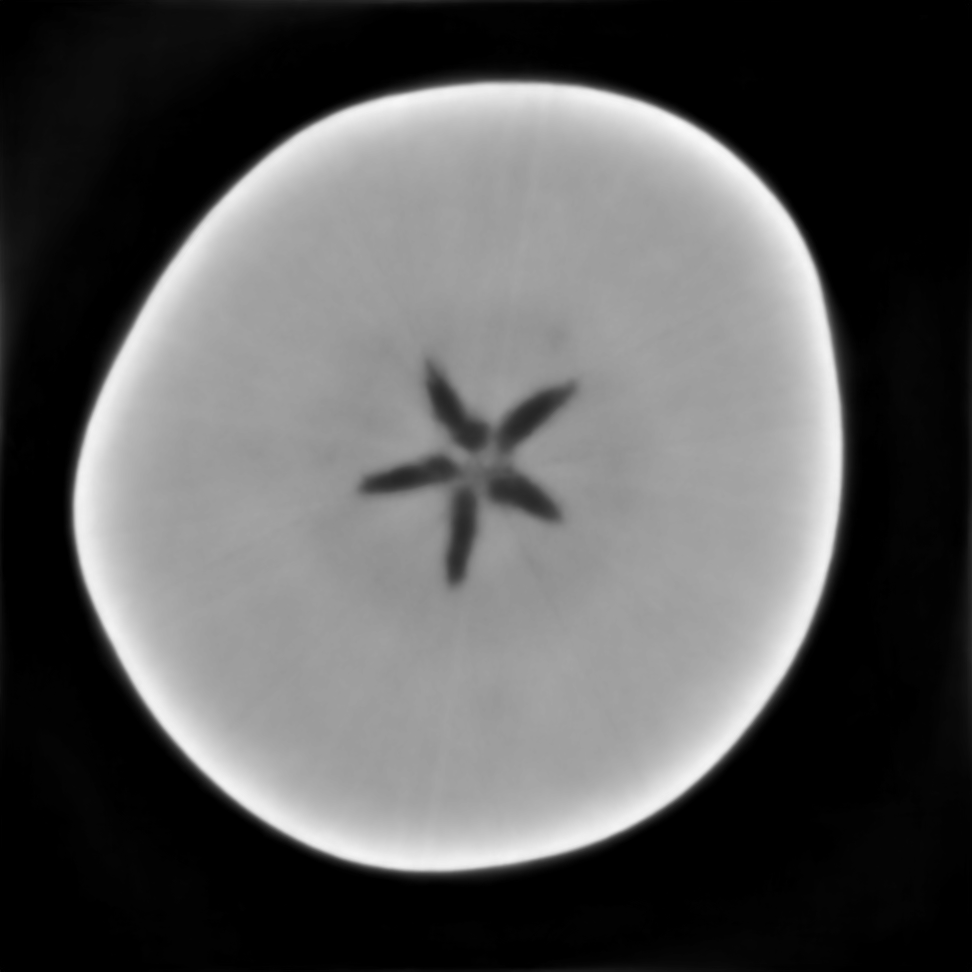} &
                \includegraphics[width=.11\textwidth,valign=t]{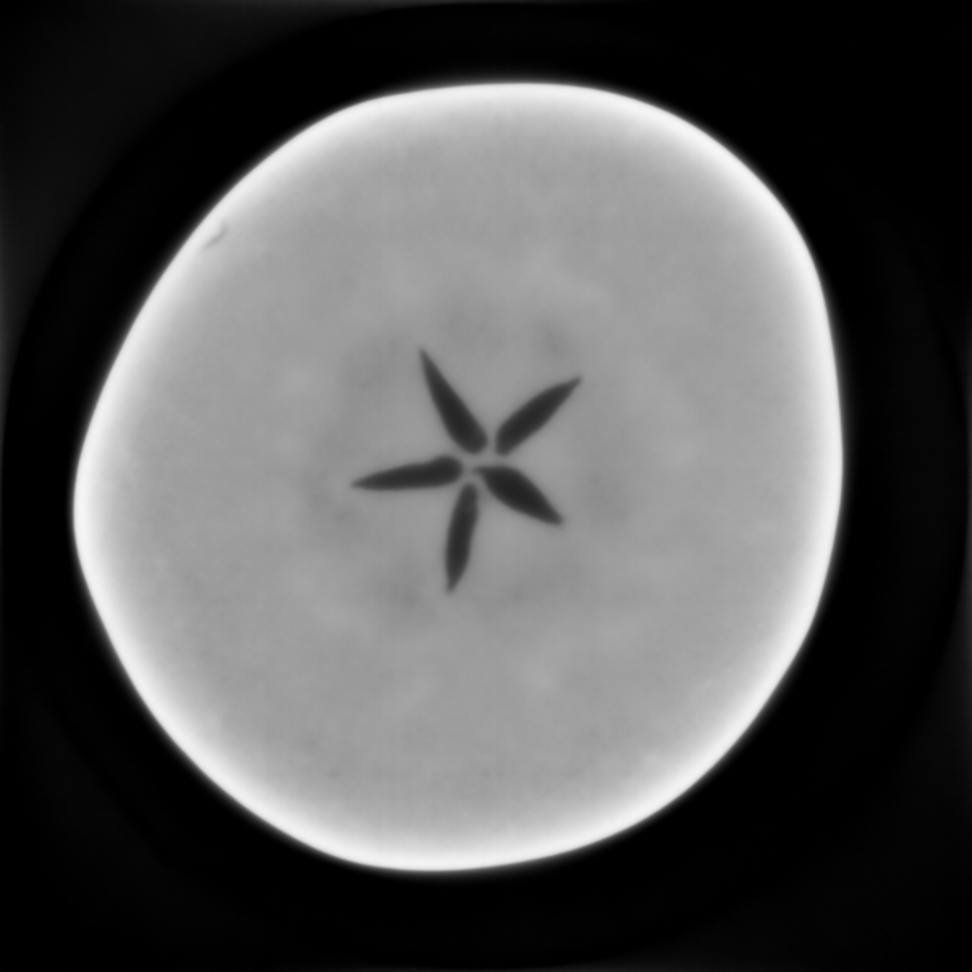}\\
                \scriptsize PSNR~(dB)&
                \scriptsize 16.49&
                \scriptsize 21.16&
                \scriptsize 27.68&
                \scriptsize 29.54&
                \scriptsize 34.25&
                \scriptsize 33.86&
                \scriptsize 33.26&
                \scriptsize 34.03&
                \scriptsize \textcolor{blue}{35.02}&
                \scriptsize 34.95&
                \scriptsize \textcolor{red}{37.32}\\
                \scriptsize /SSIM&
                \scriptsize /0.5503&
                \scriptsize /0.6168&
                \scriptsize /0.7534&
                \scriptsize /0.8587&
                \scriptsize /\textcolor{blue}{0.8751}&
                \scriptsize /0.8235&
                \scriptsize /0.8141&
                \scriptsize /0.8609&
                \scriptsize /0.8694&
                \scriptsize /0.8675&
                \scriptsize /\textcolor{red}{0.9057}\\

                \includegraphics[width=.11\textwidth,valign=t]{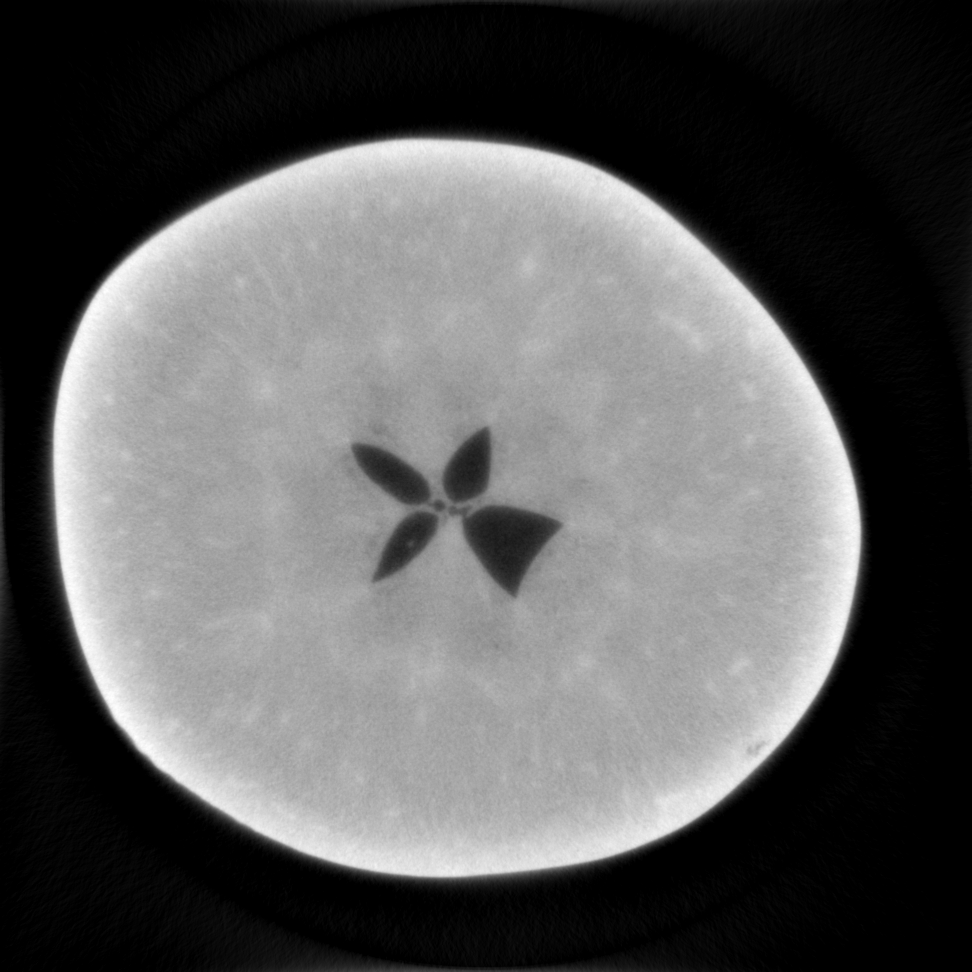} &
                \includegraphics[width=.11\textwidth,valign=t]{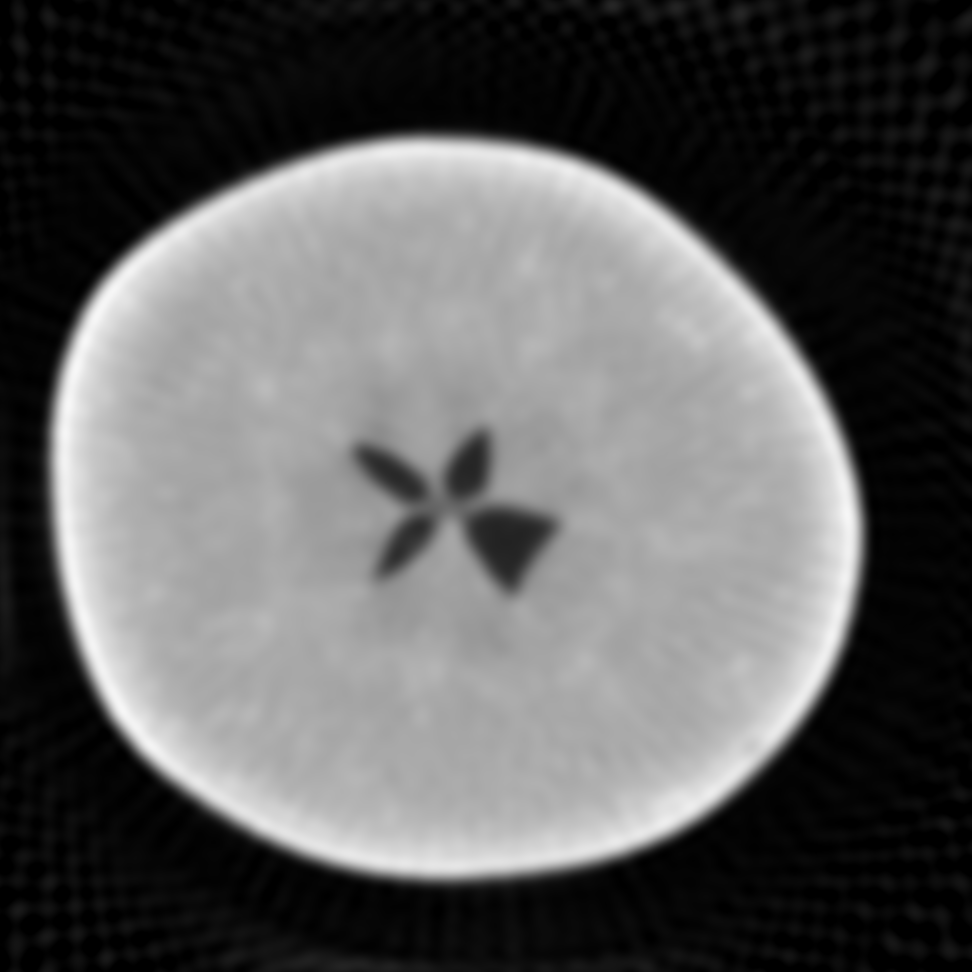} &
                \includegraphics[width=.11\textwidth,valign=t]{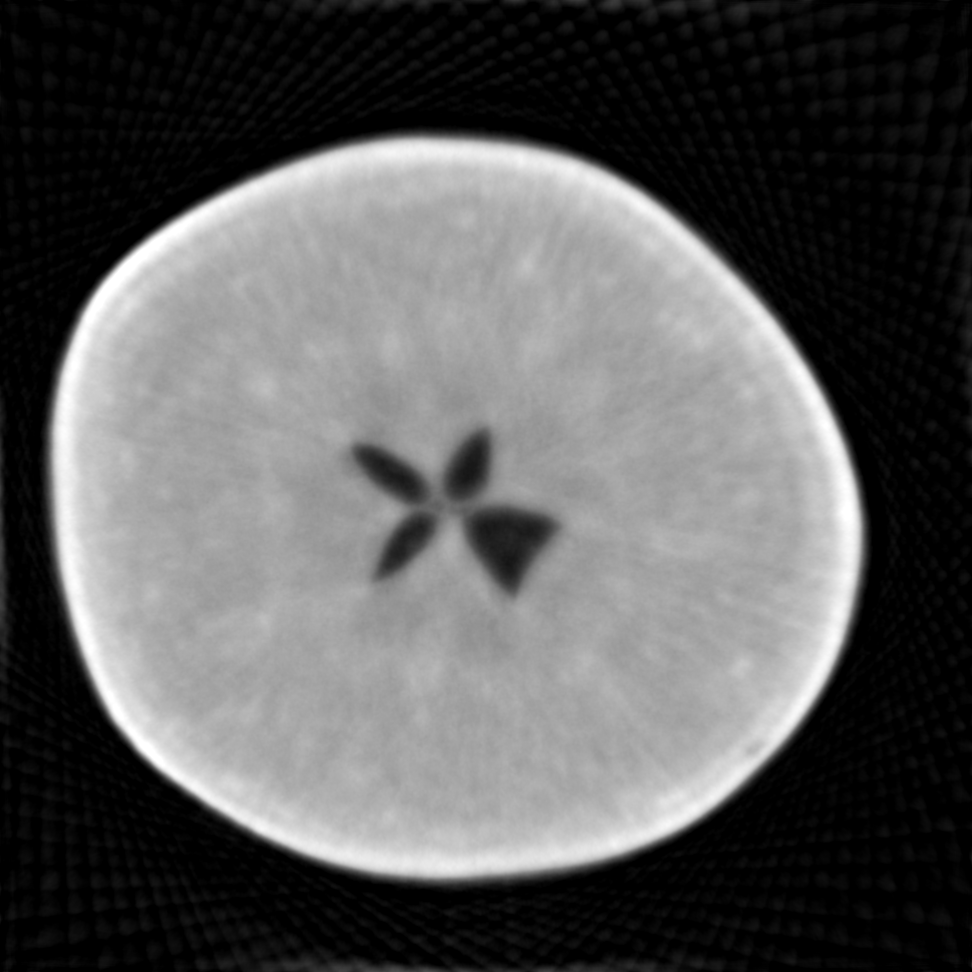} &
                \includegraphics[width=.11\textwidth,valign=t]{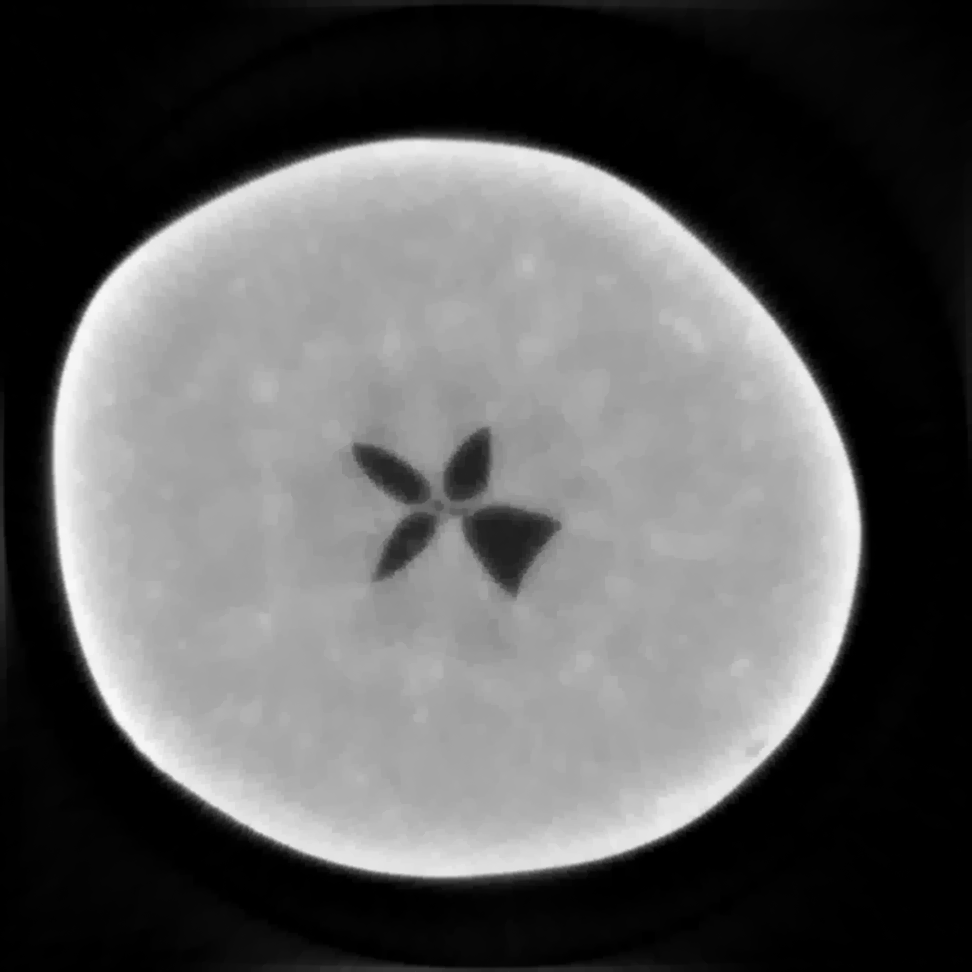} &
                \includegraphics[width=.11\textwidth,valign=t]{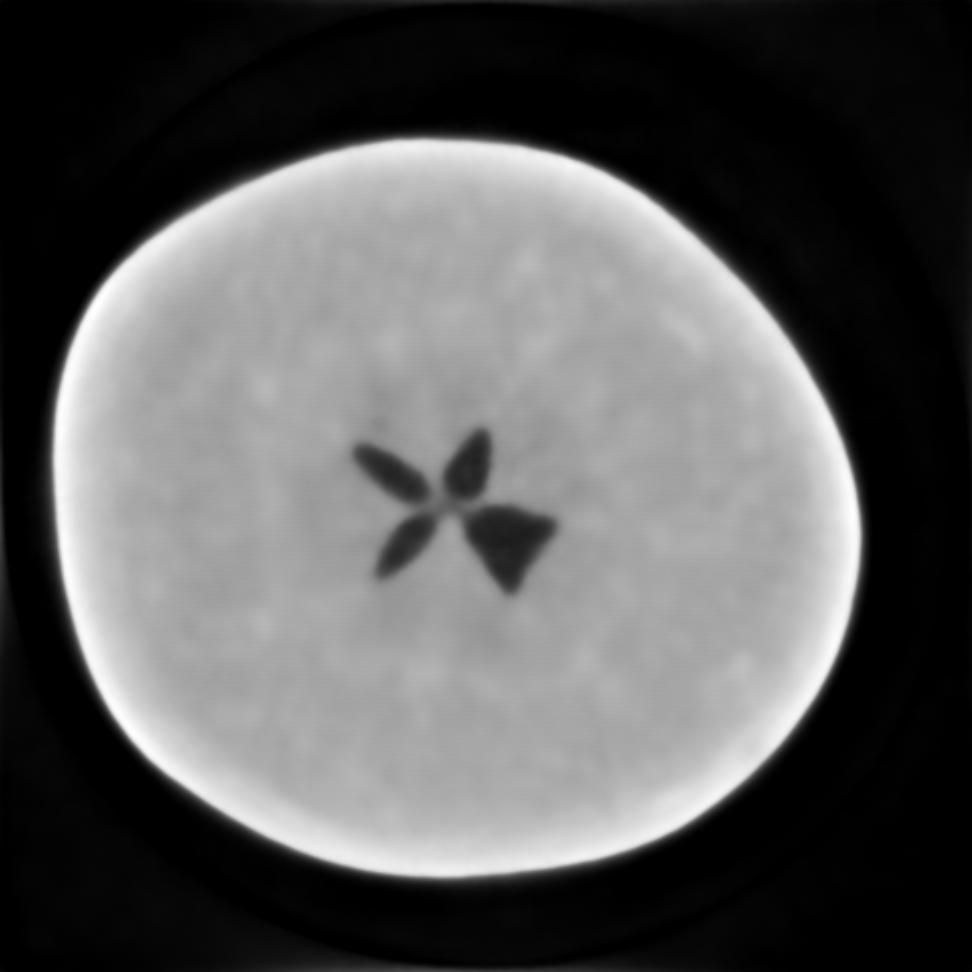} &
                \includegraphics[width=.11\textwidth,valign=t]{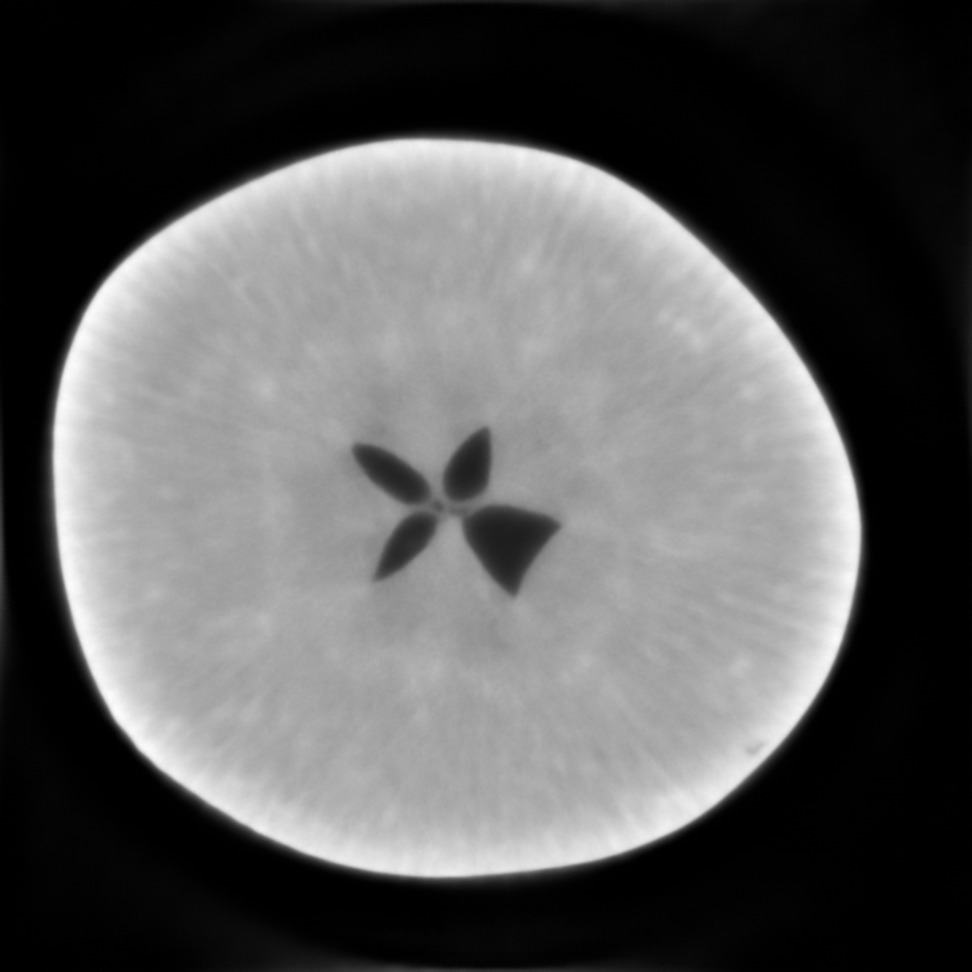} &
                \includegraphics[width=.11\textwidth,valign=t]{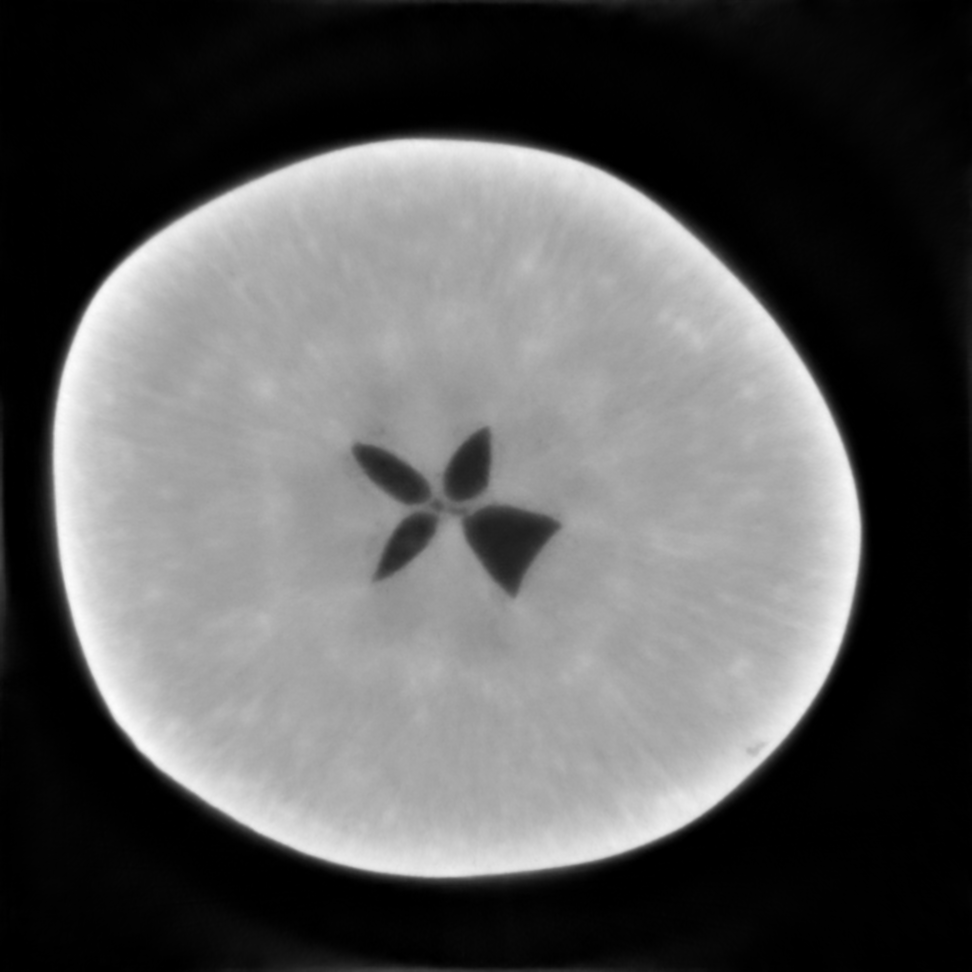} &
                \includegraphics[width=.11\textwidth,valign=t]{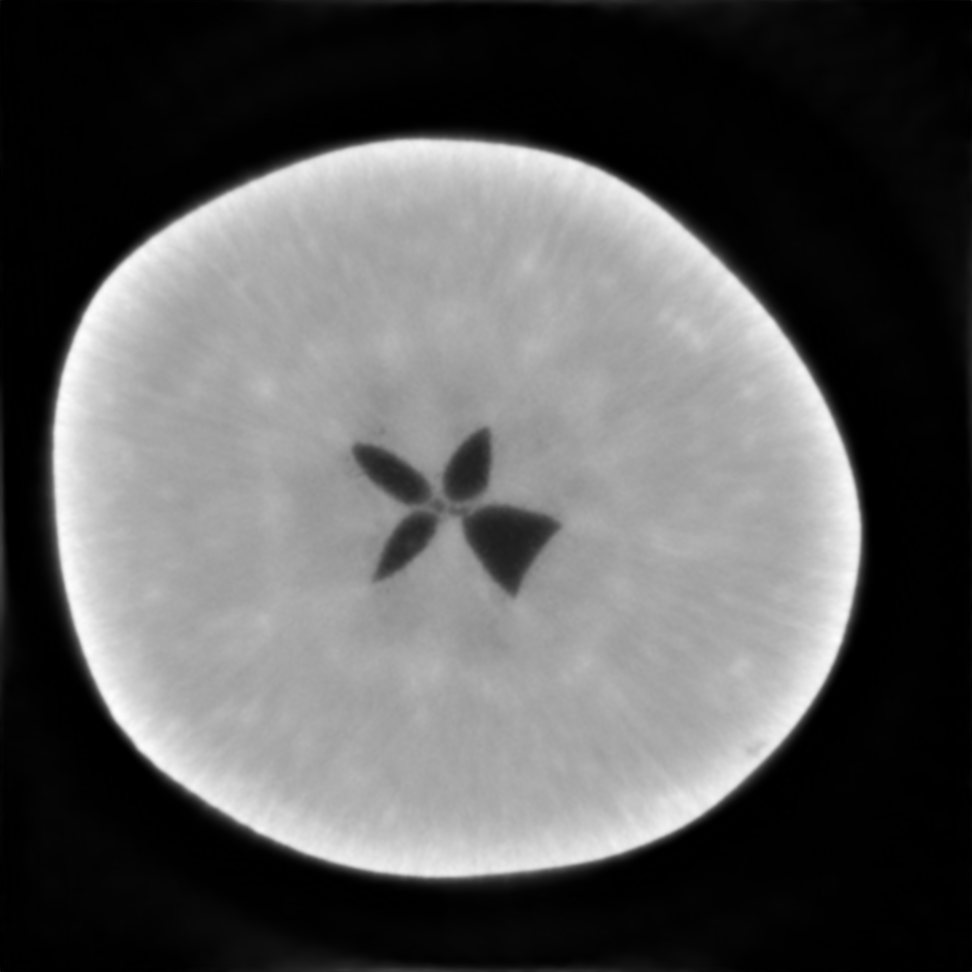} &
                \includegraphics[width=.11\textwidth,valign=t]{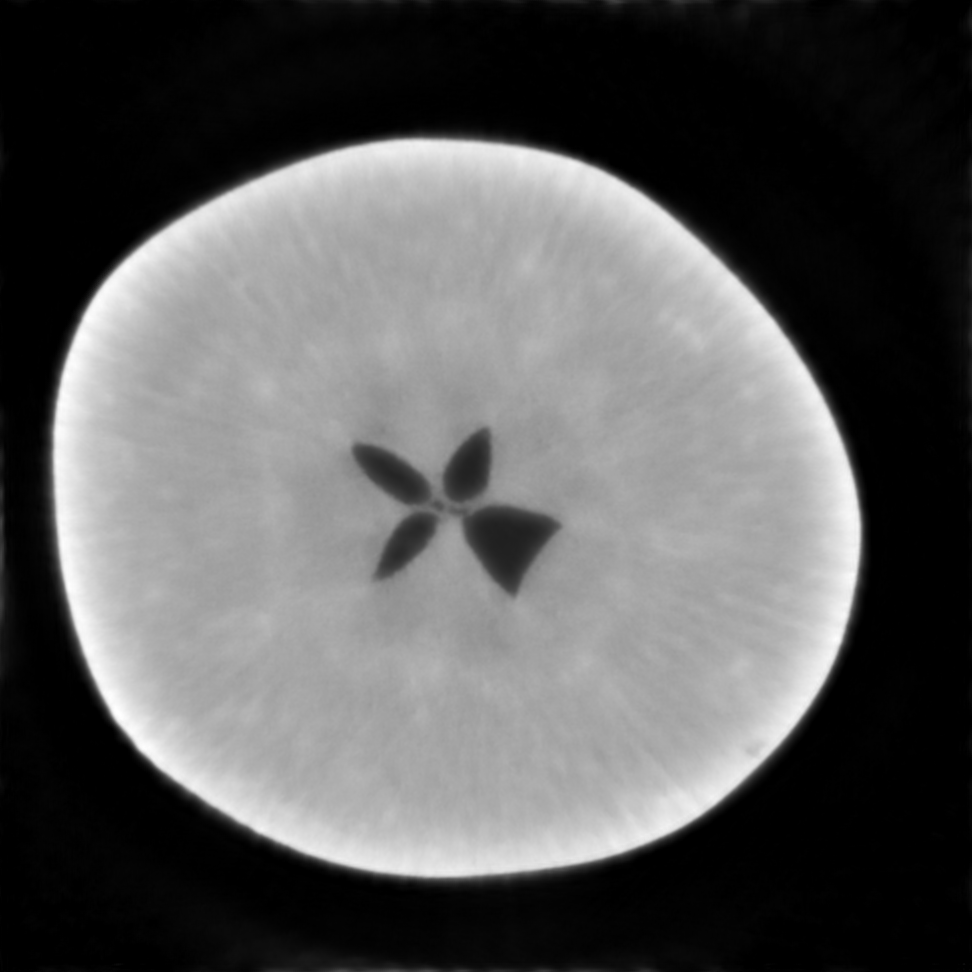} &
                \includegraphics[width=.11\textwidth,valign=t]{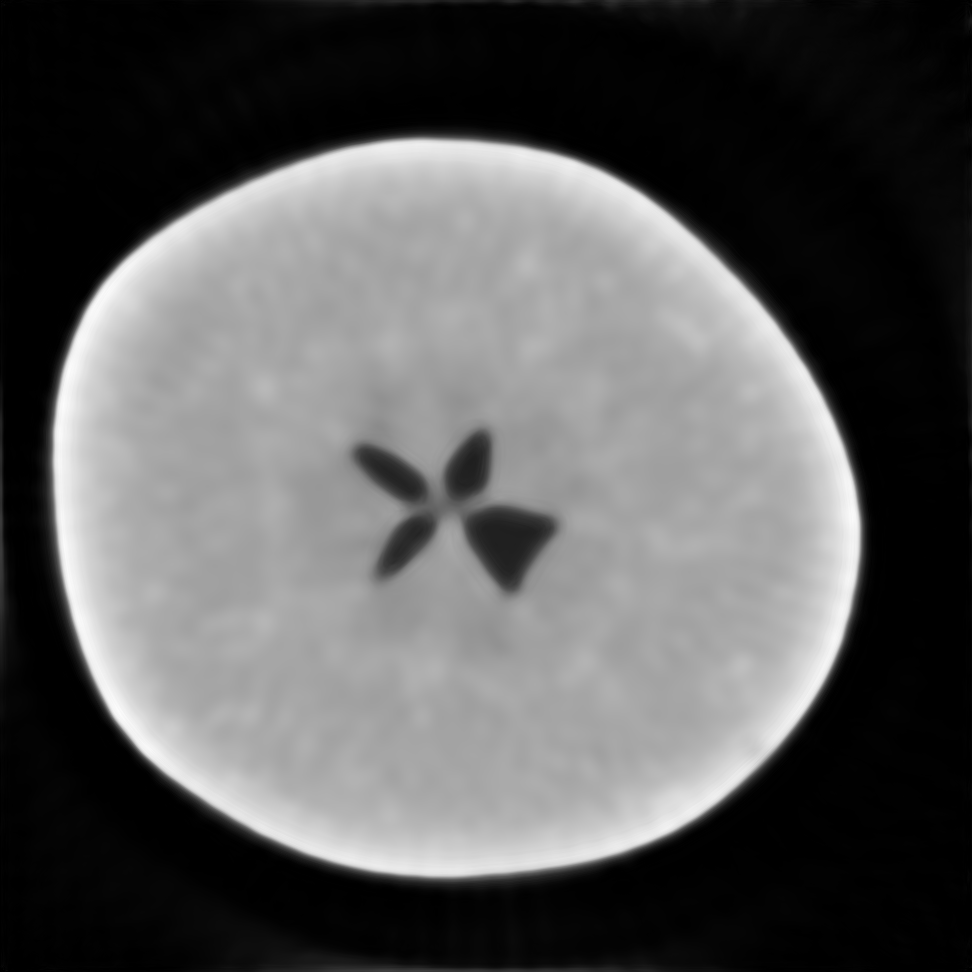} &
                \includegraphics[width=.11\textwidth,valign=t]{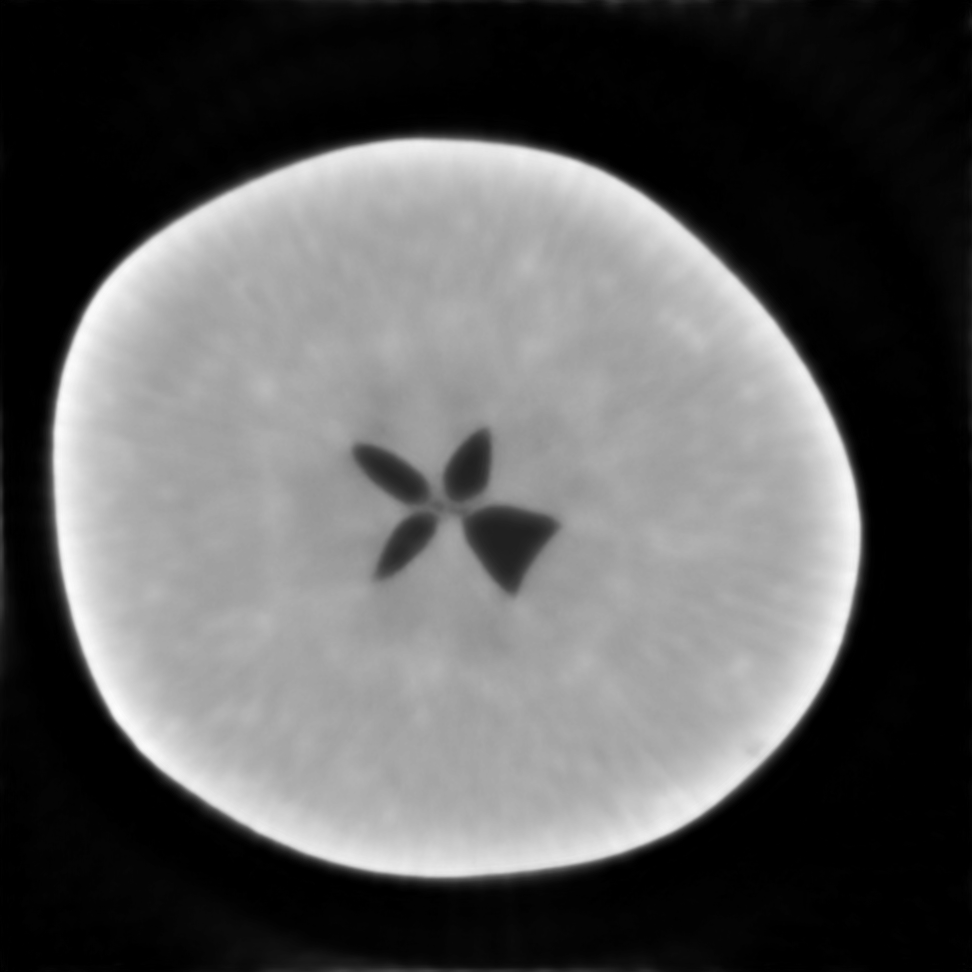} &
                \includegraphics[width=.11\textwidth,valign=t]{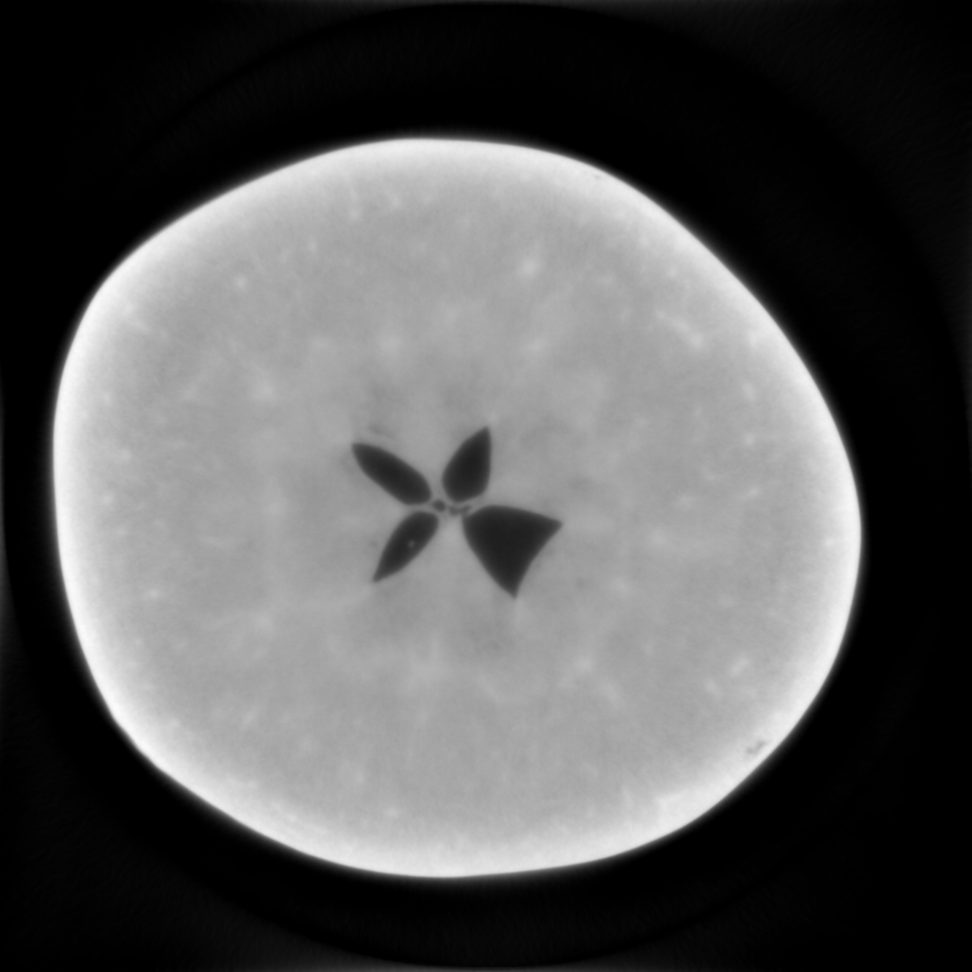}\\
                \scriptsize PSNR~(dB)&
                \scriptsize 29.57&
                \scriptsize 32.18&
                \scriptsize 38.51&
                \scriptsize 36.08&
                \scriptsize 38.73&
                \scriptsize \textcolor{blue}{39.09}&
                \scriptsize 38.67&
                \scriptsize 38.06&
                \scriptsize 37.76&
                \scriptsize 38.32&
                \scriptsize \textcolor{red}{40.16}\\
                \scriptsize /SSIM&
                \scriptsize /0.6919&
                \scriptsize /0.7588&
                \scriptsize /\textcolor{blue}{0.9083}&
                \scriptsize /0.8842&
                \scriptsize /0.9077&
                \scriptsize /0.9066&
                \scriptsize /0.9034&
                \scriptsize /0.8905&
                \scriptsize /0.8911&
                \scriptsize /0.8980&
                \scriptsize /\textcolor{red}{0.9257}\\
		\end{tabular}}
	\end{center}
	\caption{Visual comparisons among ten methods and our PRL-PGD-1 on recovering three images from Apple CT dataset \cite{coban2020apple} in the sparse-view CT reconstruction task with angle numbers of 2 (top), 10 (middle) and 50 (bottom), respectively.}
	\label{fig:CT}
\end{figure*}

\subsection{Extension to the Real-World Sparse-View CT Reconstruction}
\label{sec:CT}
As described above, our core idea of generalizing the physics-guided unrolling to the multiscale full-channel feature recovery is both algorithm- and network block-agnostic. The proposed PRL framework tries to close the gap between deep unrolling paradigm and the generic recovery networks. To further show its extensibility, in addition to Tab.~\ref{tab:boost_existing_duns}, here we extend PRL to the specific problem of sparse-view CT reconstruction and demonstrate that PRL is also forward operator-agnostic.

In sparse-view CT, the forward operator (sampling matrix) is equivalently defined as $\mathbf{A}=\mathbf{MT}$, where $\mathbf{T}$ is the matrix form of the Radon transform \cite{radon1986determination} and $\mathbf{M}$ is a diagonal binary matrix for under-sampling. In this case, we follow the settings of the grand Apples-CT challenge 2020 \cite{leuschner2021quantitative} and evaluate PRL on the Apple CT dataset \cite{coban2020apple}. It collects more than 70000 data slices from 94 real scanned apples. The goal is to reconstruct the original one hundred of 972$\times$972 images from only 2, 10 and 50 angles of measurement projections. Utilizing the same training set and implementation of the forward operator $\mathbf{A}$ as \cite{leuschner2021quantitative}, we adopt the corresponding filtered back-projection operator \cite{radon1986determination} to substitute $\mathbf{A}^\top$ and train a PRL-PGD with $K=1$ (marked by \textbf{PRL-PGD-1}) since we find that this tiny version is competent in this task.

\begin{table*}[!t]
\caption{Quantitative comparison among three state-of-the-art methods \cite{shi2019image,zhang2020optimization,song2021memory} and our PRL-PGD network on the block-diagonal one-bit image CS task with the setting of sampling rate $\gamma =50\%$ and noise level $\sigma =0$.}
\label{tab:one_bit}
\centering
\scalebox{0.85}{\begin{tabular}{c|c|c|c|
>{\columncolor[HTML]{EFEFEF}}c }
\shline
Method   & CSNet$^+$ \cite{shi2019image} & OPINE-Net$^+$ \cite{zhang2020optimization} & MADUN \cite{song2021memory} & \textbf{PRL-PGD} \\ \hline
Average PSNR (dB) on Set11 \cite{kulkarni2016reconnet} & 18.25     & 19.06         & \textcolor{blue}{20.32} & \textcolor{red}{23.78}                             \\ \shline
\end{tabular}}
\end{table*}

We select ten representative sparse-view CT reconstruction methods for comparison: the filtered back-projection (FBP) \cite{radon1986determination}, the conjugate gradient method for least squares (CGLS) \cite{bjorck1998stability}, the total variation (TV) method \cite{niu2014sparse}, the extension of iCT-Net (iCTU-Net) \cite{li2019learning}, the conditional invertible neural network (CINN) \cite{denker2020conditional}, the mixed-scale dense CNN (MS-D-CNN) \cite{pelt2018improving}, U-Net \cite{chen2017low}, the ISTA-inspired U-Net \cite{liu2020interpreting} (5-stage), the learned primal-dual algorithm (Learned P. D.) \cite{adler2018learned} (10-stage), and FISTA-Net \cite{xiang2021fista} (7-stage). Tab.~\ref{tab:CT} summarizes the recovery PSNR/SSIM scores, the levels of average training time cost and parameter numbers of various methods. Fig.~\ref{fig:CT} provides three reconstruction instances under the three different angle number settings. We can observe that when the angle number is 2, the proposed PRL-PGD-1 has a significant performance leading over other methods since it better absorbs the prior knowledge of image and physics in multiple scales. When the angle number is larger than 2, our PRL-PGD-1 consistently recovers more sharp details than other methods, especially in the center star pattern of the apple core. The physics-engaged networks including the ISTA U-Net, Learned P. D., FISTA-Net and PRL-PGD-1 keep the accurate shapes while the recovered results of other physics-free ones suffer from some apparent distortions. It is clear that our PRL-based network outperforms the others not only in terms of the accuracy indices but also in terms of the training time and efficiency of learnable parameters.

\subsection{Extension to the One-Bit Compressed Sensing Task}
\label{sec:one_bit}
To further validate the effectiveness and generalization ability of our PRL network, here we apply our proposed PRL-PGD variant to the one-bit CS task, which refers to a classical non-linear inverse problem of reconstructing the original image signal from the signs of its linear measurements \cite{boufounos20081}. The observation model of one-bit CS is $\mathbf{y}=\text{sign}(\mathbf{Ax+\epsilon})$, where $\text{sign}(\cdot)$ is an element-wise function that maps non-negative and negative numbers to +1 and -1, respectively. Following the setup in \cite{jacques2013robust,kafle2021one}, we apply CSNet$^+$ \cite{shi2019image}, OPINE-Net$^+$ \cite{zhang2020optimization} (9-stage), MADUN \cite{song2021memory} (25-stage) and our PRL-PGD to block-diagonal one-bit image CS task. Specifically, based on the default settings of all the four networks, we use $\text{sign}(\mathbf{A}(\cdot))$ to implement the forward operator and define the derivative of $\text{sign}(\cdot)$ as one \cite{zhang2020optimization} to enable its differentiability in the back-propagation of training. Tab.~\ref{tab:one_bit} reports the evaluation results of the four networks with $\gamma =50\%$ and shows the PSNR leading of 3.46dB of PRL-PGD to existing best image CS methods, thus manifesting the consistent performance and robustness superiority, and generalization ability of our method to the non-linear inverse imaging problems.

\begin{table*}
\caption{High-level conceptual and functional classifications and qualitative comparisons among six representative CS networks \cite{shi2019image,shi2019scalable,zhang2020optimization,zhang2020amp,you2021coast,song2021memory} and the proposed \textbf{PRL} framework.}
\label{tab:cla_comp}
\centering
\hspace{-6pt}\scalebox{0.675}{
\begin{tabular}{c|c|c|c|c|c|c|c|c|c|c}
\shline
Category& Method& \begin{tabular}[c]{@{}c@{}}Recovery\\ Domain\end{tabular} & \begin{tabular}[c]{@{}c@{}}High-Throughput\\ Transmission\end{tabular} & \begin{tabular}[c]{@{}c@{}}Multiscale\\ Perception\end{tabular} & \begin{tabular}[c]{@{}c@{}}Interpre\\ -tability\end{tabular} & \begin{tabular}[c]{@{}c@{}}Unrolled\\ Framework\end{tabular} & \begin{tabular}[c]{@{}c@{}}Physics\\ Utilization\end{tabular}&Accuracy&Speed&Extensibility\\ \hline
\multirow{2}{*}{\begin{tabular}[c]{@{}c@{}}Physics\\ -Free\end{tabular}}&CSNet$^+$&\multirow{2}{*}{FD}&\multirow{2}{*}{$\checkmark$}&\multirow{7}{*}{$\times$}&\multirow{2}{*}{$\times$}& \multirow{2}{*}{-}  &\multirow{1}{*}{\begin{tabular}[c]{@{}c@{}}\ding{72}\\ (initialization)\end{tabular}} & \multirow{2}{*}{\ding{72}}& \ding{72}\ding{72}\ding{72}\ding{72} & \multirow{2}{*}{\ding{72}}\\ \hhline{~-~~~~~~~-~}
&SCSNet &&&&&&&&\ding{72}\ding{72}\ding{72}&\\ \hhline{----~------} \multirow{5}{*}{\begin{tabular}[c]{@{}c@{}}Physics\\-Engaged\end{tabular}} & \begin{tabular}[c]{@{}c@{}}OPINE\\-Net$^+$\end{tabular} & \multirow{4}{*}{ID}& \multirow{3}{*}{$\times$} && \multirow{7}{*}{\textcolor{red}{$\checkmark$}}& ISTA & \multirow{4}{*}{\begin{tabular}[c]{@{}c@{}}\ding{72}\ding{72}\\ (single-channel\\and single-scale\\utilization)\end{tabular}}& \multirow{2}{*}{\ding{72}\ding{72}}& \ding{72}\ding{72}\ding{72}\ding{72} & \multirow{4}{*}{\ding{72}\ding{72}} \\ \hhline{~-~~~~-~~-~}& AMP-Net&&&&& AMP&&& \ding{72}\ding{72}\ding{72}& \\ \hhline{~-~~~~-~--} & COAST &&&&& \multirow{2}{*}{PGD}&& \multirow{2}{*}{\ding{72}\ding{72}\ding{72}}& \ding{72}\ding{72}&\\ \hhline{~-~-~~~~~-~}&MADUN&& \multirow{4}{*}{\textcolor{red}{$\checkmark$}}&&&&&& \ding{72}&\\  \hhline{~--~-~-----}& \begin{tabular}[c]{@{}c@{}}\textbf{PRL}\\(Ours)\end{tabular}& \textcolor{red}{FD}&& \textcolor{red}{$\checkmark$}&& \begin{tabular}[c]{@{}c@{}}\textcolor{red}{PGD}\\\textcolor{red}{/RND}\end{tabular}& \begin{tabular}[c]{@{}c@{}}\textcolor{red}{\ding{72}\ding{72}\ding{72}\ding{72}}\\ \textcolor{red}{(full-channel}\\\textcolor{red}{and multiscale} \\\textcolor{red}{injection)}\end{tabular} & \textcolor{red}{\ding{72}\ding{72}\ding{72}\ding{72}} & \textcolor{red}{\ding{72}\ding{72}\ding{72}}& \textcolor{red}{\ding{72}\ding{72}\ding{72}\ding{72}} \\ \shline
\end{tabular}}
\end{table*}

\section{Conclusion}
A novel generalized deep physics-guided unrolled recovery learning framework named PRL is proposed for image CS via exploiting the network boosting potential. By extending the optimization and range-nullspace decomposition paradigms from the trivial image domain (ID) to the multiscale feature domain (FD), PRL handles the degradation problem and insufficient physical knowledge utilization in previous networks, and breaks through the traditional deep unrolling principles based on the plain single-scale/path feed-forward physics-engaged architectures via embedding the reconstruction model into a 3-scale U-shaped architecture with FD generalization of physical operators for improving the network flexibility and scalability. Beyond the commonly adopted algorithm-specific network-based methods, we show that physics guidance is quite important. Instead, the choice of the unrolling framework has relatively minor impacts on recovery performance. Two implementations: PRL-PGD and PRL-RND are interpretably established and manifest their high efficiency and robustness in keeping their significant performance leading to state-of-the-arts with real-time inference speeds. Our extension to the real-world sparse-view CT and one-bit CS tasks validate that PRL is an extensible algorithm- and physics-agnostic framework, and is appealing for practical applications.

\textbf{Discussions about Limitations:} Although the PRL networks obtain the current best results and are extensible to other inverse imaging problems, their current versions are trying to explore the improvement space of the conventional convolutional recovery network beyond deep unrolling. However, they are hindered by the large-scale parameters and the data bias with a heavy burden for mobile or medical deployments. This could be reduced by some advanced designs or appropriate performance-overhead trade-offs. In addition, the traditional image-domain deep unrolling scheme has obtained good performance-interpretability balances. Our method takes steps forward in CS by further making the network mappings more physics
utilization-efficient, flexible and general (not limited to the hand-crafted analytic forms, specific algorithms or strictly iterative architecture), while keeping the mapping processes being benefited from the well-studied and solid framework bases, thus achieving better
efficiency-interpretability trade-offs. However, the actions of our unrolled neural modules (\textit{e.g.} $\mathcal{H}^{(k)}_\text{grad}$) may lack the strong guarantees to meet their corresponding original purposes (\textit{e.g.} to enforce the data fidelity or sampling consistency). These issues warrant further research.

\begin{appendices}

\section{More Comparison Results}
\label{appxA}
\subsection{High-Level Comparisons of Deep CS Networks}
Here we provide more conceptual and functional classifications, and qualitative comparisons among existing CS networks and our PRL in Tab.~\ref{tab:cla_comp}. We can see that the PRL not only exhibits its organic absorption of the merits of optimization algorithms and advanced network structures but also breaks through the bottlenecks including performance degradation, saturation and the surge of time cost with a more general and compact FD physical framework beyond ID unrolling. It is expected and verified to better approximate the theoretically ideal target reconstruction functions.

\subsection{More Comparisons on the CBSD68 and DIV2K Benchmarks}
\begin{figure}
	\begin{center}
		\hspace{-6pt}\scalebox{0.542}{
			\begin{tabular}[b]{c@{ } c@{ }  c@{ } c@{ } c@{ }}
				\multirow{4}{*}{\includegraphics[trim={56 0 136 0 },clip,width=.312\textwidth,valign=t]{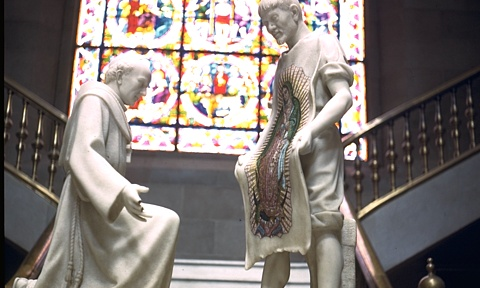}} &   
				\includegraphics[trim={250 120 210 148 },clip,width=.13\textwidth,valign=t]{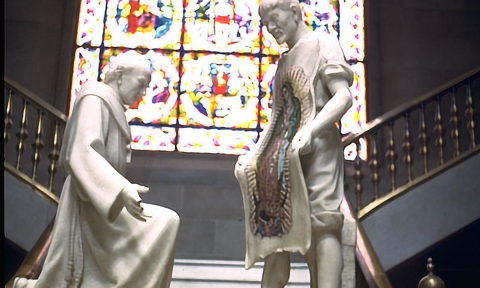}&
				\includegraphics[trim={250 120 210 148 },clip,width=.13\textwidth,valign=t]{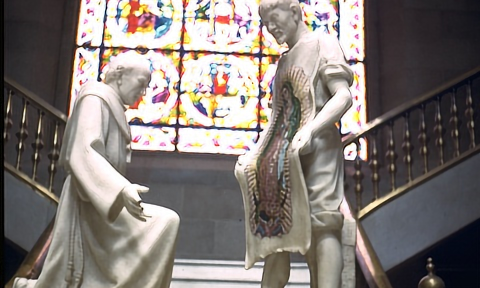}&   
				\includegraphics[trim={250 120 210 148 },clip,width=.13\textwidth,valign=t]{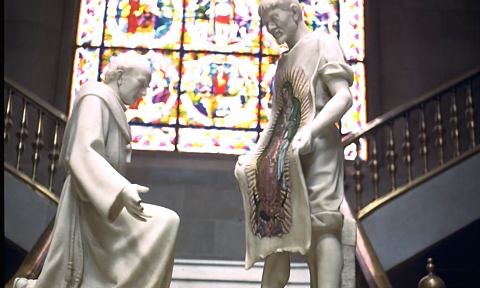}&
				\includegraphics[trim={250 120 210 148 },clip,width=.13\textwidth,valign=t]{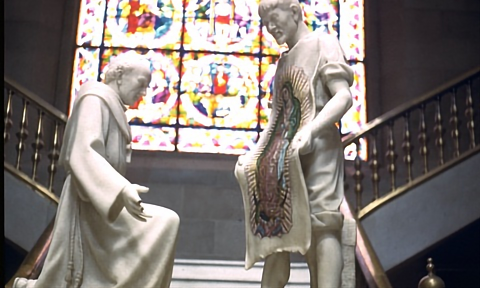}
				
				\\
				&  \scriptsize~31.59/0.9607 &\scriptsize~31.43/0.9623  & \scriptsize~34.91/0.9781& \scriptsize~35.19/0.9781\\
				& \scriptsize~CSNet$^+$& \scriptsize~SCSNet& \scriptsize~OPINE-Net$^+$& \scriptsize~AMP-Net\\
				
				&
				\includegraphics[trim={250 120 210 148 },clip,width=.13\textwidth,valign=t]{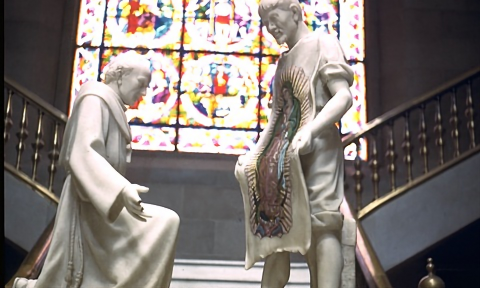}&
				\includegraphics[trim={251 136 210 165 },clip,width=.13\textwidth,valign=t]{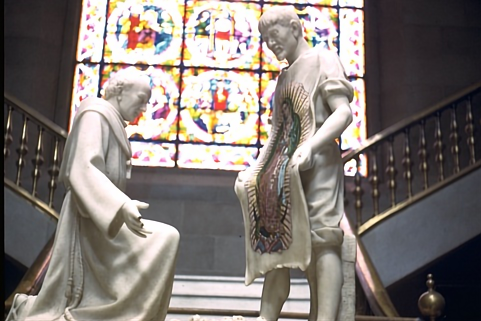}&
				\includegraphics[trim={250 120 210 148 },clip,width=.13\textwidth,valign=t]{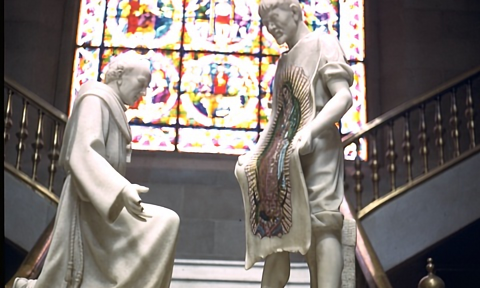}&  
				\includegraphics[trim={250 120 210 148 },clip,width=.13\textwidth,valign=t]{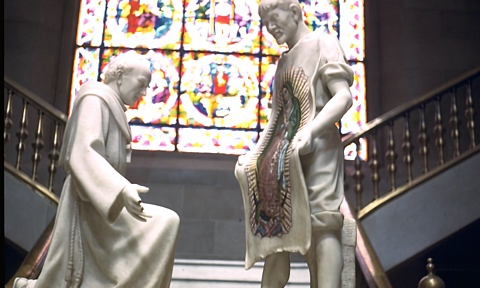}\\
				
				\scriptsize~PSNR(dB)/SSIM& \scriptsize~35.27/0.9799& \scriptsize~36.26/0.9842& \scriptsize~\textcolor{blue}{36.58}/\textcolor{blue}{0.9853}
				& \scriptsize~\textcolor{red}{37.08}/\textcolor{red}{0.9863}\\
				\scriptsize~Ground Truth& \scriptsize~COAST& \scriptsize~MADUN& \scriptsize~\textbf{PRL-PGD}&\scriptsize~\textbf{PRL-PGD$^+$}
				\\
		\end{tabular}}
  
        \hspace{-6pt}\scalebox{0.542}{
			\begin{tabular}[b]{c@{ } c@{ }  c@{ } c@{ } c@{ }}
				\multirow{4}{*}{\includegraphics[width=.312\textwidth,valign=t]{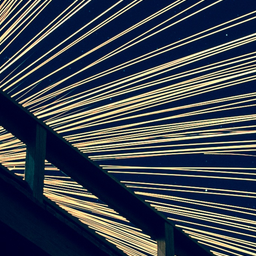}} &   
				\includegraphics[trim={108 108 113 113 },clip,width=.13\textwidth,valign=t]{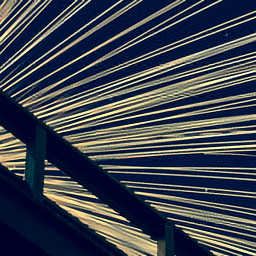}&
				\includegraphics[trim={108 108 113 113 },clip,width=.13\textwidth,valign=t]{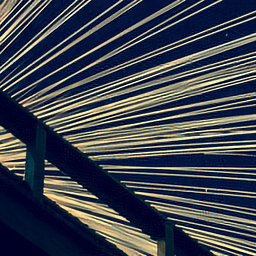}&   
				\includegraphics[trim={108 108 113 113 },clip,width=.13\textwidth,valign=t]{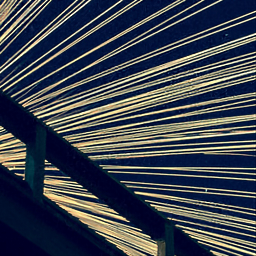}&
				\includegraphics[trim={108 108 113 113 },clip,width=.13\textwidth,valign=t]{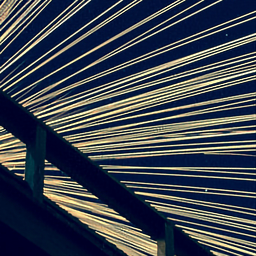}
				
				\\
				&  \scriptsize~21.00/0.9088 &\scriptsize~21.22/0.9087  & \scriptsize~26.12/0.9628& \scriptsize~27.58/0.9772\\
				& \scriptsize~CSNet$^+$&\scriptsize~SCSNet& \scriptsize~OPINE-Net$^+$& \scriptsize~AMP-Net\\
				
				&
				\includegraphics[trim={108 108 113 113 },clip,width=.13\textwidth,valign=t]{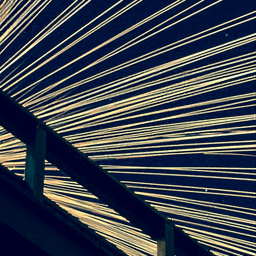}&
				\includegraphics[trim={108 108 113 113 },clip,width=.13\textwidth,valign=t]{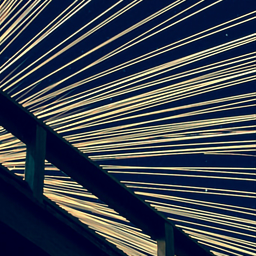}&
				\includegraphics[trim={108 108 113 113 },clip,width=.13\textwidth,valign=t]{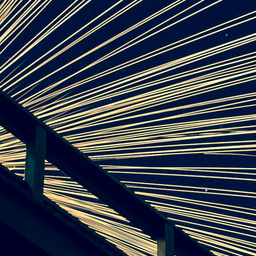}&  
				\includegraphics[trim={108 108 113 113 },clip,width=.13\textwidth,valign=t]{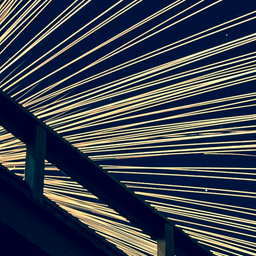}\\
				
				\scriptsize~PSNR(dB)/SSIM& \scriptsize~27.67/0.9740& \scriptsize~29.84/0.9881& \scriptsize~\textcolor{blue}{34.35}/\textcolor{blue}{0.9921}
				& \scriptsize~\textcolor{red}{36.26}/\textcolor{red}{0.9949}\\
				\scriptsize~Ground Truth& \scriptsize~COAST& \scriptsize~MADUN& \scriptsize~\textbf{PRL-PGD}& \scriptsize~\textbf{PRL-PGD$^+$}
				\\
		\end{tabular}}
	\end{center}
	\caption{Visual comparisons on recovering two images from CBSD68 \cite{kulkarni2016reconnet} (top) and DIV2K \cite{agustsson2017ntire} (bottom) respectively under the setting of $\gamma=30\%$.}
	\label{fig:vis_comp2}
\end{figure}

In this subsection, we compare our PRL and the competing methods on CBSD68 \cite{martin2001database} and DIV2K \cite{agustsson2017ntire} validation sets. From Tab.~\ref{tab:comp2}, we can see the consistent leading of PRL networks over the previous networks (about PSNR of 0.8-1.8dB, 1.4-2.1dB, 0.3-1.5dB and 0.7-1.6dB on Set11 \cite{kulkarni2016reconnet}, Urban100 \cite{huang2015single}, CBSD68 and DIV2K). Fig.~\ref{fig:vis_comp2} shows that PRL networks recover more details and fine edges, and can especially reconstruct the correct shapes and accurate line textures. We also find that the accuracy leading on CBSD68 is relatively less significant, which comes from the greater recovery difficulties of CBSD68 \cite{zhang2017beyond,zhang2021plug} and our default extremely unbalanced training set. More further experiments and related analyses about the training set are provided in Sec.~\ref{sec:effect_dataset}.

\begin{table}
        \caption{PSNR(dB)/SSIM comparisons of various CS methods on CBSD68 \cite{kulkarni2016reconnet} and DIV2K \cite{agustsson2017ntire}. The existing physics-free, physics-engaged, and our methods are in the yellow, green and cyan backgrounds, respectively, with the corresponding highlighted \textcolor{red}{best} and \textcolor{blue}{second best} results.}
	\label{tab:comp2}
	\centering
	\scalebox{0.68}{
	\begin{tabular}{l|ccc|ccc}
		\shline
		Dataset                                               & \multicolumn{3}{c|}{CBSD68}                                                                         & \multicolumn{3}{c}{DIV2K}   \\ \hline
		CS Ratio $\gamma$                                     & 10\%                            & 30\%                            & 50\%                            & 10\%    & 30\%    & 50\%    \\ \hline
		\rowcolor[HTML]{FFFFC7} 
		CSNet$^+$ \cite{shi2019image}                                            & \cellcolor[HTML]{FFFFC7}27.04   & \cellcolor[HTML]{FFFFC7}31.60   & \cellcolor[HTML]{FFFFC7}35.27   & 28.23   & 33.63   & 37.59   \\
		\rowcolor[HTML]{FFFFC7} 
		(TIP'19)                                              & \cellcolor[HTML]{FFFFC7}/0.7735 & \cellcolor[HTML]{FFFFC7}/0.9140 & \cellcolor[HTML]{FFFFC7}/0.9608 & /0.8041 & /0.9317 & /0.9692 \\ \hline
		\rowcolor[HTML]{FFFFC7} 
		SCSNet \cite{shi2019scalable}                                               & \cellcolor[HTML]{FFFFC7}27.14   & \cellcolor[HTML]{FFFFC7}31.72   & \cellcolor[HTML]{FFFFC7}35.62   & 28.41   & 33.85   & 37.97   \\
		\rowcolor[HTML]{FFFFC7} 
		(CVPR'19)                                             & \cellcolor[HTML]{FFFFC7}/0.7763 & \cellcolor[HTML]{FFFFC7}/0.9162 & \cellcolor[HTML]{FFFFC7}/0.9636 & /0.8087 & /0.9340 & /0.9711 \\ \hline
		\rowcolor[HTML]{E8FFE8} 
		OPINE-Net$^+$ \cite{zhang2020optimization}                                        & \cellcolor[HTML]{E8FFE8}27.66   & \cellcolor[HTML]{E8FFE8}32.38   & \cellcolor[HTML]{E8FFE8}36.21   & 29.26   & 35.03   & 39.27   \\
		\rowcolor[HTML]{E8FFE8} 
		(JSTSP'20)                                            & \cellcolor[HTML]{E8FFE8}/0.8022 & \cellcolor[HTML]{E8FFE8}/0.9229 & \cellcolor[HTML]{E8FFE8}/0.9655 & /0.8400 & /0.9444 & /0.9758 \\ \hline
		\rowcolor[HTML]{E8FFE8} 
		AMP-Net \cite{zhang2020amp}                                              & \cellcolor[HTML]{E8FFE8}27.71   & \cellcolor[HTML]{E8FFE8}32.72   & \cellcolor[HTML]{E8FFE8}36.72   & 29.08   & 35.41   & 39.46   \\
		\rowcolor[HTML]{E8FFE8} 
		(TIP'21)                                              & \cellcolor[HTML]{E8FFE8}/0.7900 & \cellcolor[HTML]{E8FFE8}/0.9233 & \cellcolor[HTML]{E8FFE8}/0.9678 & /0.8278 & /0.9439 & /0.9767 \\ \hline
		\rowcolor[HTML]{E8FFE8} 
		COAST \cite{you2021coast}                                                & \cellcolor[HTML]{E8FFE8}27.76   & \cellcolor[HTML]{E8FFE8}32.56   & \cellcolor[HTML]{E8FFE8}36.34   & 29.46   & 35.32   & 39.43   \\
		\rowcolor[HTML]{E8FFE8} 
		(TIP'21)                                              & \cellcolor[HTML]{E8FFE8}/0.8061 & \cellcolor[HTML]{E8FFE8}/0.9256 & \cellcolor[HTML]{E8FFE8}/0.9663 & /0.8451  & /0.9473  & /0.9766  \\ \hline
		\rowcolor[HTML]{E8FFE8} 
		MADUN \cite{song2021memory}                                                & \cellcolor[HTML]{E8FFE8}28.04   & \cellcolor[HTML]{E8FFE8}33.07   & \cellcolor[HTML]{E8FFE8}36.99   & 29.62   & 36.04   & 40.06   \\
		\rowcolor[HTML]{E8FFE8} 
		(ACM MM'21)                                           & \cellcolor[HTML]{E8FFE8}/0.8185 & \cellcolor[HTML]{E8FFE8}/0.9350 & \cellcolor[HTML]{E8FFE8}/0.9722 & /0.8570  & /0.9563  & /0.9808  \\ \hline
		\rowcolor[HTML]{CCFFFD} 
		\cellcolor[HTML]{CCFFFD}                              & \cellcolor[HTML]{CCFFFD}28.41   & \cellcolor[HTML]{CCFFFD}33.29   & \cellcolor[HTML]{CCFFFD}37.25   & 30.61   & 36.43   & 40.76   \\
		\rowcolor[HTML]{CCFFFD} 
		\multirow{-2}{*}{\cellcolor[HTML]{CCFFFD}\textbf{PRL-PGD}}     & \cellcolor[HTML]{CCFFFD}/0.8239 & \cellcolor[HTML]{CCFFFD}/0.9357 & \cellcolor[HTML]{CCFFFD}/0.9724 & /0.8698  & /0.9570  & /0.9819  \\ \hline
		\rowcolor[HTML]{CCFFFD} 
		\cellcolor[HTML]{CCFFFD}                              & \cellcolor[HTML]{CCFFFD}\textcolor{red}{28.50}   & \cellcolor[HTML]{CCFFFD}\textcolor{blue}{33.41}   & \cellcolor[HTML]{CCFFFD}\textcolor{blue}{37.37}   & \textcolor{red}{30.75}   & \textcolor{blue}{36.64}   & \textcolor{blue}{40.97}   \\
		\rowcolor[HTML]{CCFFFD} 
		\multirow{-2}{*}{\cellcolor[HTML]{CCFFFD}\textbf{PRL-PGD$^+$}} & \cellcolor[HTML]{CCFFFD}/\textcolor{red}{0.8257} & \cellcolor[HTML]{CCFFFD}/\textcolor{red}{0.9365} & \cellcolor[HTML]{CCFFFD}/\textcolor{blue}{0.9729} & /\textcolor{red}{0.8718}  & /\textcolor{blue}{0.9580}  & /\textcolor{blue}{0.9823}  \\ \hline
		\rowcolor[HTML]{CCFFFD} 
		\cellcolor[HTML]{CCFFFD}                              & \cellcolor[HTML]{CCFFFD}28.40   & \cellcolor[HTML]{CCFFFD}33.32   & \cellcolor[HTML]{CCFFFD}37.28   & 30.58   & 36.54   & 40.83   \\
		\rowcolor[HTML]{CCFFFD} 
		\multirow{-2}{*}{\cellcolor[HTML]{CCFFFD}\textbf{PRL-RND}}     & \cellcolor[HTML]{CCFFFD}/0.8237 & \cellcolor[HTML]{CCFFFD}/\textcolor{blue}{0.9359} & /0.9726                         & /0.8692  & /0.9575  & /0.9825  \\ \hline
		\rowcolor[HTML]{CCFFFD} 
		\cellcolor[HTML]{CCFFFD}                              & \cellcolor[HTML]{CCFFFD}\textcolor{blue}{28.48}   & \cellcolor[HTML]{CCFFFD}\textcolor{red}{33.42}   & \cellcolor[HTML]{CCFFFD}\textcolor{red}{37.38}   & \textcolor{blue}{30.73}   & \textcolor{red}{36.68}   & \textcolor{red}{41.02}   \\
		\rowcolor[HTML]{CCFFFD} 
		\multirow{-2}{*}{\cellcolor[HTML]{CCFFFD}\textbf{PRL-RND$^+$}} & \cellcolor[HTML]{CCFFFD}/\textcolor{blue}{0.8254} & \cellcolor[HTML]{CCFFFD}/\textcolor{red}{0.9365} & \cellcolor[HTML]{CCFFFD}/\textcolor{red}{0.9732} & /\textcolor{blue}{0.8714}  & /\textcolor{red}{0.9583}  & /\textcolor{red}{0.9827}  \\ \shline
	\end{tabular}}
\end{table}

\begin{table*}[!t]
    \centering
    \caption{PSNR comparison and \textcolor{purple}{vertical} increments of the trained \underline{improved} network variants with seven different training configurations on Set11 \cite{kulkarni2016reconnet} under the setting of $\gamma=10\%$.}
    \scalebox{0.95}{
    \begin{tabular}{c|c|c}
    \shline
    Globally Unified Value of the Random Seed     & Setting & Final PSNR (dB) \\ \hline
    $v_\text{seed}=2021$& $K=20, q=64$  & 29.53     \\
                          (the default setting)& $K=30, q=64$  & 29.51     \\ \hline
    \multirow{5}{*}{\begin{tabular}[c]{@{}c@{}}$v_\text{seed}=2022$\\(the supplementary setting in our retrainings)\end{tabular}      } & $K=20, q=64$  & 29.49     \\
                          & $K=30, q=~1$   & 28.80\scriptsize~(\textcolor{blue}{severely degraded})     \\
                          & $K=30, q=~4$   & 29.17\scriptsize~$\textcolor{purple}{+0.37}$ (\textcolor{blue}{partially alleviated})     \\
                          & $K=30, q=16$  & 29.45\scriptsize~$\textcolor{purple}{+0.28}$     \\
                          & $K=30, q=64$  & 29.54\scriptsize~$\textcolor{purple}{+0.09}$     \\ \shline
    \end{tabular}}
    \label{tab:Train400_supp}
\end{table*}

\section{More Ablation Studies and Discussions}
\label{appx:abla}
\subsection{Effect of Fully-Utilized Physics in Eliminating Degradation}
Under the single-scale/path plain framework, the existing ID unrolling scheme suffers from not only the performance saturation and large inference time increase with the unfavorable noise delicacy due to its sensitive ``slender" architecture, but also the training difficulty with serious degradation as our pilot experiment exhibited. It is eliminated by activating the full-channel physics engagements. Here we conduct more experiments on five variants of our PGD-unrolled \underline{improved} baseline in Fig.~\ref{fig:Train400}(d) with four degrees of physics utilization to give a better understanding of the effect of physical information guidance.

\begin{figure}
    \centering
    \hspace{-4pt}\includegraphics[width=0.482\textwidth]{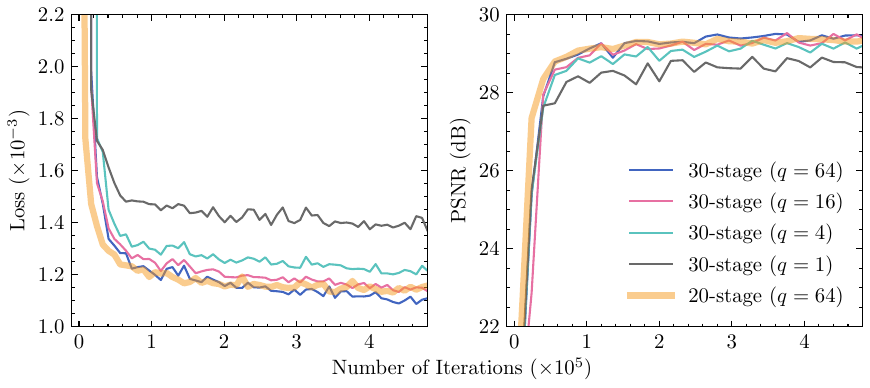}
	\caption{Training loss (left) and test PSNR (right) curves on Set11~\cite{kulkarni2016reconnet} of our five retrained \underline{improved} variants with $\gamma =10\%$. The learning of 30-stage (212-layer) network is stabilized and enhanced by more feature channels of physics guidance from $\{\mathbf{A}, \mathbf{y}\}$.}
	\label{fig:Train400_supp}
\end{figure}

Specifically, we retrain two fully-activated networks with $r\equiv 1$, $K\in\left\{20, 30\right\}$, $C=D\equiv 64$ and the maximized physical feature dimensionality $q=64$, and three partially-activated ones with $K=30$ and $q\in\left\{1,4,16\right\}$ by disabling the latter $(64-q)$ Conv kernels and activations in $\mathcal{H}^{(k)}_\text{grad}$ for comparison. Fig.~\ref{fig:Train400_supp} shows the training processes and exhibits that the 30-stage (212-layer) recovery networks are inevitably difficult to converge as expected when $q\in\left\{1, 4\right\}$, and the degradation is alleviated by $q=16$ and eliminated by $q=64$. Tab.~\ref{tab:Train400_supp} gives the final evaluations of all these trained networks on Set11~\cite{kulkarni2016reconnet} and demonstrates the effectiveness of the physics-feature fusion enhancement with a 0.74dB PSNR improvement in total brought by the sufficiently large $q$ values. Note that our neural physical guidance is implemented by multiscale unrolling in FD, and can not be achieved by only the introduction of extra parameters. We also find that this plain architecture quickly goes saturated when $K\ge 20$, and obtain similar conclusions from extra trainings with $C=D\equiv 32$ and $K\in \{80,100,120\}$. These results reveal the importance of adequate physics injection in stabilizing and enhancing the learning processes of deep physics-engaged networks for improving performance.

\begin{figure*}
    \centering
	\includegraphics[width=1.0\textwidth]{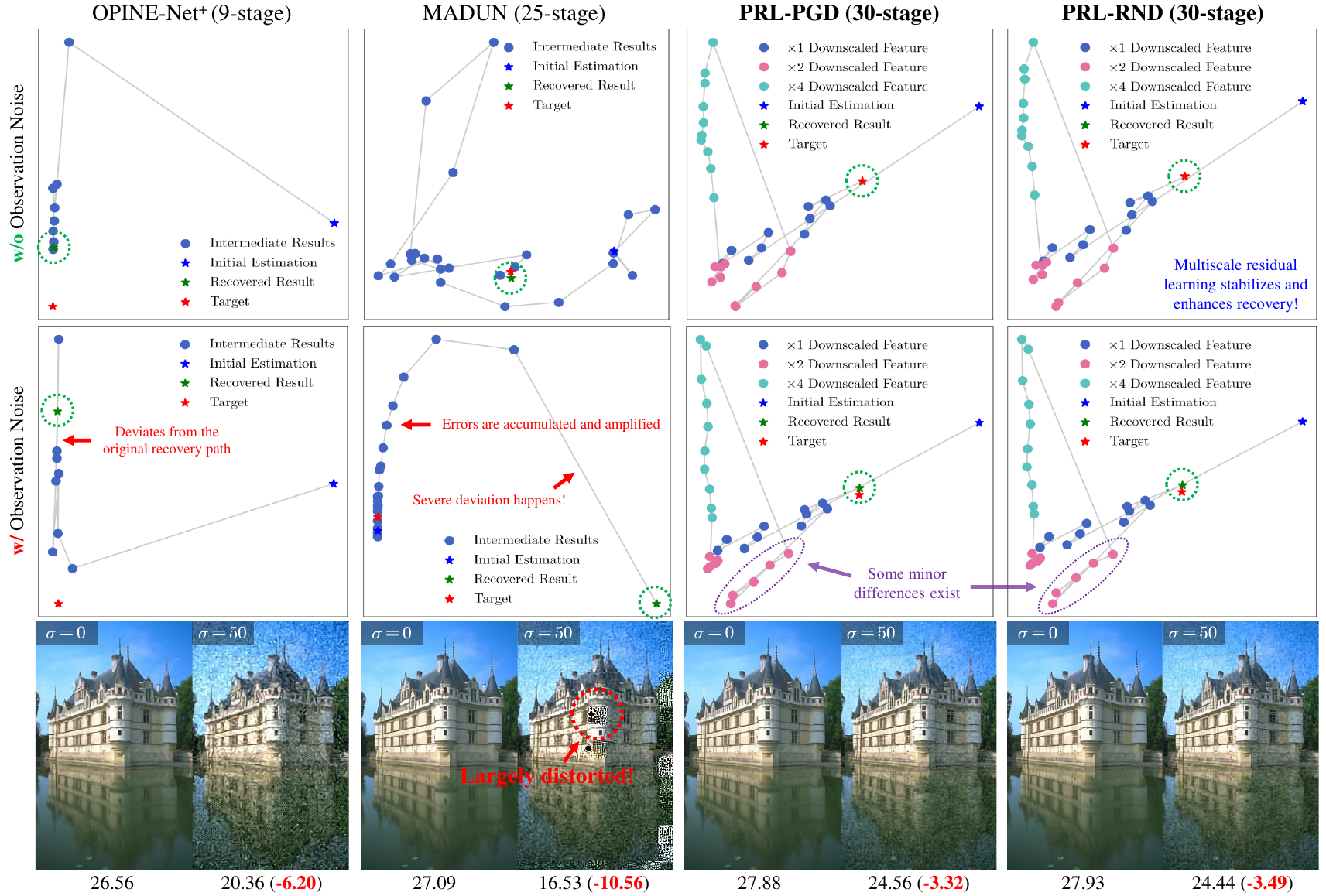}
	\caption{Visualization of the trajectories with relative data positions based on the principal component analysis (PCA) technique about recovering an image from CBSD68 \cite{martin2001database} by the plain OPINE-Net$^+$ \cite{zhang2020optimization}, MADUN \cite{song2021memory} and our PRL networks with/without noise $\mathbf{\epsilon}$ of a large standard deviation (noise level) $\sigma=50$ in observation $\mathbf{y}=\mathbf{Ax}+\mathbf{\epsilon}$ (top/middle) when $\gamma =10\%$. The recovered results (bottom) are marked by $\sigma$ and PSNR (dB) with corresponding \textcolor{red}{drops}. Compared with the existing plain unrolling networks, PRL can stabilize the recovery process with heavy noise interference.}	\label{fig:PCA_comp}
\end{figure*}

\begin{figure*}[!t]
    \centering
	\includegraphics[width=1.0\textwidth]{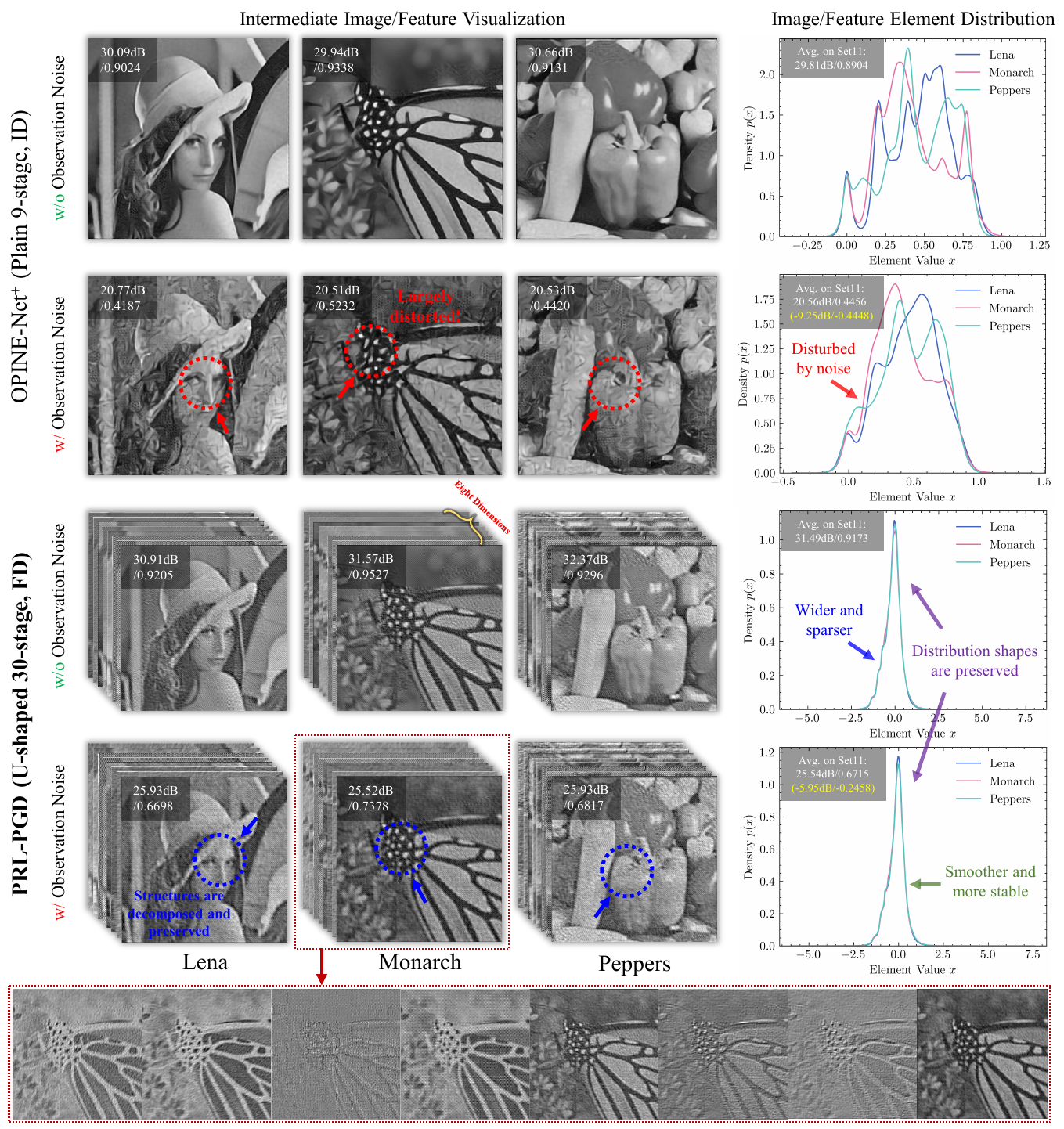}
	\caption{Visualization of the $[0,255]$-normalized intermediate images/features (left) of recovering three Set11 \cite{kulkarni2016reconnet} images (including ``Lena", ``Monarch" and ``Peppers") from the 5-/15-th stage of OPINE-Net$^+$ (top) and our default PRL-PGD (bottom) with $\sigma \in \{0,50\}$ and $\gamma =10\%$. Features from the third level of PRL-PGD are upscaled to be in 8-dimensional space and visualized channel-by-channel. Their distribution curves are on the right side. All visualizations are marked by PSNR/SSIM of the final recoveries (with drop margins) in their upper left corner. PRL-PGD recovers better under noise disturbance by preserving the image structure and feature distribution shapes based on its effective U-shaped FD unrolling.}
	\label{fig:dis_vis}
\end{figure*}

\subsection{Effect of U-Shaped Residual Recovery in Improving the Noise Robustness}
To better understand the reconstruction characteristics of different unrolled architectures, based on Fig.~\ref{fig:abla_noise}, we conduct further image- and feature-level visualizations on the refinement processings of various networks from the perspective of noise robustness. Fig.~\ref{fig:PCA_comp} provides the restoration trajectories of four PGD-unrolled networks including the plain 9-stage OPINE-Net$^+$ \cite{zhang2020optimization} and 25-stage MADUN \cite{song2021memory} in ID, and our 30-stage PRL-PGD and PRL-RND in FD about the recovery of an image from CBSD68 \cite{martin2001database} (with/without observation noise). Here we use the principal component analysis (PCA) instead of the more advanced t-SNE \cite{van2008visualizing} to reduce the image data dimensionality for all stages, since the PCA technique is simpler and more stable without hyperparameters or randomness in our evaluation. We map the high-dimensional features in PRL networks from FD to ID by $\times r$-upscaling and transforming them to be with the shape of $H\times W\times 1$ through PixelShuffle \cite{shi2016real} and $\mathcal{H}_\text{rec}$. From the visualization in Fig.~\ref{fig:PCA_comp}, we observe that compared with OPINE-Net$^+$, MADUN reconstructs better with 25-step milder content refinements in the noise-free case. Both of them are severely affected by observation noise which brings large error accumulation and a ``domino" effect of amplification in such a long and slender architecture, especially the deeper ones (see the blocking/ripple/honeycomb artifacts in OPINE-Net$^+$ recovery with a 6.20dB PSNR drop and more totally distorted blocks from MADUN with a 10.56dB PSNR drop). Similar phenomena are observed in other plain networks \cite{zhang2020amp,you2021coast} and our baseline variants. PRL preserves the trajectory structures and well keeps the recovery stable under the interference of noise (with $<$3.50dB PSNR drops). Note that the intra-stage connections and inter-stage short-/long-term memories are introduced in OPINE-Net$^+$ and MADUN for transmission enhancements. These results demonstrate the positive effect of improving noise robustness from our compact multiscale unrolling with additive long-range skip connections, which conducts three levels of residual FD recoveries and plays a major role in stabilizing the 30-stage reconstructions since our PGD-/RND-unrolling networks share the similar recovery trajectories and are with minor differences in the path detail.

\begin{table*}[!t]
	\caption{Evaluation of the recovery accuracy in PSNR, inference times and parameter numbers of four prior networks (left) and twenty PRL-PGD variants (right) on Set11 \cite{kulkarni2016reconnet} when CS ratio $\gamma =50\%$. The selected most comparable networks (considered in parameter number, network depth and inference speed) are highlighted in the same color backgrounds. Our three default settings are marked by * (PRL-PGD-S, PRL-PGD and PRL-PGD$^+$). Note that PRL-PGD gives full play to its real strength with sufficient $d\in\{8,16\}$ and is largely restricted by the poor feature capacity when $d\in\{2,4\}$.}
	\label{tab:abla_c}
	\hspace{-10pt}
	\begin{minipage}{0.5\textwidth}
	\centering
	\scalebox{0.80}{\begin{tabular}{c|c|c|c}
	\shline
    Method        & \begin{tabular}[c]{@{}c@{}}Stage\\ Number\\ $K$\end{tabular} & \begin{tabular}[c]{@{}c@{}}Feature\\ Settings\end{tabular}                                           & \begin{tabular}[c]{@{}c@{}}PSNR (dB)\\ /Time (ms)\\ /\#Param. (M)\end{tabular}              \\ \hline
    OPINE-Net$^+$ & 9                                                            &                                                                                                      & \cellcolor[HTML]{9AFF99}\begin{tabular}[c]{@{}c@{}}40.19\\ /17.305\\ /0.622\end{tabular} \\ \hhline{--~-}    AMP-Net       & 9                                                            &                                                                                                      & \cellcolor[HTML]{9AFF99}\begin{tabular}[c]{@{}c@{}}40.34\\ /27.378\\ /0.868\end{tabular} \\ \hhline{--~-} 
    COAST         & 20                                                           &                                                                                                      & \cellcolor[HTML]{96FFFB}\begin{tabular}[c]{@{}c@{}}40.33\\ /45.544\\ /1.122\end{tabular} \\ \hhline{--~-} 
    MADUN         & 25                                                           & \multirow{-10}{*}{\begin{tabular}[c]{@{}c@{}}$r\equiv 1$,\\ $C\equiv 1$,\\ $D\equiv 32$\end{tabular}} & \cellcolor[HTML]{FFFFC7}\begin{tabular}[c]{@{}c@{}}40.75\\ /92.146\\ /3.126\end{tabular} \\ \shline
    \end{tabular}}
    \end{minipage}
    \begin{minipage}{0.5\textwidth}
    \scalebox{0.8}{\begin{tabular}{c|c|c|c|c|c}
    \shline
    \begin{tabular}[c]{@{}c@{}}FD Dimen\\ -sionality\\ $C=D=d$\end{tabular} & $K=1$                                                            & $K=3$                                                             & $K=5$                                                             & \begin{tabular}[c]{@{}c@{}}$K=5$\\ (shared)\end{tabular}                                 & $K=7$                                                             \\ \hline
    2                                                                       & \begin{tabular}[c]{@{}c@{}}39.88\\ /8.506\\ /0.659\end{tabular}  & \begin{tabular}[c]{@{}c@{}}40.37\\ /23.230\\ /0.925\end{tabular}  & \begin{tabular}[c]{@{}c@{}}40.63\\ /37.325\\ /1.190\end{tabular}  & \cellcolor[HTML]{9AFF99}\begin{tabular}[c]{@{}c@{}}40.59\\ /35.798\\ /0.659\end{tabular} & \begin{tabular}[c]{@{}c@{}}40.86\\ /52.293\\ /1.456\end{tabular}  \\ \hline
    4                                                                       & \begin{tabular}[c]{@{}c@{}}40.41\\ /8.611\\ /1.043\end{tabular}  & \begin{tabular}[c]{@{}c@{}}40.93\\ /23.251\\ /2.063\end{tabular}  & \begin{tabular}[c]{@{}c@{}}41.15\\ /38.260\\ /3.083\end{tabular}  & \cellcolor[HTML]{96FFFB}\begin{tabular}[c]{@{}c@{}}41.13\\ /38.354\\ /1.043\end{tabular} & \begin{tabular}[c]{@{}c@{}}41.33\\ /53.199\\ /4.104\end{tabular}  \\ \hline
    8                                                                       & \begin{tabular}[c]{@{}c@{}}40.77\\ /8.709\\ /2.558\end{tabular}  & \begin{tabular}[c]{@{}c@{}}41.31\\ /24.017\\ /6.555\end{tabular}  & \begin{tabular}[c]{@{}c@{}}*41.52\\ /41.402\\ /10.173\end{tabular} & \cellcolor[HTML]{FFFFC7}\begin{tabular}[c]{@{}c@{}}*41.47\\ /40.873\\ /2.558\end{tabular} & \begin{tabular}[c]{@{}c@{}}41.59\\ /54.743\\ /14.549\end{tabular} \\ \hline
    16                                                                      & \begin{tabular}[c]{@{}c@{}}40.95\\ /11.169\\ /8.575\end{tabular} & \begin{tabular}[c]{@{}c@{}}41.45\\ /31.397\\ /24.397\end{tabular} & \begin{tabular}[c]{@{}c@{}}41.65\\ /51.746\\ /40.218\end{tabular} & \begin{tabular}[c]{@{}c@{}}41.61\\ /51.685\\ /8.575\end{tabular}                         & \begin{tabular}[c]{@{}c@{}}*41.78\\ /72.067\\ /55.621 \end{tabular}\\
    \shline
    \end{tabular}}
    \end{minipage}
\end{table*}

We further extract the intermediate refined images/features of three Set11 \cite{kulkarni2016reconnet} instances from the recoveries of the 5-/15-th stage in OPINE-Net$^+$/PRL-PGD, and visualize them in Fig.~\ref{fig:dis_vis} with $\sigma\in\{0,50\}$. Specifically, we provide their original distribution density curves (right) and visualize stage outputs channel-by-channel (left) with $[0,255]$-normalization and upscaling  $(H/4)\times(W/4)\times 128$ features from PRL-PGD to the shape of $H\times W\times 8$. From Fig.~\ref{fig:dis_vis} we observe that the ID recoveries of OPINE-Net$^+$ are with unstable signal distributions and loss of accurate structures of both the image content and distribution shapes under the noise disturbance (with an average PSNR/SSIM drop of 9.25dB/0.4448). Our FD recoveries (with a drop of 5.95dB/0.2458) decompose the images into multiple physics channels and preserve their spatial structures with smooth and stable distribution curves. We also find that with only an end-to-end $\ell_2$ loss function, OPINE-Net$^+$ can be regarded as a special FD network reconstructing a single-channel feature instead of totally working in ID since its non-linearly transformed intermediate results tend to be distributed in $[-0.3,1.2]$ (not the original $[0,1]$ for ID). PRL-PGD recovers better in 8-dimensional space with wider and sparser zero-mean distributions which may be helpful to enhance the edges and textures. These results validate the necessity of our FD restoration for supporting the high-throughput transmissions and sufficient physics guidance with leaving large freedom degrees for information distributions that makes PRL robust.

\begin{figure*}
    \centering
    \includegraphics[width=1.0\textwidth]{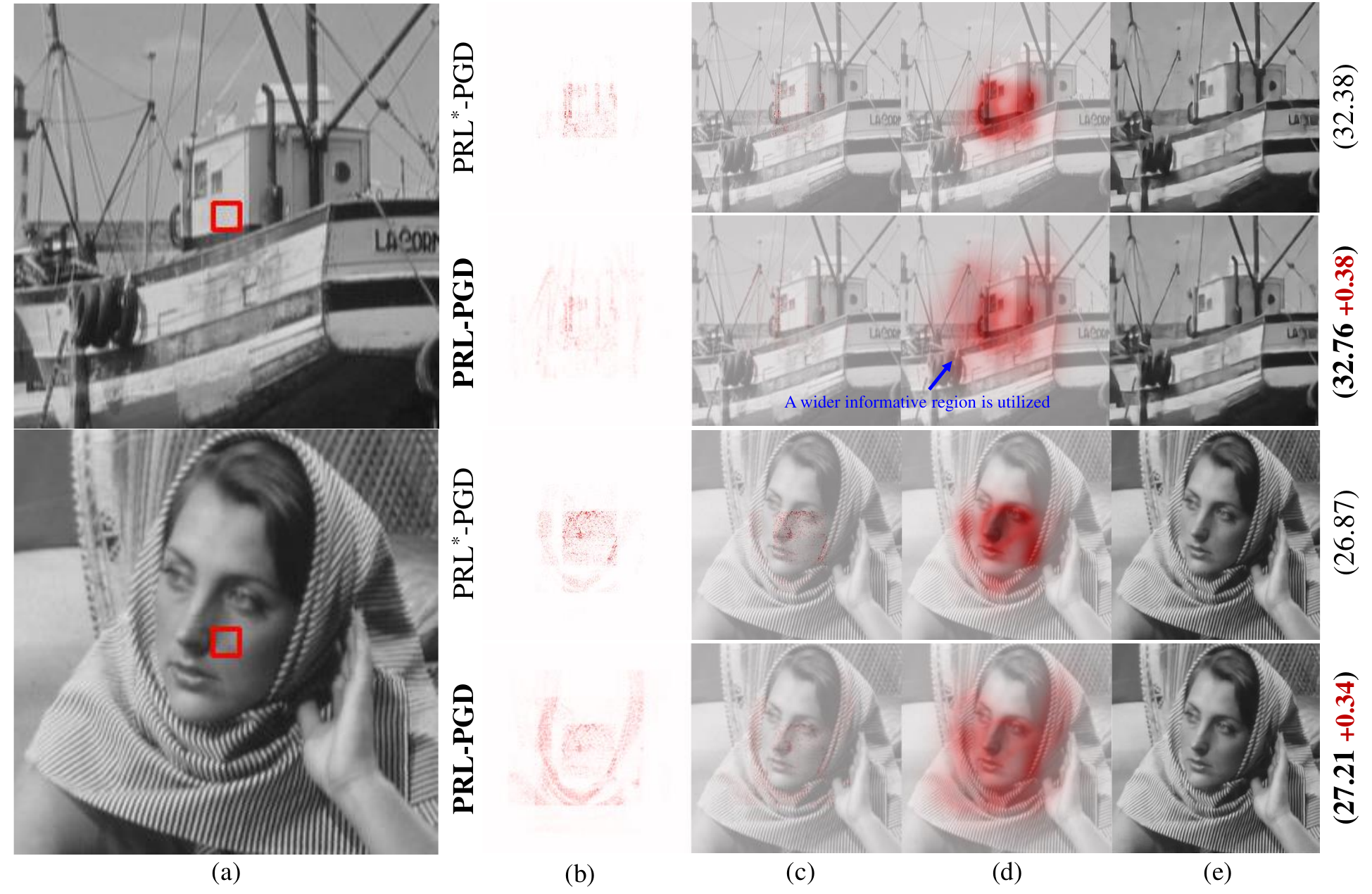}
	\caption{Visualization of the LAM \cite{gu2021interpreting} attribution results of PRL$^*$-PGD and PRL-PGD with $K=1$ and $C=D=8$ about the recovery of two images named ``Boats" (top) and ``Barbara" (bottom) from Set11 \cite{kulkarni2016reconnet} when $\gamma =10\%$. The LAM maps represent the importance of different pixels for the selected central $16\times 16$ local patches and the more informative regions are marked with darker red color. (a) is the original image with the selected region marked in a red box; (b) is the LAM result; (c) is the LAM result with input image; (d) is the informative area with input image; (e) is the CS reconstructed result. The upper/lower rows of each instance correspond to the single-/multi-scale PRL$^*$/PRL, respectively. Their recovery PSNR (dB) with accuracy distances are exhibited on the right side. One can see that PRL produces much more accurate recovered outputs by significantly increasing the range of pivotal information utilization.}
	\label{fig:LAM}
\end{figure*}

\subsection{Study of FD Dimensionality and the Inter-Stage Weight Sharing Strategy}
We evaluate 20 PRL-PGD variants with the FD dimensionalities $d\in\{2,4,8,16\}$, the stage numbers of each group $K\in\{1,3,5,7\}$ and the weight sharing across all the stages in each group with $K=5$, and report their recovery accuracy, speeds, and parameter numbers when CS ratio $\gamma =50\%$ in Tab.~\ref{tab:abla_c} with four competing unrolled methods \cite{zhang2020optimization,zhang2020amp,you2021coast,song2021memory}. We observe that the PRL performance gives full play when $d\in\{8,16\}$ and is restricted by $d\in\{2,4\}$ with poor feature capacity, so a larger FD dimensionality can sufficiently stimulate their high performance. The inter-stage weight sharing strategy brings little impact on recovery accuracy (only 0.04dB PSNR drop on average) and speed (0.506ms time saving) with about $\times(1/K)$ parameter reduction. It benefits from the compact and efficient PRL structure and is quite effective and considerable in deployments. The PRL-PGD variants most comparable to other methods (in parameter number, depth and speed) bring the average improvements including PSNR leading of 0.66dB, 16\% time reduction and 26\% parameter number saving under such a high sampling rate. Note that these PRL variants are limited by insufficient configurations but the competing networks are their own optimal and saturated versions. These results verify the higher structural efficiency and appealing compression potential of PRL with its flexible and competent optional settings for real-world applications.

\begin{table*}
\caption{PSNR(dB)/SSIM comparisons among our default PGD-unrolled PRL networks with three different training set configurations (a), (b) and (c) on four benchmarks in the case of $\gamma =50\%$. The recovery accuracy differences between two settings (x) and (y) are marked by (x)-(y) with relative \textcolor{red}{positive} and \textcolor{blue}{negative} effects.}
\label{tab:abla_set}
\hspace{-4pt}
\scalebox{0.75}{
\begin{tabular}{c|cc|cc|cc|cc|cc}
\shline
\multirow{2}{*}{Setting}                                & \multicolumn{1}{c|}{\multirow{2}{*}{Training Set}}                                     & \multirow{2}{*}{\begin{tabular}[c]{@{}c@{}}Image\\Number\end{tabular}} & \multicolumn{2}{c|}{Set11 \cite{kulkarni2016reconnet}} & \multicolumn{2}{c|}{Urban100 \cite{huang2015single}}  & \multicolumn{2}{c|}{CBSD68 \cite{martin2001database}} & \multicolumn{2}{c}{DIV2K \cite{agustsson2017ntire}} \\ \hhline{~~~--------}& \multicolumn{1}{c|}{}&& \multicolumn{1}{c|}{\begin{tabular}[c]{@{}c@{}}PRL\\-PGD\end{tabular}} & \begin{tabular}[c]{@{}c@{}}PRL\\-PGD$^+$\end{tabular} & \multicolumn{1}{c|}{\begin{tabular}[c]{@{}c@{}}PRL\\-PGD\end{tabular}} & \begin{tabular}[c]{@{}c@{}}PRL\\-PGD$^+$\end{tabular} & \multicolumn{1}{c|}{\begin{tabular}[c]{@{}c@{}}PRL\\-PGD\end{tabular}} & \begin{tabular}[c]{@{}c@{}}PRL\\-PGD$^+$\end{tabular}& \multicolumn{1}{c|}{\begin{tabular}[c]{@{}c@{}}PRL\\-PGD\end{tabular}} & \begin{tabular}[c]{@{}c@{}}PRL\\-PGD$^+$\end{tabular}  \\ \hline
(a)                                                     & \multicolumn{1}{c|}{Train400}                                                          & 400                           & \multicolumn{1}{c|}{\begin{tabular}[c]{@{}c@{}}41.40\\ /0.9837\end{tabular}}  & \begin{tabular}[c]{@{}c@{}}41.64\\ /0.9840\end{tabular}  & \multicolumn{1}{c|}{\begin{tabular}[c]{@{}c@{}}38.01\\ /0.9778\end{tabular}}  & \begin{tabular}[c]{@{}c@{}}38.49\\ /0.9785\end{tabular}  & \multicolumn{1}{c|}{\begin{tabular}[c]{@{}c@{}}37.42\\ /0.9728\end{tabular}}  & \begin{tabular}[c]{@{}c@{}}37.50\\ /0.9734\end{tabular}  & \multicolumn{1}{c|}{\begin{tabular}[c]{@{}c@{}}40.68\\ /0.9816\end{tabular}}  & \begin{tabular}[c]{@{}c@{}}40.90\\ /0.9821\end{tabular}  \\ \hline
(b)                                                     & \multicolumn{1}{c|}{\begin{tabular}[c]{@{}c@{}}T91+Train400\\ +DIV2K\end{tabular}}     & 1291                          & \multicolumn{1}{c|}{\begin{tabular}[c]{@{}c@{}}41.38\\ /0.9836\end{tabular}}  & \begin{tabular}[c]{@{}c@{}}41.60\\ /0.9839\end{tabular}  & \multicolumn{1}{c|}{\begin{tabular}[c]{@{}c@{}}37.91\\ /0.9774\end{tabular}}  & \begin{tabular}[c]{@{}c@{}}38.32\\ /0.9781\end{tabular}  & \multicolumn{1}{c|}{\begin{tabular}[c]{@{}c@{}}37.31\\ /0.9726\end{tabular}}  & \begin{tabular}[c]{@{}c@{}}37.40\\ /0.9732\end{tabular}  & \multicolumn{1}{c|}{\begin{tabular}[c]{@{}c@{}}40.72\\ /0.9817\end{tabular}}  & \begin{tabular}[c]{@{}c@{}}40.95\\ /0.9822\end{tabular}  \\ \hline
(b)-(a)                                                 & \multicolumn{2}{c|}{-}                                                                                                 & \multicolumn{1}{c|}{\begin{tabular}[c]{@{}c@{}}\textcolor{blue}{-0.02}\\ /\textcolor{blue}{-0.0001}\end{tabular}} & \begin{tabular}[c]{@{}c@{}}\textcolor{blue}{-0.04}\\ /\textcolor{blue}{-0.0001}\end{tabular} & \multicolumn{1}{c|}{\begin{tabular}[c]{@{}c@{}}\textcolor{blue}{-0.10}\\ /\textcolor{blue}{-0.0004}\end{tabular}} & \begin{tabular}[c]{@{}c@{}}\textcolor{blue}{-0.17}\\ /\textcolor{blue}{-0.0004}\end{tabular} & \multicolumn{1}{c|}{\begin{tabular}[c]{@{}c@{}}\textcolor{blue}{-0.09}\\ /\textcolor{blue}{-0.0002}\end{tabular}} & \begin{tabular}[c]{@{}c@{}}\textcolor{blue}{-0.10}\\ /\textcolor{blue}{-0.0002}\end{tabular} & \multicolumn{1}{c|}{\begin{tabular}[c]{@{}c@{}}\textcolor{red}{+0.04}\\ /\textcolor{red}{+0.0001}\end{tabular}} & \begin{tabular}[c]{@{}c@{}}\textcolor{red}{+0.05}\\ /\textcolor{red}{+0.0001}\end{tabular} \\ \hline
\begin{tabular}[c]{@{}c@{}}(c)\\ (Default)\end{tabular} & \multicolumn{1}{c|}{\begin{tabular}[c]{@{}c@{}}T91+Train400\\ +DIV2K+WED\end{tabular}} & 6035                          & \multicolumn{1}{c|}{\begin{tabular}[c]{@{}c@{}}41.52\\ /0.9842\end{tabular}}  & \begin{tabular}[c]{@{}c@{}}41.78\\ /0.9847\end{tabular}  & \multicolumn{1}{c|}{\begin{tabular}[c]{@{}c@{}}38.12\\ /0.9783\end{tabular}}  & \begin{tabular}[c]{@{}c@{}}38.57\\ /0.9794\end{tabular}  & \multicolumn{1}{c|}{\begin{tabular}[c]{@{}c@{}}37.25\\ /0.9724\end{tabular}}  & \begin{tabular}[c]{@{}c@{}}37.37\\ /0.9729\end{tabular}  & \multicolumn{1}{c|}{\begin{tabular}[c]{@{}c@{}}40.76\\ /0.9819\end{tabular}}  & \begin{tabular}[c]{@{}c@{}}40.97\\ /0.9823\end{tabular}  \\ \hline
(c)-(b)                                                 & \multicolumn{2}{c|}{-}                                                                                                 & \multicolumn{1}{c|}{\begin{tabular}[c]{@{}c@{}}\textcolor{red}{+0.14}\\ /\textcolor{red}{+0.0006}\end{tabular}} & \begin{tabular}[c]{@{}c@{}}\textcolor{red}{+0.18}\\ /\textcolor{red}{+0.0008}\end{tabular} & \multicolumn{1}{c|}{\begin{tabular}[c]{@{}c@{}}\textcolor{red}{+0.21}\\ /\textcolor{red}{+0.0009}\end{tabular}} & \begin{tabular}[c]{@{}c@{}}\textcolor{red}{+0.25}\\ /\textcolor{red}{+0.0013}\end{tabular} & \multicolumn{1}{c|}{\begin{tabular}[c]{@{}c@{}}\textcolor{blue}{-0.06}\\ /\textcolor{blue}{-0.0002}\end{tabular}} & \begin{tabular}[c]{@{}c@{}}\textcolor{blue}{-0.03}\\ /\textcolor{blue}{-0.0003}\end{tabular} & \multicolumn{1}{c|}{\begin{tabular}[c]{@{}c@{}}\textcolor{red}{+0.04}\\ /\textcolor{red}{+0.0002}\end{tabular}} & \begin{tabular}[c]{@{}c@{}}\textcolor{red}{+0.02}\\ /\textcolor{red}{+0.0001}\end{tabular}\\
\shline
\end{tabular}}
\end{table*}

\subsection{Effect of Multiscale Architecture and Large Training Set in Improving Recovery Accuracy}
\label{sec:effect_dataset}
One important characteristic of PRL different from existing unrolled networks is the multiscale design that not only speeds up inference but also provides inter-scale FD communications for capturing the image context and structure. With a parameter number close to PRL, the single-scale U-shaped PRL$^*$ has a receptive field size of $R_{\text{PRL}^*}=(84K+13)^2=(2L+1)^2$ in its longest fully-convolutional path, where $K$ is group stage number and $L$ is the $3\times 3$ Conv layer number or network depth. But PRL has a much larger size of $R_{\text{PRL}}=(140K+8)^2=(10L/3-12)^2$ based on the introduction of $2\times 2$ strided/transposed (S-/T-) Conv layers, which leads to $R_{\text{PRL}}=(5/3)^2 R_{\text{PRL}^*}$ when $L\rightarrow \infty$. It is quite important to reconstruct a specific region by utilizing a larger informative image part \cite{zhang2017learning}. Here we analyze the effect of the receptive field differences between the single-/multi-scale PRL variants by using the local attribution map (LAM) \cite{gu2021interpreting} to compare the input image parts that have a strong influence on the CS recovered outputs when recovering two images from Set11 \cite{kulkarni2016reconnet} with $\gamma =10\%$. The two LAM visualization results in Fig.~\ref{fig:LAM} indicate that both PGD-unrolled PRL and PRL$^*$ perform the sampling and recovery for the selected local patch with some edges and textures by intensively utilizing a region with rich and relevant information. PRL significantly takes notice of a wider range of informative clean areas to guide its more accurate recoveries and achieves an average PSNR gain of 0.36dB over the PRL$^*$, thus revealing the stronger information capture capability and higher structural efficiency of our multiscale design. We also find that due to the block-based CS scheme (with a globally unified block size of $32\times 32$) and the iterative stage-by-stage physical knowledge injections of $\{\mathbf{A}, \mathbf{y}\}$, the actual network receptive field is larger than their fully-convolutional paths as we calculated above. The most heavily utilized informative regions are with complicated and content-correlated specific spatial shapes and distributions.

To enhance the training sample diversity, we follow \cite{zhang2021plug,you2021ista,song2021memory} to use four datasets including T91 \cite{yang2008image,dong2014learning}, Train400 \cite{chen2016trainable,zhang2017beyond}, DIV2K \cite{agustsson2017ntire} training set and the Waterloo exploration database (WED) \cite{ma2016waterloo}. They cover a large image space and can enrich the network prior to many classic image restoration tasks \cite{zhang2021plug}. Here we conduct more PRL-PGD trainings with three different dataset settings, report their evaluation results in Tab.~\ref{tab:abla_set}, and observe from which that our network trainings on only the Train400 dataset can bring the consistent recovery accuracy average leading of PRL (PGD-unrolled versions) over the unrolled network MADUN \cite{song2021memory} in PSNR/SSIM of about 0.77dB/0.0008, 1.56dB/0.0036, 0.47dB/0.0009 and 0.71dB/0.0011 on Set11 \cite{kulkarni2016reconnet}, Urban100 \cite{huang2015single}, CBSD68 \cite{martin2001database} and DIV2K validation set. The introductions of T91/DIV2K and WED generally bring negative and positive effects in most benchmarks, respectively. The combination of them provides the average accuracy changes of +0.13dB/+0.0006, +0.10dB/+0.0007, -0.14dB/-0.0005, and +0.08dB/+0.0003. Especially, WED overturns the negative impacts of T91/DIV2K in Set11 and Urban100 with appreciable improvements. We also find that the recovery performance on CBSD68 can not directly benefit from the larger training set due to its unbalanced sample distribution (only $6.6\%$ training images are $180\times 180$ grayscale BSD patches), thus verifying the data bias limitation that restricts network learning and causes the overfitting on PRL variants with large capacities.

\end{appendices}

\bibliography{sn-bibliography}


\begin{thebibliography}{94}
\ifx \bisbn   \undefined \def \bisbn  #1{ISBN #1}\fi
\ifx \binits  \undefined \def \binits#1{#1}\fi
\ifx \bauthor  \undefined \def \bauthor#1{#1}\fi
\ifx \batitle  \undefined \def \batitle#1{#1}\fi
\ifx \bjtitle  \undefined \def \bjtitle#1{#1}\fi
\ifx \bvolume  \undefined \def \bvolume#1{\textbf{#1}}\fi
\ifx \byear  \undefined \def \byear#1{#1}\fi
\ifx \bissue  \undefined \def \bissue#1{#1}\fi
\ifx \bfpage  \undefined \def \bfpage#1{#1}\fi
\ifx \blpage  \undefined \def \blpage #1{#1}\fi
\ifx \burl  \undefined \def \burl#1{\textsf{#1}}\fi
\ifx \doiurl  \undefined \def \doiurl#1{\url{https://doi.org/#1}}\fi
\ifx \betal  \undefined \def \betal{\textit{et al.}}\fi
\ifx \binstitute  \undefined \def \binstitute#1{#1}\fi
\ifx \binstitutionaled  \undefined \def \binstitutionaled#1{#1}\fi
\ifx \bctitle  \undefined \def \bctitle#1{#1}\fi
\ifx \beditor  \undefined \def \beditor#1{#1}\fi
\ifx \bpublisher  \undefined \def \bpublisher#1{#1}\fi
\ifx \bbtitle  \undefined \def \bbtitle#1{#1}\fi
\ifx \bedition  \undefined \def \bedition#1{#1}\fi
\ifx \bseriesno  \undefined \def \bseriesno#1{#1}\fi
\ifx \blocation  \undefined \def \blocation#1{#1}\fi
\ifx \bsertitle  \undefined \def \bsertitle#1{#1}\fi
\ifx \bsnm \undefined \def \bsnm#1{#1}\fi
\ifx \bsuffix \undefined \def \bsuffix#1{#1}\fi
\ifx \bparticle \undefined \def \bparticle#1{#1}\fi
\ifx \barticle \undefined \def \barticle#1{#1}\fi
\bibcommenthead
\ifx \bconfdate \undefined \def \bconfdate #1{#1}\fi
\ifx \botherref \undefined \def \botherref #1{#1}\fi
\ifx \url \undefined \def \url#1{\textsf{#1}}\fi
\ifx \bchapter \undefined \def \bchapter#1{#1}\fi
\ifx \bbook \undefined \def \bbook#1{#1}\fi
\ifx \bcomment \undefined \def \bcomment#1{#1}\fi
\ifx \oauthor \undefined \def \oauthor#1{#1}\fi
\ifx \citeauthoryear \undefined \def \citeauthoryear#1{#1}\fi
\ifx \endbibitem  \undefined \def \endbibitem {}\fi
\ifx \bconflocation  \undefined \def \bconflocation#1{#1}\fi
\ifx \arxivurl  \undefined \def \arxivurl#1{\textsf{#1}}\fi
\csname PreBibitemsHook\endcsname

\bibitem{donoho2006compressed}
\begin{barticle}
\bauthor{\bsnm{Donoho}, \binits{D.L.}}:
\batitle{{Compressed Sensing}}.
\bjtitle{{IEEE Transactions on Information Theory}}
\bvolume{52}(\bissue{4}),
\bfpage{1289}--\blpage{1306}
(\byear{2006})
\end{barticle}
\endbibitem

\bibitem{candes2008introduction}
\begin{barticle}
\bauthor{\bsnm{Cand{\`e}s}, \binits{E.J.}},
\bauthor{\bsnm{Wakin}, \binits{M.B.}}:
\batitle{{An Introduction to Compressive Sampling}}.
\bjtitle{{IEEE Signal Processing Magazine}}
\bvolume{25}(\bissue{2}),
\bfpage{21}--\blpage{30}
(\byear{2008})
\end{barticle}
\endbibitem

\bibitem{lustig2008compressed}
\begin{barticle}
\bauthor{\bsnm{Lustig}, \binits{M.}},
\bauthor{\bsnm{Donoho}, \binits{D.L.}},
\bauthor{\bsnm{Santos}, \binits{J.M.}},
\bauthor{\bsnm{Pauly}, \binits{J.M.}}:
\batitle{{Compressed Sensing MRI}}.
\bjtitle{{IEEE Signal Processing Magazine}}
\bvolume{25}(\bissue{2}),
\bfpage{72}--\blpage{82}
(\byear{2008})
\end{barticle}
\endbibitem

\bibitem{lustig2007sparse}
\begin{barticle}
\bauthor{\bsnm{Lustig}, \binits{M.}},
\bauthor{\bsnm{Donoho}, \binits{D.}},
\bauthor{\bsnm{Pauly}, \binits{J.M.}}:
\batitle{{Sparse MRI: The Application of Compressed Sensing for Rapid MR
  Imaging}}.
\bjtitle{{Magnetic Resonance in Medicine}}
\bvolume{58}(\bissue{6}),
\bfpage{1182}--\blpage{1195}
(\byear{2007})
\end{barticle}
\endbibitem

\bibitem{szczykutowicz2010dual}
\begin{barticle}
\bauthor{\bsnm{Szczykutowicz}, \binits{T.P.}},
\bauthor{\bsnm{Chen}, \binits{G.-H.}}:
\batitle{{Dual Energy CT Using Slow kVp Switching Acquisition and Prior Image
  Constrained Compressed Sensing}}.
\bjtitle{{Physics in Medicine \& Biology}}
\bvolume{55}(\bissue{21}),
\bfpage{6411}
(\byear{2010})
\end{barticle}
\endbibitem

\bibitem{gan2007block}
\begin{bchapter}
\bauthor{\bsnm{Gan}, \binits{L.}}:
\bctitle{{Block Compressed Sensing of Natural Images}}.
In: \bbtitle{{Proceedings of IEEE International Conference on Digital Signal
  Processing (ICDSP)}},
pp. \bfpage{403}--\blpage{406}
(\byear{2007}).
\bcomment{IEEE}
\end{bchapter}
\endbibitem

\bibitem{mun2009block}
\begin{bchapter}
\bauthor{\bsnm{Mun}, \binits{S.}},
\bauthor{\bsnm{Fowler}, \binits{J.E.}}:
\bctitle{{Block Compressed Sensing of Images Using Directional Transforms}}.
In: \bbtitle{{Proceedings of IEEE International Conference on Image Processing
  (ICIP)}},
pp. \bfpage{3021}--\blpage{3024}
(\byear{2009})
\end{bchapter}
\endbibitem

\bibitem{fowler2012block}
\begin{barticle}
\bauthor{\bsnm{Fowler}, \binits{J.E.}},
\bauthor{\bsnm{Mun}, \binits{S.}},
\bauthor{\bsnm{Tramel}, \binits{E.W.}}, \betal:
\batitle{{Block-Based Compressed Sensing of Images and Video}}.
\bjtitle{{Foundations and Trends in Signal Processing}}
\bvolume{4}(\bissue{4}),
\bfpage{297}--\blpage{416}
(\byear{2012})
\end{barticle}
\endbibitem

\bibitem{chen2022content}
\begin{barticle}
\bauthor{\bsnm{Chen}, \binits{B.}},
\bauthor{\bsnm{Zhang}, \binits{J.}}:
\batitle{{Content-Aware Scalable Deep Compressed Sensing}}.
\bjtitle{IEEE Transactions on Image Processing}
\bvolume{31},
\bfpage{5412}--\blpage{5426}
(\byear{2022})
\end{barticle}
\endbibitem

\bibitem{song2023dynamic}
\begin{barticle}
\bauthor{\bsnm{Song}, \binits{J.}},
\bauthor{\bsnm{Chen}, \binits{B.}},
\bauthor{\bsnm{Zhang}, \binits{J.}}:
\batitle{{Dynamic Path-Controllable Deep Unfolding Network for Compressive
  Sensing}}.
\bjtitle{IEEE Transactions on Image Processing}
\bvolume{32},
\bfpage{2202}--\blpage{2214}
(\byear{2023})
\end{barticle}
\endbibitem

\bibitem{chen2020deep}
\begin{bchapter}
\bauthor{\bsnm{Chen}, \binits{D.}},
\bauthor{\bsnm{Davies}, \binits{M.E.}}:
\bctitle{{Deep Decomposition Learning for Inverse Imaging Problems}}.
In: \bbtitle{{Proceedings of European Conference on Computer Vision (ECCV)}},
pp. \bfpage{510}--\blpage{526}
(\byear{2020})
\end{bchapter}
\endbibitem

\bibitem{chen2021equivariant}
\begin{bchapter}
\bauthor{\bsnm{Chen}, \binits{D.}},
\bauthor{\bsnm{Tachella}, \binits{J.}},
\bauthor{\bsnm{Davies}, \binits{M.E.}}:
\bctitle{{Equivariant Imaging: Learning Beyond the Range Space}}.
In: \bbtitle{{Proceedings of IEEE International Conference on Computer Vision
  (ICCV)}},
pp. \bfpage{4379}--\blpage{4388}
(\byear{2021})
\end{bchapter}
\endbibitem

\bibitem{zhao2016reducing}
\begin{barticle}
\bauthor{\bsnm{Zhao}, \binits{C.}},
\bauthor{\bsnm{Zhang}, \binits{J.}},
\bauthor{\bsnm{Ma}, \binits{S.}},
\bauthor{\bsnm{Fan}, \binits{X.}},
\bauthor{\bsnm{Zhang}, \binits{Y.}},
\bauthor{\bsnm{Gao}, \binits{W.}}:
\batitle{{Reducing Image Compression Artifacts by Structural Sparse
  Representation and Quantization Constraint Prior}}.
\bjtitle{{IEEE Transactions on Circuits and Systems for Video Technology}}
\bvolume{27}(\bissue{10}),
\bfpage{2057}--\blpage{2071}
(\byear{2016})
\end{barticle}
\endbibitem

\bibitem{zhao2016video}
\begin{barticle}
\bauthor{\bsnm{Zhao}, \binits{C.}},
\bauthor{\bsnm{Ma}, \binits{S.}},
\bauthor{\bsnm{Zhang}, \binits{J.}},
\bauthor{\bsnm{Xiong}, \binits{R.}},
\bauthor{\bsnm{Gao}, \binits{W.}}:
\batitle{{Video Compressive Sensing Reconstruction via Reweighted Residual
  Sparsity}}.
\bjtitle{{IEEE Transactions on Circuits and Systems for Video Technology}}
\bvolume{27}(\bissue{6}),
\bfpage{1182}--\blpage{1195}
(\byear{2016})
\end{barticle}
\endbibitem

\bibitem{zhang2014image}
\begin{barticle}
\bauthor{\bsnm{Zhang}, \binits{J.}},
\bauthor{\bsnm{Zhao}, \binits{C.}},
\bauthor{\bsnm{Zhao}, \binits{D.}},
\bauthor{\bsnm{Gao}, \binits{W.}}:
\batitle{{Image Compressive Sensing Recovery Using Adaptively Learned
  Sparsifying Basis via L0 Minimization}}.
\bjtitle{{Signal Processing}}
\bvolume{103},
\bfpage{114}--\blpage{126}
(\byear{2014})
\end{barticle}
\endbibitem

\bibitem{elad2010sparse}
\begin{bbook}
\bauthor{\bsnm{Elad}, \binits{M.}}:
\bbtitle{{Sparse and Redundant Representations: From Theory to Applications in
  Signal and Image Processing}}
vol. \bseriesno{2}.
\bpublisher{{Springer Science \& Business Media}}
(\byear{2010})
\end{bbook}
\endbibitem

\bibitem{nam2013cosparse}
\begin{barticle}
\bauthor{\bsnm{Nam}, \binits{S.}},
\bauthor{\bsnm{Davies}, \binits{M.E.}},
\bauthor{\bsnm{Elad}, \binits{M.}},
\bauthor{\bsnm{Gribonval}, \binits{R.}}:
\batitle{{The Cosparse Analysis Model and Algorithms}}.
\bjtitle{{Applied and Computational Harmonic Analysis}}
\bvolume{34}(\bissue{1}),
\bfpage{30}--\blpage{56}
(\byear{2013})
\end{barticle}
\endbibitem

\bibitem{dong2014compressive}
\begin{barticle}
\bauthor{\bsnm{Dong}, \binits{W.}},
\bauthor{\bsnm{Shi}, \binits{G.}},
\bauthor{\bsnm{Li}, \binits{X.}},
\bauthor{\bsnm{Ma}, \binits{Y.}},
\bauthor{\bsnm{Huang}, \binits{F.}}:
\batitle{{Compressive Sensing via Nonlocal Low-Rank Regularization}}.
\bjtitle{{IEEE Transactions on Image Processing}}
\bvolume{23}(\bissue{8}),
\bfpage{3618}--\blpage{3632}
(\byear{2014})
\end{barticle}
\endbibitem

\bibitem{cai2010singular}
\begin{barticle}
\bauthor{\bsnm{Cai}, \binits{J.-F.}},
\bauthor{\bsnm{Cand{\`e}s}, \binits{E.J.}},
\bauthor{\bsnm{Shen}, \binits{Z.}}:
\batitle{{A Singular Value Thresholding Algorithm for Matrix Completion}}.
\bjtitle{{SIAM Journal on Optimization}}
\bvolume{20}(\bissue{4}),
\bfpage{1956}--\blpage{1982}
(\byear{2010})
\end{barticle}
\endbibitem

\bibitem{long2019low}
\begin{barticle}
\bauthor{\bsnm{Long}, \binits{Z.}},
\bauthor{\bsnm{Liu}, \binits{Y.}},
\bauthor{\bsnm{Chen}, \binits{L.}},
\bauthor{\bsnm{Zhu}, \binits{C.}}:
\batitle{Low rank tensor completion for multiway visual data}.
\bjtitle{{Signal Processing}}
\bvolume{155},
\bfpage{301}--\blpage{316}
(\byear{2019})
\end{barticle}
\endbibitem

\bibitem{liu2018image}
\begin{barticle}
\bauthor{\bsnm{Liu}, \binits{Y.}},
\bauthor{\bsnm{Long}, \binits{Z.}},
\bauthor{\bsnm{Zhu}, \binits{C.}}:
\batitle{{Image Completion Using Low Tensor Tree Rank and Total Variation
  Minimization}}.
\bjtitle{{IEEE Transactions on Multimedia}}
\bvolume{21}(\bissue{2}),
\bfpage{338}--\blpage{350}
(\byear{2018})
\end{barticle}
\endbibitem

\bibitem{liu2019low}
\begin{barticle}
\bauthor{\bsnm{Liu}, \binits{Y.}},
\bauthor{\bsnm{Long}, \binits{Z.}},
\bauthor{\bsnm{Huang}, \binits{H.}},
\bauthor{\bsnm{Zhu}, \binits{C.}}:
\batitle{{Low CP Rank and Tucker Rank Tensor Completion for Estimating Missing
  Components in Image Data}}.
\bjtitle{{IEEE Transactions on Circuits and Systems for Video Technology}}
\bvolume{30}(\bissue{4}),
\bfpage{944}--\blpage{954}
(\byear{2019})
\end{barticle}
\endbibitem

\bibitem{long2021bayesian}
\begin{barticle}
\bauthor{\bsnm{Long}, \binits{Z.}},
\bauthor{\bsnm{Zhu}, \binits{C.}},
\bauthor{\bsnm{Liu}, \binits{J.}},
\bauthor{\bsnm{Liu}, \binits{Y.}}:
\batitle{{Bayesian Low Rank Tensor Ring for Image Recovery}}.
\bjtitle{{IEEE Transactions on Image Processing}}
\bvolume{30},
\bfpage{3568}--\blpage{3580}
(\byear{2021})
\end{barticle}
\endbibitem

\bibitem{long2022trainable}
\begin{barticle}
\bauthor{\bsnm{Long}, \binits{Z.}},
\bauthor{\bsnm{Zhu}, \binits{C.}},
\bauthor{\bsnm{Liu}, \binits{J.}},
\bauthor{\bsnm{Comon}, \binits{P.}},
\bauthor{\bsnm{Liu}, \binits{Y.}}:
\batitle{{Trainable Subspaces for Low Rank Tensor Completion: Model and
  Analysis}}.
\bjtitle{{IEEE Transactions on Signal Processing}}
\bvolume{70},
\bfpage{2502}--\blpage{2517}
(\byear{2022})
\end{barticle}
\endbibitem

\bibitem{kulkarni2016reconnet}
\begin{bchapter}
\bauthor{\bsnm{Kulkarni}, \binits{K.}},
\bauthor{\bsnm{Lohit}, \binits{S.}},
\bauthor{\bsnm{Turaga}, \binits{P.}},
\bauthor{\bsnm{Kerviche}, \binits{R.}},
\bauthor{\bsnm{Ashok}, \binits{A.}}:
\bctitle{{ReconNet: Non-iterative Reconstruction of Images from Compressively
  Sensed Measurements}}.
In: \bbtitle{{Proceedings of IEEE Conference on Computer Vision and Pattern
  Recognition (CVPR)}},
pp. \bfpage{449}--\blpage{458}
(\byear{2016})
\end{bchapter}
\endbibitem

\bibitem{ravishankar2019image}
\begin{barticle}
\bauthor{\bsnm{Ravishankar}, \binits{S.}},
\bauthor{\bsnm{Ye}, \binits{J.C.}},
\bauthor{\bsnm{Fessler}, \binits{J.A.}}:
\batitle{{Image Reconstruction: From Sparsity to Data-Adaptive Methods and
  Machine Learning}}.
\bjtitle{{Proceedings of the IEEE}}
\bvolume{108}(\bissue{1}),
\bfpage{86}--\blpage{109}
(\byear{2019})
\end{barticle}
\endbibitem

\bibitem{mousavi2015deep}
\begin{bchapter}
\bauthor{\bsnm{Mousavi}, \binits{A.}},
\bauthor{\bsnm{Patel}, \binits{A.B.}},
\bauthor{\bsnm{Baraniuk}, \binits{R.G.}}:
\bctitle{{A Deep Learning Approach to Structured Signal Recovery}}.
In: \bbtitle{{Proceedings of IEEE Allerton Conference on Communication,
  Control, and Computing}},
pp. \bfpage{1336}--\blpage{1343}
(\byear{2015})
\end{bchapter}
\endbibitem

\bibitem{shi2019image}
\begin{barticle}
\bauthor{\bsnm{Shi}, \binits{W.}},
\bauthor{\bsnm{Jiang}, \binits{F.}},
\bauthor{\bsnm{Liu}, \binits{S.}},
\bauthor{\bsnm{Zhao}, \binits{D.}}:
\batitle{{Image Compressed Sensing Using Convolutional Neural Network}}.
\bjtitle{{IEEE Transactions on Image Processing}}
\bvolume{29},
\bfpage{375}--\blpage{388}
(\byear{2019})
\end{barticle}
\endbibitem

\bibitem{sun2020dual}
\begin{barticle}
\bauthor{\bsnm{Sun}, \binits{Y.}},
\bauthor{\bsnm{Chen}, \binits{J.}},
\bauthor{\bsnm{Liu}, \binits{Q.}},
\bauthor{\bsnm{Liu}, \binits{B.}},
\bauthor{\bsnm{Guo}, \binits{G.}}:
\batitle{{Dual-Path Attention Network for Compressed Sensing Image
  Reconstruction}}.
\bjtitle{{IEEE Transactions on Image Processing}}
\bvolume{29},
\bfpage{9482}--\blpage{9495}
(\byear{2020})
\end{barticle}
\endbibitem

\bibitem{shi2019scalable}
\begin{bchapter}
\bauthor{\bsnm{Shi}, \binits{W.}},
\bauthor{\bsnm{Jiang}, \binits{F.}},
\bauthor{\bsnm{Liu}, \binits{S.}},
\bauthor{\bsnm{Zhao}, \binits{D.}}:
\bctitle{{Scalable Convolutional Neural Network for Image Compressed Sensing}}.
In: \bbtitle{{Proceedings of IEEE Conference on Computer Vision and Pattern
  Recognition (CVPR)}},
pp. \bfpage{12290}--\blpage{12299}
(\byear{2019})
\end{bchapter}
\endbibitem

\bibitem{fan2022global}
\begin{bchapter}
\bauthor{\bsnm{Fan}, \binits{Z.-E.}},
\bauthor{\bsnm{Lian}, \binits{F.}},
\bauthor{\bsnm{Quan}, \binits{J.-N.}}:
\bctitle{{Global Sensing and Measurements Reuse for Image Compressed Sensing}}.
In: \bbtitle{{Proceedings of IEEE Conference on Computer Vision and Pattern
  Recognition (CVPR)}},
pp. \bfpage{8954}--\blpage{8963}
(\byear{2022})
\end{bchapter}
\endbibitem

\bibitem{huang2018some}
\begin{bchapter}
\bauthor{\bsnm{Huang}, \binits{Y.}},
\bauthor{\bsnm{W{\"u}rfl}, \binits{T.}},
\bauthor{\bsnm{Breininger}, \binits{K.}},
\bauthor{\bsnm{Liu}, \binits{L.}},
\bauthor{\bsnm{Lauritsch}, \binits{G.}},
\bauthor{\bsnm{Maier}, \binits{A.}}:
\bctitle{{Some Investigations on Robustness of Deep Learning in Limited Angle
  Tomography}}.
In: \bbtitle{{Proceedings of International Conference on Medical Image
  Computing and Computer-Assisted Intervention (MICCAI)}},
pp. \bfpage{145}--\blpage{153}
(\byear{2018})
\end{bchapter}
\endbibitem

\bibitem{sun2016deep}
\begin{barticle}
\bauthor{\bsnm{Sun}, \binits{J.}},
\bauthor{\bsnm{Li}, \binits{H.}},
\bauthor{\bsnm{Xu}, \binits{Z.}}, \betal:
\batitle{{Deep ADMM-Net for Compressive Sensing MRI}}.
\bjtitle{{Proceedings of Neural Information Processing Systems (NeurIPS)}}
\bvolume{29},
\bfpage{10}--\blpage{18}
(\byear{2016})
\end{barticle}
\endbibitem

\bibitem{zhang2018ista}
\begin{bchapter}
\bauthor{\bsnm{Zhang}, \binits{J.}},
\bauthor{\bsnm{Ghanem}, \binits{B.}}:
\bctitle{{ISTA-Net: Interpretable Optimization-Inspired Deep Network for Image
  Compressive Sensing}}.
In: \bbtitle{{Proceedings of IEEE Conference on Computer Vision and Pattern
  Recognition (CVPR)}},
pp. \bfpage{1828}--\blpage{1837}
(\byear{2018})
\end{bchapter}
\endbibitem

\bibitem{zhang2020optimization}
\begin{barticle}
\bauthor{\bsnm{Zhang}, \binits{J.}},
\bauthor{\bsnm{Zhao}, \binits{C.}},
\bauthor{\bsnm{Gao}, \binits{W.}}:
\batitle{{Optimization-Inspired Compact Deep Compressive Sensing}}.
\bjtitle{{IEEE Journal of Selected Topics in Signal Processing}}
\bvolume{14}(\bissue{4}),
\bfpage{765}--\blpage{774}
(\byear{2020})
\end{barticle}
\endbibitem

\bibitem{zhang2020amp}
\begin{barticle}
\bauthor{\bsnm{Zhang}, \binits{Z.}},
\bauthor{\bsnm{Liu}, \binits{Y.}},
\bauthor{\bsnm{Liu}, \binits{J.}},
\bauthor{\bsnm{Wen}, \binits{F.}},
\bauthor{\bsnm{Zhu}, \binits{C.}}:
\batitle{{AMP-Net: Denoising-Based Deep Unfolding for Compressive Image
  Sensing}}.
\bjtitle{{IEEE Transactions on Image Processing}}
\bvolume{30},
\bfpage{1487}--\blpage{1500}
(\byear{2020})
\end{barticle}
\endbibitem

\bibitem{zhang2023physics}
\begin{barticle}
\bauthor{\bsnm{Zhang}, \binits{J.}},
\bauthor{\bsnm{Chen}, \binits{B.}},
\bauthor{\bsnm{Xiong}, \binits{R.}},
\bauthor{\bsnm{Zhang}, \binits{Y.}}:
\batitle{{Physics-Inspired Compressive Sensing: Beyond deep unrolling}}.
\bjtitle{IEEE Signal Processing Magazine}
\bvolume{40}(\bissue{1}),
\bfpage{58}--\blpage{72}
(\byear{2023})
\end{barticle}
\endbibitem

\bibitem{gilton2019neumann}
\begin{barticle}
\bauthor{\bsnm{Gilton}, \binits{D.}},
\bauthor{\bsnm{Ongie}, \binits{G.}},
\bauthor{\bsnm{Willett}, \binits{R.}}:
\batitle{{Neumann Networks for Linear Inverse Problems in Imaging}}.
\bjtitle{{IEEE Transactions on Computational Imaging}}
\bvolume{6},
\bfpage{328}--\blpage{343}
(\byear{2019})
\end{barticle}
\endbibitem

\bibitem{chen2020learning}
\begin{bchapter}
\bauthor{\bsnm{Chen}, \binits{J.}},
\bauthor{\bsnm{Sun}, \binits{Y.}},
\bauthor{\bsnm{Liu}, \binits{Q.}},
\bauthor{\bsnm{Huang}, \binits{R.}}:
\bctitle{{Learning Memory Augmented Cascading Network for Compressed Sensing of
  Images}}.
In: \bbtitle{{Proceedings of European Conference on Computer Vision (ECCV)}},
pp. \bfpage{513}--\blpage{529}
(\byear{2020})
\end{bchapter}
\endbibitem

\bibitem{song2021memory}
\begin{bchapter}
\bauthor{\bsnm{Song}, \binits{J.}},
\bauthor{\bsnm{Chen}, \binits{B.}},
\bauthor{\bsnm{Zhang}, \binits{J.}}:
\bctitle{Memory-augmented deep unfolding network for compressive sensing}.
In: \bbtitle{{Proceedings of ACM International Conference on Multimedia (ACM
  MM)}},
pp. \bfpage{4249}--\blpage{4258}
(\byear{2021})
\end{bchapter}
\endbibitem

\bibitem{you2021coast}
\begin{barticle}
\bauthor{\bsnm{You}, \binits{D.}},
\bauthor{\bsnm{Zhang}, \binits{J.}},
\bauthor{\bsnm{Xie}, \binits{J.}},
\bauthor{\bsnm{Chen}, \binits{B.}},
\bauthor{\bsnm{Ma}, \binits{S.}}:
\batitle{{COAST: COntrollable Arbitrary-Sampling NeTwork for Compressive
  Sensing}}.
\bjtitle{{IEEE Transactions on Image Processing}}
\bvolume{30},
\bfpage{6066}--\blpage{6080}
(\byear{2021})
\end{barticle}
\endbibitem

\bibitem{you2021ista}
\begin{bchapter}
\bauthor{\bsnm{You}, \binits{D.}},
\bauthor{\bsnm{Xie}, \binits{J.}},
\bauthor{\bsnm{Zhang}, \binits{J.}}:
\bctitle{{ISTA-Net$^{++}$: Flexible Deep Unfolding Network for Compressive
  Sensing}}.
In: \bbtitle{{Proceedings of IEEE International Conference on Multimedia and
  Expo (ICME)}},
pp. \bfpage{1}--\blpage{6}
(\byear{2021})
\end{bchapter}
\endbibitem

\bibitem{chen2021deep}
\begin{barticle}
\bauthor{\bsnm{Chen}, \binits{Z.}},
\bauthor{\bsnm{Guo}, \binits{W.}},
\bauthor{\bsnm{Feng}, \binits{Y.}},
\bauthor{\bsnm{Li}, \binits{Y.}},
\bauthor{\bsnm{Zhao}, \binits{C.}},
\bauthor{\bsnm{Ren}, \binits{Y.}},
\bauthor{\bsnm{Shao}, \binits{L.}}:
\batitle{{Deep-Learned Regularization and Proximal Operator for Image
  Compressive Sensing}}.
\bjtitle{{IEEE Transactions on Image Processing}}
\bvolume{30},
\bfpage{7112}--\blpage{7126}
(\byear{2021})
\end{barticle}
\endbibitem

\bibitem{he2016deep}
\begin{bchapter}
\bauthor{\bsnm{He}, \binits{K.}},
\bauthor{\bsnm{Zhang}, \binits{X.}},
\bauthor{\bsnm{Ren}, \binits{S.}},
\bauthor{\bsnm{Sun}, \binits{J.}}:
\bctitle{{Deep Residual Learning for Image Recognition}}.
In: \bbtitle{{Proceedings of IEEE Conference on Computer Vision and Pattern
  Recognition (CVPR)}},
pp. \bfpage{770}--\blpage{778}
(\byear{2016})
\end{bchapter}
\endbibitem

\bibitem{ronneberger2015u}
\begin{bchapter}
\bauthor{\bsnm{Ronneberger}, \binits{O.}},
\bauthor{\bsnm{Fischer}, \binits{P.}},
\bauthor{\bsnm{Brox}, \binits{T.}}:
\bctitle{{U-Net: Convolutional Networks for Biomedical Image Segmentation}}.
In: \bbtitle{{Proceedings of International Conference on Medical Image
  Computing and Computer-Assisted Intervention (MICCAI)}},
pp. \bfpage{234}--\blpage{241}
(\byear{2015})
\end{bchapter}
\endbibitem

\bibitem{zhang2021plug}
\begin{botherref}
\oauthor{\bsnm{Zhang}, \binits{K.}},
\oauthor{\bsnm{Li}, \binits{Y.}},
\oauthor{\bsnm{Zuo}, \binits{W.}},
\oauthor{\bsnm{Zhang}, \binits{L.}},
\oauthor{\bsnm{Van~Gool}, \binits{L.}},
\oauthor{\bsnm{Timofte}, \binits{R.}}:
{Plug-and-Play Image Restoration with Deep Denoiser Prior}.
{IEEE Transactions on Pattern Analysis and Machine Intelligence}
(2021)
\end{botherref}
\endbibitem

\bibitem{parikh2014proximal}
\begin{barticle}
\bauthor{\bsnm{Parikh}, \binits{N.}},
\bauthor{\bsnm{Boyd}, \binits{S.}}, \betal:
\batitle{{Proximal Algorithms}}.
\bjtitle{{Foundations and in Optimization}}
\bvolume{1}(\bissue{3}),
\bfpage{127}--\blpage{239}
(\byear{2014})
\end{barticle}
\endbibitem

\bibitem{lefkimmiatis2018universal}
\begin{bchapter}
\bauthor{\bsnm{Lefkimmiatis}, \binits{S.}}:
\bctitle{{Universal Denoising Networks: A Novel CNN Architecture for Image
  Denoising}}.
In: \bbtitle{{Proceedings of IEEE Conference on Computer Vision and Pattern
  Recognition (CVPR)}},
pp. \bfpage{3204}--\blpage{3213}
(\byear{2018})
\end{bchapter}
\endbibitem

\bibitem{chen2016trainable}
\begin{barticle}
\bauthor{\bsnm{Chen}, \binits{Y.}},
\bauthor{\bsnm{Pock}, \binits{T.}}:
\batitle{{Trainable Nonlinear Reaction Diffusion: A Flexible Framework for Fast
  and Effective Image Restoration}}.
\bjtitle{{IEEE Transactions on Pattern Analysis and Machine Intelligence}}
\bvolume{39}(\bissue{6}),
\bfpage{1256}--\blpage{1272}
(\byear{2016})
\end{barticle}
\endbibitem

\bibitem{lefkimmiatis2017non}
\begin{bchapter}
\bauthor{\bsnm{Lefkimmiatis}, \binits{S.}}:
\bctitle{{Non-Local Color Image Denoising with Convolutional Neural Networks}}.
In: \bbtitle{{Proceedings of IEEE Conference on Computer Vision and Pattern
  Recognition (CVPR)}},
pp. \bfpage{3587}--\blpage{3596}
(\byear{2017})
\end{bchapter}
\endbibitem

\bibitem{kruse2017learning}
\begin{bchapter}
\bauthor{\bsnm{Kruse}, \binits{J.}},
\bauthor{\bsnm{Rother}, \binits{C.}},
\bauthor{\bsnm{Schmidt}, \binits{U.}}:
\bctitle{{Learning to Push the Limits of Efficient FFT-Based Image
  Deconvolution}}.
In: \bbtitle{{Proceedings of IEEE International Conference on Computer Vision
  (ICCV)}},
pp. \bfpage{4586}--\blpage{4594}
(\byear{2017})
\end{bchapter}
\endbibitem

\bibitem{wang2020stacking}
\begin{bchapter}
\bauthor{\bsnm{Wang}, \binits{H.}},
\bauthor{\bsnm{Zhang}, \binits{T.}},
\bauthor{\bsnm{Yu}, \binits{M.}},
\bauthor{\bsnm{Sun}, \binits{J.}},
\bauthor{\bsnm{Ye}, \binits{W.}},
\bauthor{\bsnm{Wang}, \binits{C.}},
\bauthor{\bsnm{Zhang}, \binits{S.}}:
\bctitle{{Stacking Networks Dynamically for Image Restoration Based on the
  Plug-and-Play Framework}}.
In: \bbtitle{{Proceedings of European Conference on Computer Vision (ECCV)}},
pp. \bfpage{446}--\blpage{462}
(\byear{2020})
\end{bchapter}
\endbibitem

\bibitem{kokkinos2018deep}
\begin{bchapter}
\bauthor{\bsnm{Kokkinos}, \binits{F.}},
\bauthor{\bsnm{Lefkimmiatis}, \binits{S.}}:
\bctitle{{Deep Image Demosaicking Using a Cascade of Convolutional Residual
  Denoising Networks}}.
In: \bbtitle{{Proceedings of European Conference on Computer Vision (ECCV)}},
pp. \bfpage{303}--\blpage{319}
(\byear{2018})
\end{bchapter}
\endbibitem

\bibitem{zhang2020deep}
\begin{bchapter}
\bauthor{\bsnm{Zhang}, \binits{K.}},
\bauthor{\bsnm{Gool}, \binits{L.V.}},
\bauthor{\bsnm{Timofte}, \binits{R.}}:
\bctitle{{Deep Unfolding Network for Image Super-Resolution}}.
In: \bbtitle{{Proceedings of IEEE Conference on Computer Vision and Pattern
  Recognition (CVPR)}},
pp. \bfpage{3217}--\blpage{3226}
(\byear{2020})
\end{bchapter}
\endbibitem

\bibitem{blumensath2009iterative}
\begin{barticle}
\bauthor{\bsnm{Blumensath}, \binits{T.}},
\bauthor{\bsnm{Davies}, \binits{M.E.}}:
\batitle{{Iterative Hard Thresholding for Compressed Sensing}}.
\bjtitle{{Applied and Computational Harmonic Analysis}}
\bvolume{27}(\bissue{3}),
\bfpage{265}--\blpage{274}
(\byear{2009})
\end{barticle}
\endbibitem

\bibitem{song2023deep}
\begin{botherref}
\oauthor{\bsnm{Song}, \binits{J.}},
\oauthor{\bsnm{Chen}, \binits{B.}},
\oauthor{\bsnm{Zhang}, \binits{J.}}:
{Deep Memory-Augmented Proximal Unrolling Network for Compressive Sensing}.
International Journal of Computer Vision,
1--20
(2023)
\end{botherref}
\endbibitem

\bibitem{zhang2014group}
\begin{barticle}
\bauthor{\bsnm{Zhang}, \binits{J.}},
\bauthor{\bsnm{Zhao}, \binits{D.}},
\bauthor{\bsnm{Gao}, \binits{W.}}:
\batitle{{Group-Based Sparse Representation for Image Restoration}}.
\bjtitle{{IEEE Transactions on Image Processing}}
\bvolume{23}(\bissue{8}),
\bfpage{3336}--\blpage{3351}
(\byear{2014})
\end{barticle}
\endbibitem

\bibitem{mousavi2017learning}
\begin{bchapter}
\bauthor{\bsnm{Mousavi}, \binits{A.}},
\bauthor{\bsnm{Baraniuk}, \binits{R.G.}}:
\bctitle{{Learning to Invert: Signal Recovery via Deep Convolutional
  Networks}}.
In: \bbtitle{{Proceedings of IEEE International Conference on Acoustics, Speech
  and Signal Processing (ICASSP)}},
pp. \bfpage{2272}--\blpage{2276}
(\byear{2017})
\end{bchapter}
\endbibitem

\bibitem{dabov2007image}
\begin{barticle}
\bauthor{\bsnm{Dabov}, \binits{K.}},
\bauthor{\bsnm{Foi}, \binits{A.}},
\bauthor{\bsnm{Katkovnik}, \binits{V.}},
\bauthor{\bsnm{Egiazarian}, \binits{K.}}:
\batitle{{Image Denoising by Sparse 3-D Transform-Domain Collaborative
  Filtering}}.
\bjtitle{{IEEE Transactions on Image Processing}}
\bvolume{16}(\bissue{8}),
\bfpage{2080}--\blpage{2095}
(\byear{2007})
\end{barticle}
\endbibitem

\bibitem{ren2021adaptive}
\begin{bchapter}
\bauthor{\bsnm{Ren}, \binits{C.}},
\bauthor{\bsnm{He}, \binits{X.}},
\bauthor{\bsnm{Wang}, \binits{C.}},
\bauthor{\bsnm{Zhao}, \binits{Z.}}:
\bctitle{{Adaptive Consistency Prior Based Deep Network for Image Denoising}}.
In: \bbtitle{{Proceedings of IEEE Conference on Computer Vision and Pattern
  Recognition (CVPR)}},
pp. \bfpage{8596}--\blpage{8606}
(\byear{2021})
\end{bchapter}
\endbibitem

\bibitem{zheng2021deep}
\begin{bchapter}
\bauthor{\bsnm{Zheng}, \binits{H.}},
\bauthor{\bsnm{Yong}, \binits{H.}},
\bauthor{\bsnm{Zhang}, \binits{L.}}:
\bctitle{{Deep Convolutional Dictionary Learning for Image Denoising}}.
In: \bbtitle{{Proceedings of IEEE Conference on Computer Vision and Pattern
  Recognition (CVPR)}},
pp. \bfpage{630}--\blpage{641}
(\byear{2021})
\end{bchapter}
\endbibitem

\bibitem{wu2021dense}
\begin{bchapter}
\bauthor{\bsnm{Wu}, \binits{Z.}},
\bauthor{\bsnm{Zhang}, \binits{J.}},
\bauthor{\bsnm{Mou}, \binits{C.}}:
\bctitle{{Dense Deep Unfolding Network with 3D-CNN Prior for Snapshot
  Compressive Imaging}}.
In: \bbtitle{{Proceedings of IEEE International Conference on Computer Vision
  (ICCV)}},
pp. \bfpage{4892}--\blpage{4901}
(\byear{2021})
\end{bchapter}
\endbibitem

\bibitem{zamir2021multi}
\begin{bchapter}
\bauthor{\bsnm{Zamir}, \binits{S.W.}},
\bauthor{\bsnm{Arora}, \binits{A.}},
\bauthor{\bsnm{Khan}, \binits{S.}},
\bauthor{\bsnm{Hayat}, \binits{M.}},
\bauthor{\bsnm{Khan}, \binits{F.S.}},
\bauthor{\bsnm{Yang}, \binits{M.-H.}},
\bauthor{\bsnm{Shao}, \binits{L.}}:
\bctitle{{Multi-Stage Progressive Image Restoration}}.
In: \bbtitle{{Proceedings of IEEE Conference on Computer Vision and Pattern
  Recognition (CVPR)}},
pp. \bfpage{14821}--\blpage{14831}
(\byear{2021})
\end{bchapter}
\endbibitem

\bibitem{shi2016real}
\begin{bchapter}
\bauthor{\bsnm{Shi}, \binits{W.}},
\bauthor{\bsnm{Caballero}, \binits{J.}},
\bauthor{\bsnm{Husz{\'a}r}, \binits{F.}},
\bauthor{\bsnm{Totz}, \binits{J.}},
\bauthor{\bsnm{Aitken}, \binits{A.P.}},
\bauthor{\bsnm{Bishop}, \binits{R.}},
\bauthor{\bsnm{Rueckert}, \binits{D.}},
\bauthor{\bsnm{Wang}, \binits{Z.}}:
\bctitle{{Real-Time Single Image and Video Super-Resolution Using An Efficient
  Sub-Pixel Convolutional Neural Network}}.
In: \bbtitle{{Proceedings of IEEE Conference on Computer Vision and Pattern
  Recognition (CVPR)}},
pp. \bfpage{1874}--\blpage{1883}
(\byear{2016})
\end{bchapter}
\endbibitem

\bibitem{huang2015single}
\begin{bchapter}
\bauthor{\bsnm{Huang}, \binits{J.-B.}},
\bauthor{\bsnm{Singh}, \binits{A.}},
\bauthor{\bsnm{Ahuja}, \binits{N.}}:
\bctitle{{Single Image Super-Resolution from Transformed Self-Exemplars}}.
In: \bbtitle{{Proceedings of IEEE Conference on Computer Vision and Pattern
  Recognition (CVPR)}},
pp. \bfpage{5197}--\blpage{5206}
(\byear{2015})
\end{bchapter}
\endbibitem

\bibitem{yang2008image}
\begin{bchapter}
\bauthor{\bsnm{Yang}, \binits{J.}},
\bauthor{\bsnm{Wright}, \binits{J.}},
\bauthor{\bsnm{Huang}, \binits{T.}},
\bauthor{\bsnm{Ma}, \binits{Y.}}:
\bctitle{{Image Super-Resolution as Sparse Representation of Raw Image
  Patches}}.
In: \bbtitle{{Proceedings of IEEE Conference on Computer Vision and Pattern
  Recognition (CVPR)}},
pp. \bfpage{1}--\blpage{8}
(\byear{2008})
\end{bchapter}
\endbibitem

\bibitem{dong2014learning}
\begin{bchapter}
\bauthor{\bsnm{Dong}, \binits{C.}},
\bauthor{\bsnm{Loy}, \binits{C.C.}},
\bauthor{\bsnm{He}, \binits{K.}},
\bauthor{\bsnm{Tang}, \binits{X.}}:
\bctitle{{Learning a Deep Convolutional Network for Image Super-Resolution}}.
In: \bbtitle{{Proceedings of European Conference on Computer Vision (ECCV)}},
pp. \bfpage{184}--\blpage{199}
(\byear{2014})
\end{bchapter}
\endbibitem

\bibitem{zhang2017beyond}
\begin{barticle}
\bauthor{\bsnm{Zhang}, \binits{K.}},
\bauthor{\bsnm{Zuo}, \binits{W.}},
\bauthor{\bsnm{Chen}, \binits{Y.}},
\bauthor{\bsnm{Meng}, \binits{D.}},
\bauthor{\bsnm{Zhang}, \binits{L.}}:
\batitle{{Beyond a Gaussian Denoiser: Residual Learning of Deep CNN for Image
  Denoising}}.
\bjtitle{{IEEE Transactions on Image Processing}}
\bvolume{26}(\bissue{7}),
\bfpage{3142}--\blpage{3155}
(\byear{2017})
\end{barticle}
\endbibitem

\bibitem{agustsson2017ntire}
\begin{bchapter}
\bauthor{\bsnm{Agustsson}, \binits{E.}},
\bauthor{\bsnm{Timofte}, \binits{R.}}:
\bctitle{{NTIRE 2017 Challenge on Single Image Super-Resolution: Dataset and
  Study}}.
In: \bbtitle{{Proceedings of IEEE Conference on Computer Vision and Pattern
  Recognition Workshops (CVPRW)}},
pp. \bfpage{126}--\blpage{135}
(\byear{2017})
\end{bchapter}
\endbibitem

\bibitem{ma2016waterloo}
\begin{barticle}
\bauthor{\bsnm{Ma}, \binits{K.}},
\bauthor{\bsnm{Duanmu}, \binits{Z.}},
\bauthor{\bsnm{Wu}, \binits{Q.}},
\bauthor{\bsnm{Wang}, \binits{Z.}},
\bauthor{\bsnm{Yong}, \binits{H.}},
\bauthor{\bsnm{Li}, \binits{H.}},
\bauthor{\bsnm{Zhang}, \binits{L.}}:
\batitle{{Waterloo Exploration Database: New Challenges for Image Quality
  Assessment Models}}.
\bjtitle{{IEEE Transactions on Image Processing}}
\bvolume{26}(\bissue{2}),
\bfpage{1004}--\blpage{1016}
(\byear{2016})
\end{barticle}
\endbibitem

\bibitem{martin2001database}
\begin{bchapter}
\bauthor{\bsnm{Martin}, \binits{D.}},
\bauthor{\bsnm{Fowlkes}, \binits{C.}},
\bauthor{\bsnm{Tal}, \binits{D.}},
\bauthor{\bsnm{Malik}, \binits{J.}}:
\bctitle{{A Database of Human Segmented Natural Images and Its Application to
  Evaluating Segmentation Algorithms and Measuring Ecological Statistics}}.
In: \bbtitle{{Proceedings of IEEE International Conference on Computer Vision
  (ICCV)}},
vol. \bseriesno{2},
pp. \bfpage{416}--\blpage{423}
(\byear{2001})
\end{bchapter}
\endbibitem

\bibitem{wang2004image}
\begin{barticle}
\bauthor{\bsnm{Wang}, \binits{Z.}},
\bauthor{\bsnm{Bovik}, \binits{A.C.}},
\bauthor{\bsnm{Sheikh}, \binits{H.R.}},
\bauthor{\bsnm{Simoncelli}, \binits{E.P.}}:
\batitle{{Image Quality Assessment: From Error Visibility to Structural
  Similarity}}.
\bjtitle{{IEEE Transactions on Image Processing}}
\bvolume{13}(\bissue{4}),
\bfpage{600}--\blpage{612}
(\byear{2004})
\end{barticle}
\endbibitem

\bibitem{paszke2019pytorch}
\begin{botherref}
\oauthor{\bsnm{Paszke}, \binits{A.}},
\oauthor{\bsnm{Gross}, \binits{S.}},
\oauthor{\bsnm{Massa}, \binits{F.}},
\oauthor{\bsnm{Lerer}, \binits{A.}},
\oauthor{\bsnm{Bradbury}, \binits{J.}},
\oauthor{\bsnm{Chanan}, \binits{G.}},
\oauthor{\bsnm{Killeen}, \binits{T.}},
\oauthor{\bsnm{Lin}, \binits{Z.}},
\oauthor{\bsnm{Gimelshein}, \binits{N.}},
\oauthor{\bsnm{Antiga}, \binits{L.}}, et al.:
{PyTorch: An Imperative Style, High-Performance Deep Learning Library}.
{Proceedings of Neural Information Processing Systems (NeurIPS)}
\textbf{32}
(2019)
\end{botherref}
\endbibitem

\bibitem{kingma2014adam}
\begin{bchapter}
\bauthor{\bsnm{Kingma}, \binits{D.P.}},
\bauthor{\bsnm{Ba}, \binits{J.}}:
\bctitle{{Adam: A Method for Stochastic Optimization}}.
In: \bbtitle{{Proceedings of International Conference on Learning
  Representations (ICLR)}},
pp. \bfpage{1}--\blpage{15}
(\byear{2015})
\end{bchapter}
\endbibitem

\bibitem{tian2020attention}
\begin{barticle}
\bauthor{\bsnm{Tian}, \binits{C.}},
\bauthor{\bsnm{Xu}, \binits{Y.}},
\bauthor{\bsnm{Li}, \binits{Z.}},
\bauthor{\bsnm{Zuo}, \binits{W.}},
\bauthor{\bsnm{Fei}, \binits{L.}},
\bauthor{\bsnm{Liu}, \binits{H.}}:
\batitle{{Attention-Guided CNN for Image Denoising}}.
\bjtitle{{Neural Networks}}
\bvolume{124},
\bfpage{117}--\blpage{129}
(\byear{2020})
\end{barticle}
\endbibitem

\bibitem{coban2020apple}
\begin{botherref}
\oauthor{\bsnm{Coban}, \binits{S.}},
\oauthor{\bsnm{Andriiashen}, \binits{V.}},
\oauthor{\bsnm{Ganguly}, \binits{P.}}:
{Apple CT Data: Simulated Parallel-Beam Tomographic Datasets}.
{Zenodo}
(2020)
\end{botherref}
\endbibitem

\bibitem{zhang2018image}
\begin{bchapter}
\bauthor{\bsnm{Zhang}, \binits{Y.}},
\bauthor{\bsnm{Li}, \binits{K.}},
\bauthor{\bsnm{Li}, \binits{K.}},
\bauthor{\bsnm{Wang}, \binits{L.}},
\bauthor{\bsnm{Zhong}, \binits{B.}},
\bauthor{\bsnm{Fu}, \binits{Y.}}:
\bctitle{{Image Super-Resolution Using Very Deep Residual Channel Attention
  Networks}}.
In: \bbtitle{{Proceedings of European Conference on Computer Vision (ECCV)}},
pp. \bfpage{286}--\blpage{301}
(\byear{2018})
\end{bchapter}
\endbibitem

\bibitem{radon1986determination}
\begin{barticle}
\bauthor{\bsnm{Radon}, \binits{J.}}:
\batitle{{On the Determination of Functions from Their Integral Values Along
  Certain Manifolds}}.
\bjtitle{{IEEE Transactions on Medical Imaging}}
\bvolume{5}(\bissue{4}),
\bfpage{170}--\blpage{176}
(\byear{1986})
\end{barticle}
\endbibitem

\bibitem{leuschner2021quantitative}
\begin{barticle}
\bauthor{\bsnm{Leuschner}, \binits{J.}},
\bauthor{\bsnm{Schmidt}, \binits{M.}},
\bauthor{\bsnm{Ganguly}, \binits{P.S.}},
\bauthor{\bsnm{Andriiashen}, \binits{V.}},
\bauthor{\bsnm{Coban}, \binits{S.B.}},
\bauthor{\bsnm{Denker}, \binits{A.}},
\bauthor{\bsnm{Bauer}, \binits{D.}},
\bauthor{\bsnm{Hadjifaradji}, \binits{A.}},
\bauthor{\bsnm{Batenburg}, \binits{K.J.}},
\bauthor{\bsnm{Maass}, \binits{P.}}, \betal:
\batitle{{Quantitative Comparison of Deep Learning-Based Image Reconstruction
  Methods for Low-Dose and Sparse-Angle CT Applications}}.
\bjtitle{{Journal of Imaging}}
\bvolume{7}(\bissue{3}),
\bfpage{44}
(\byear{2021})
\end{barticle}
\endbibitem

\bibitem{bjorck1998stability}
\begin{barticle}
\bauthor{\bsnm{Bj{\"o}rck}, \binits{{\AA}.}},
\bauthor{\bsnm{Elfving}, \binits{T.}},
\bauthor{\bsnm{Strakos}, \binits{Z.}}:
\batitle{{Stability of Conjugate Gradient and Lanczos Methods for Linear Least
  Squares Problems}}.
\bjtitle{{SIAM Journal on Matrix Analysis and Applications}}
\bvolume{19}(\bissue{3}),
\bfpage{720}--\blpage{736}
(\byear{1998})
\end{barticle}
\endbibitem

\bibitem{niu2014sparse}
\begin{barticle}
\bauthor{\bsnm{Niu}, \binits{S.}},
\bauthor{\bsnm{Gao}, \binits{Y.}},
\bauthor{\bsnm{Bian}, \binits{Z.}},
\bauthor{\bsnm{Huang}, \binits{J.}},
\bauthor{\bsnm{Chen}, \binits{W.}},
\bauthor{\bsnm{Yu}, \binits{G.}},
\bauthor{\bsnm{Liang}, \binits{Z.}},
\bauthor{\bsnm{Ma}, \binits{J.}}:
\batitle{{Sparse-View X-Ray CT Reconstruction via Total Generalized Variation
  Regularization}}.
\bjtitle{{Physics in Medicine \& Biology}}
\bvolume{59}(\bissue{12}),
\bfpage{2997}
(\byear{2014})
\end{barticle}
\endbibitem

\bibitem{li2019learning}
\begin{barticle}
\bauthor{\bsnm{Li}, \binits{Y.}},
\bauthor{\bsnm{Li}, \binits{K.}},
\bauthor{\bsnm{Zhang}, \binits{C.}},
\bauthor{\bsnm{Montoya}, \binits{J.}},
\bauthor{\bsnm{Chen}, \binits{G.-H.}}:
\batitle{{Learning to Reconstruct Computed Tomography Images Directly from
  Sinogram Data under a Variety of Data Acquisition Conditions}}.
\bjtitle{{IEEE Transactions on Medical Imaging}}
\bvolume{38}(\bissue{10}),
\bfpage{2469}--\blpage{2481}
(\byear{2019})
\end{barticle}
\endbibitem

\bibitem{denker2020conditional}
\begin{botherref}
\oauthor{\bsnm{Denker}, \binits{A.}},
\oauthor{\bsnm{Schmidt}, \binits{M.}},
\oauthor{\bsnm{Leuschner}, \binits{J.}},
\oauthor{\bsnm{Maass}, \binits{P.}},
\oauthor{\bsnm{Behrmann}, \binits{J.}}:
{Conditional Normalizing Flows for Low-Dose Computed Tomography Image
  Reconstruction}.
arXiv preprint arXiv:2006.06270
(2020)
\end{botherref}
\endbibitem

\bibitem{pelt2018improving}
\begin{barticle}
\bauthor{\bsnm{Pelt}, \binits{D.M.}},
\bauthor{\bsnm{Batenburg}, \binits{K.J.}},
\bauthor{\bsnm{Sethian}, \binits{J.A.}}:
\batitle{Improving tomographic reconstruction from limited data using
  mixed-scale dense convolutional neural networks}.
\bjtitle{{Journal of Imaging}}
\bvolume{4}(\bissue{11}),
\bfpage{128}
(\byear{2018})
\end{barticle}
\endbibitem

\bibitem{chen2017low}
\begin{barticle}
\bauthor{\bsnm{Chen}, \binits{H.}},
\bauthor{\bsnm{Zhang}, \binits{Y.}},
\bauthor{\bsnm{Zhang}, \binits{W.}},
\bauthor{\bsnm{Liao}, \binits{P.}},
\bauthor{\bsnm{Li}, \binits{K.}},
\bauthor{\bsnm{Zhou}, \binits{J.}},
\bauthor{\bsnm{Wang}, \binits{G.}}:
\batitle{{Low-Dose CT via Convolutional Neural Network}}.
\bjtitle{{Biomedical Optics Express}}
\bvolume{8}(\bissue{2}),
\bfpage{679}--\blpage{694}
(\byear{2017})
\end{barticle}
\endbibitem

\bibitem{liu2020interpreting}
\begin{botherref}
\oauthor{\bsnm{Liu}, \binits{T.}},
\oauthor{\bsnm{Chaman}, \binits{A.}},
\oauthor{\bsnm{Belius}, \binits{D.}},
\oauthor{\bsnm{Dokmanic}, \binits{I.}}:
{Interpreting U-Nets via Task-Driven Multiscale Dictionary Learning}.
arXiv preprint arXiv:2011.12815
(2020)
\end{botherref}
\endbibitem

\bibitem{adler2018learned}
\begin{barticle}
\bauthor{\bsnm{Adler}, \binits{J.}},
\bauthor{\bsnm{{\"O}ktem}, \binits{O.}}:
\batitle{{Learned Primal-Dual Reconstruction}}.
\bjtitle{{IEEE Transactions on Medical Imaging}}
\bvolume{37}(\bissue{6}),
\bfpage{1322}--\blpage{1332}
(\byear{2018})
\end{barticle}
\endbibitem

\bibitem{xiang2021fista}
\begin{barticle}
\bauthor{\bsnm{Xiang}, \binits{J.}},
\bauthor{\bsnm{Dong}, \binits{Y.}},
\bauthor{\bsnm{Yang}, \binits{Y.}}:
\batitle{{FISTA-Net: Learning A Fast Iterative Shrinkage Thresholding Network
  for Inverse Problems in Imaging}}.
\bjtitle{{IEEE Transactions on Medical Imaging}}
\bvolume{40}(\bissue{5}),
\bfpage{1329}--\blpage{1339}
(\byear{2021})
\end{barticle}
\endbibitem

\bibitem{boufounos20081}
\begin{bchapter}
\bauthor{\bsnm{Boufounos}, \binits{P.T.}},
\bauthor{\bsnm{Baraniuk}, \binits{R.G.}}:
\bctitle{{1-bit Compressive Sensing}}.
In: \bbtitle{{Proceedings of IEEE Conference on Information Sciences and
  Systems (CISS)}},
pp. \bfpage{16}--\blpage{21}
(\byear{2008})
\end{bchapter}
\endbibitem

\bibitem{jacques2013robust}
\begin{barticle}
\bauthor{\bsnm{Jacques}, \binits{L.}},
\bauthor{\bsnm{Laska}, \binits{J.N.}},
\bauthor{\bsnm{Boufounos}, \binits{P.T.}},
\bauthor{\bsnm{Baraniuk}, \binits{R.G.}}:
\batitle{{Robust 1-bit Compressive Sensing via Binary Stable Embeddings of
  Sparse Vectors}}.
\bjtitle{{IEEE Transactions on Information Theory}}
\bvolume{59}(\bissue{4}),
\bfpage{2082}--\blpage{2102}
(\byear{2013})
\end{barticle}
\endbibitem

\bibitem{kafle2021one}
\begin{bchapter}
\bauthor{\bsnm{Kafle}, \binits{S.}},
\bauthor{\bsnm{Joseph}, \binits{G.}},
\bauthor{\bsnm{Varshney}, \binits{P.K.}}:
\bctitle{{One-bit Compressed Sensing Using Untrained Network Prior}}.
In: \bbtitle{{Proceedings of IEEE International Conference on Acoustics, Speech
  and Signal Processing (ICASSP)}},
pp. \bfpage{2875}--\blpage{2879}
(\byear{2021})
\end{bchapter}
\endbibitem

\bibitem{van2008visualizing}
\begin{botherref}
\oauthor{\bparticle{Van~der} \bsnm{Maaten}, \binits{L.}},
\oauthor{\bsnm{Hinton}, \binits{G.}}:
{Visualizing Data Using t-SNE}.
{Journal of Machine Learning Research}
\textbf{9}(11)
(2008)
\end{botherref}
\endbibitem

\bibitem{gu2021interpreting}
\begin{bchapter}
\bauthor{\bsnm{Gu}, \binits{J.}},
\bauthor{\bsnm{Dong}, \binits{C.}}:
\bctitle{{Interpreting Super-Resolution Networks with Local Attribution Maps}}.
In: \bbtitle{{Proceedings of IEEE Conference on Computer Vision and Pattern
  Recognition (CVPR)}},
pp. \bfpage{9199}--\blpage{9208}
(\byear{2021})
\end{bchapter}
\endbibitem

\bibitem{zhang2017learning}
\begin{bchapter}
\bauthor{\bsnm{Zhang}, \binits{K.}},
\bauthor{\bsnm{Zuo}, \binits{W.}},
\bauthor{\bsnm{Gu}, \binits{S.}},
\bauthor{\bsnm{Zhang}, \binits{L.}}:
\bctitle{{Learning Deep CNN Denoiser Prior for Image Restoration}}.
In: \bbtitle{{Proceedings of IEEE Conference on Computer Vision and Pattern
  Recognition (CVPR)}},
pp. \bfpage{3929}--\blpage{3938}
(\byear{2017})
\end{bchapter}
\endbibitem

\end{thebibliography}

\end{document}